\documentclass{mva_style}
\usepackage{graphicx}
\usepackage{times}
\usepackage{epsfig}
\usepackage{amsmath}
\usepackage{amssymb}
\usepackage{amsthm}
\usepackage{mathtools}
\usepackage{times}
\usepackage{epsfig}
\usepackage{subfig}
\usepackage[ruled]{algorithm2e}
\usepackage{mathtools}
\usepackage{xcolor,colortbl}
\usepackage{xurl}
\usepackage{hyperref}
\hypersetup{
    colorlinks=true,
    linkcolor=blue,
    filecolor=magenta,      
    urlcolor=cyan,
    pdftitle={Overleaf Example},
    pdfpagemode=FullScreen,
    }

\finalcopy 

\begin{document}
\title{INV-Flow2PoseNet: Light-Resistant Rigid Object Pose from Optical Flow of RGB-D Images using Images, Normals and Vertices}

\author{
  \\Torben Fetzer$^1$\\
  \and
  \\Gerd Reis$^2$
  \and
  \\Didier Stricker$^{1,2}$
  \and
  \vspace{-0.8cm}\\$^1$Department of Computer Science, University of Kaiserslautern\\
  $^2$Department Augmented Vision, DFKI GmbH
  \and
  {\tt\small \{torben.fetzer, gerd.reis, didier.stricker\}@dfki.de}
}

\maketitle
\section*{\centering Abstract}
\textit{
This paper presents a novel architecture for simultaneous estimation of highly accurate optical flows and rigid scene transformations for difficult scenarios where the brightness assumption is violated by strong shading changes. 
In the case of rotating objects or moving light sources, such as those encountered for driving cars in the dark, the scene appearance often changes significantly from one view to the next.
Unfortunately, standard methods for calculating optical flows or poses are based on the expectation that the appearance of features in the scene remain constant between views.
These methods may fail frequently in the investigated cases.\\
The presented method fuses texture and geometry information by combining image, vertex and normal data to compute an illumination-invariant optical flow. 
By using a coarse-to-fine strategy, 
globally anchored optical flows are learned, reducing the impact of erroneous shading-based pseudo-correspondences.
Based on the learned optical flows, a second architecture is proposed that predicts robust rigid transformations from the warped vertex and normal maps.
Particular attention is payed to situations with strong rotations, which often cause such shading changes.
Therefore a 3-step procedure is proposed that profitably exploits correlations between the normals and vertices.\\
The method has been evaluated on a newly created dataset containing both synthetic and real data with strong rotations and shading effects. 
This data represents the typical use case in 3D reconstruction, where the object often rotates in large steps between the partial reconstructions.
Additionally, we apply the method to the well-known Kitti Odometry dataset. 
Even if, due to fulfillment of the brighness assumption, this is not the typical use case of the method, the applicability to standard situations and the relation to other methods is therefore established.
}

\section{Introduction}
3D reconstructions of objects and depth information of scenes play an increasingly important role in industry.
Whether it is quality control in production or the recognition of the environment in autonomous driving, the number of applications is continuously increasing.
Due to the simplicity of applicability, depth cameras are more and more used in parallel to flexible 3D scanners, and the availability of depth data for a wide variety of applications is steadily increasing.
At the same time, the demand for scene understanding methods, represented by optical flow estimation, is constantly increasing, especially in the field of automation.
Since in addition to images alone, more and more information is available, also the demand for higher quality scene understanding increases.

For the vast majority of applications, rigid scenes can be assumed and taken into account. 
And even for dynamic scenes, the optical flow can be approximated by rigid models if not too large motions of the camera or the environment are expected.
This rigidity assumption can even guide the estimation of optical flow, whose accuracy can benefit from it.
The simultaneous extraction of the rigid transformation between two subsequent frames is then also desirable.
In this way, the method can be used for automatic alignment of point clouds in difficult scenarios including large motion and rotation, that yield strong shading changes.

The presented method uses an optical flow approach based on \textit{PWC-Net}, that has been adapted to use data from texture images, normal maps and vertex maps simultaneously.
This optical flow method is moreover combined with the extraction of rigid transformations, that are computed from the normal and vertex maps, warped by the predicted optical flow. 
The so predicted pose can mutually benefit from the coarse to fine strategy of the optical flow, since the optical flow can find dense correspondences over the whole scene using a pyramidal approach, even in the presence of large motion. 
Textural, geometric and shading features are included, which partly compensate for each other's weaknesses (sparsity of normal and vertex maps, illumination susceptibility of the texture images).
From the warped 3D information of the scene, the rigid transformation can then be stably determined in a second step.

\begin{figure*}[t]
    \subfloat[SIFT Matches]{
    \includegraphics[height=.195\textwidth]{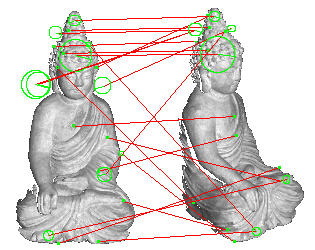}}
    \subfloat[Illuminated Scene]{
    \includegraphics[height=.195\textwidth]{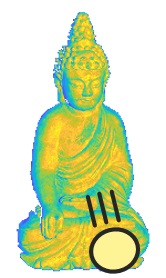}
    \includegraphics[height=.195\textwidth]{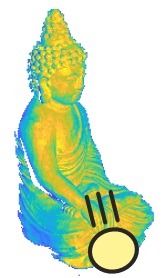}}
    \subfloat[Overlap of Views]{
    \includegraphics[height=.195\textwidth]{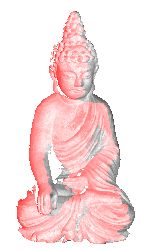}
    \includegraphics[height=.195\textwidth]{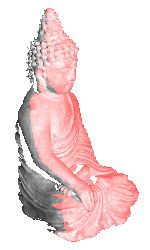}}
    \subfloat[Flow Matches]{
    \includegraphics[height=.195\textwidth]{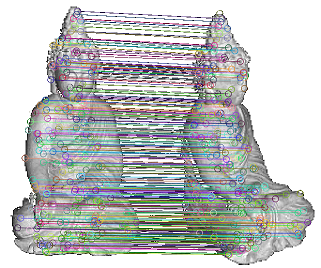}}
    \caption{(a) and (d) show matches based on SIFT features and optical flow. (b) shows the scene, which has been illuminated by a strong spot light, in a different color space. This more clearly visualizes the different shadings of the object, which is the reason for the failure of the common method based on SIFT features. (c) shows overlapping regions of subsequent scans. Even a rotation of approx. 45° yields a large overlap of more than 80\%.\vspace{-0.3cm}}
    \label{Matches_Overlap}
\end{figure*}

\subsection{Motivation: Flow-Based Alignment}
In order to compute the alignment of two subsequently reconstructed frames, usually robust and transformation invariant features (SIFT, KAZE, ...) are detected and matched between the frames.
Robust and outlier resistant methods like RANSAC based \textit{PnP-solvers} are subsequently used to compute the rigid transformation between the views.
It is commonly known, that this approach, applied with some few good features only, results in way better alignments than using many worse features jointly.
Modern deep learning approaches adopt this scheme and deliver competitive results on a wide range of data in real time.

The basis of all common feature methods is the \textit{brightness assumption}, which expects that the appearance of the object does not change significantly from one frame to another. This is fulfilled for many applications, especially when the camera moves smoothly through a scene or an object undergoes slow motion. 
If, on the contrary, the direction of the light incidence changes, the shading of the scene also differs dramatically and the brightness assumption gets strongly violated.
This leads to a very probable failure of the standard methods based on this requirement, especially in the following situations:
\begin{itemize}
    \item Outdoor scenes where lighting conditions can change suddenly. 
    This can occur from direct sun light, as well as indirect light reflections from other objects.
    \item Moving objects, especially rotating ones, inevitably change the direction of light incidence. 
    This leads in particular to considerable difficulties in the application area of 3D reconstruction, where the object is often rotated in order to capture it successively from all sides.
    \item Driving cars in the dark may cause strong shading differences in the captured images of the environment. 
    Visible elements in the scene are illuminated by the car's headlights. 
    These light sources move together with the car through the scene, which may yield strong variation of the direction of light incidence.
\end{itemize}
In order to illustrate this problem and investigate it, a setup with a static direct light source, a static camera and a rotated object is considered.  
Figure \ref{Matches_Overlap} (a) shows how the standard approach based on SIFT matches fails, due to different light incidence.
Figure \ref{Matches_Overlap} (b) shows the scene in a different color space, where the different shading becomes more obvious.  
While the features in the scene change in appearance, it can still be assumed that a significant portion of the scene overlaps in the different views.
In the important case of object rotation in 3D reconstruction, our research shows that in the vast majority of cases a typical rotation of 45° still yields more than 80\% overlap of the scene.
Figure \ref{Matches_Overlap} (c) visualizes the overlapping areas of the two views.
Optical flow methods can benefit from this in turn, as they view and match the motion as a whole, using pyramidal approaches.
Finally, Figure \ref{Matches_Overlap} (d) shows correspondences determined using an optical flow method, as introduced in the following. The correspondences do contain noise and smaller errors, especially in feature-poor regions. 
They are nevertheless capable of predicting stable orientations of the object, significantly more stable ones than feature-based methods.

\section{Related Work}
Optical flow estimation is a well-known problem in applied machine vision and has wide spread use cases in industrial applications such as robotics, automotive driving, and quality control.
The task is to determine dense motion at pixel level between image pairs as accurately as possible.
Starting with the method of Horn and Schunck \cite{horn1981determining}, variational methods were the state of the art for a long time. 
Since the problem itself is an ill-posed problem, further assumptions have to be made on the flow field, which led to a multitude of different methods that use the most diverse regularization procedures to make the problem solvable according to the specific application.
In recent years the problem of \textit{optical flow} estimation increasingly expanded to the problem of \textit{scene flow} estimation, which deals with the 3D motion of scene points in space, whereas optical flow was limited to 2D point motion on the image plane.
Based on the variational approaches for optical flow, a number of variational scene flow methods have been developed.
Most of them use rectified stereo image pairs as input and thus estimate scene flow with different regularization methods or partial rigidity assumptions (\cite{vcech2011scene}, \cite{huguet2007variational}, \cite{isard2006dense}, \cite{li2008multi}, \cite{basha2013multi}, \cite{park2012tensor}, \cite{zhang2012dense}, \cite{ferstl2014atgv}).
At the same time, methods were developed to determine the scene flow directly from RGB-D data. 
With an increasingly number of depth sensors that became available, this approach is quite justified.
Several variants of methods handle this case (\cite{letouzey2011scene}, \cite{gottfried2011computing}, \cite{herbst2013rgb}, \cite{quiroga2014dense}).

The appearance of \textit{FlowNet} \cite{dosovitskiy2015flownet} revolutionized the field of optical flow estimation.
It became possible to treat the problem in real time with the help of \textit{convolutional neural networks (CNNs)}. 
In contrast, the variational methods were extremely time consuming and computationally expensive.
A higher accuracy at the expense of a much larger network was subsequently achieved with \textit{FlowNet2} \cite{ilg2017flownet}.
This was followed by the release of \textit{PWC-Net} \cite{sun2018pwc}, which uses warping layers at different levels of an image pyramid, representing the current state of the art that is in addition much smaller than the previously released \textit{FlowNet2}.
Based on \textit{PWC-Net}, Saxena \textit{et al.} have presented a method for estimating scene flow from rectified stereo image pairs. 
In addition, they handle occlusions within the forward pass.
Previous methods required at least one forward and one backward warping to stably detect occlusions (\cite{hur2017mirrorflow}, \cite{meister2018unflow}, \cite{wang2018occlusion}).

Similar to earlier in the variational path, methods that extract scene flow directly from RGB-D data also evolved over time.
Qiao \textit{et al.} showed how scene flow based on \textit{FlowNet} can be improved by fusion with features of depth data extracted in an extra network pass.
Based on \textit{PWC-Net}, Rishav \textit{et al.} \cite{rishav2020deeplidarflow} use depth data from a Lidar sensor to determine scene flow. 
In doing so, they account for the lower resolution of Lidar data using appropriate reliability weights from \cite{eldesokey2019confidence}.
In general, scene flow networks based on RGB-D data show poor performance for outdoor scenes, due to range limitations of the sensors.
A number of approaches attempt to address this issue (\cite{wang2018occlusion}, \cite{yang2018every}, \cite{yin2018geonet}, \cite{zou2018df}).
Since the omission of active components removes the range limitations, but is accompanied by a loss of quality of the depth information, we will nevertheless restrict ourselves to this limited case.
We are content with the scene flow within the sensor limits, since it is sufficient for an overwhelming number of practical applications, where the limits of the sensor can be planned accordingly.
\newline

In order to predict the pose of an object, a long time RANSAC approaches using explicit pose estimates based on the singular value decomposition were used.
In recent years, first deep learning approaches predicted the pose directly using neural networks.
Kendall \textit{et al.} \cite{kendall2015posenet} use in their \textit{PoseNet} several convolutional layers, followed by linear layers to directly predict rotation and translation from RGB images.
This way, they were the first to solve the problem of camera relocalization in static scenes by a deep learning approach.
A few years later Vijayanarasimhan \textit{et al.} \cite{vijayanarasimhan2017sfm} extend this principal in \textit{SfM-Net} in order to predict simultaneously the rigid transformations and the depth of the scene.
They basically adopt the principals of the famous \textit{Structure-from-Motion} pipeline to a deep learning framework.
In parallel Zhou \textit{et al.} \cite{zhou2017unsupervised} developed a related model and showed how to train it in an unsupervised manner.

Finally, there has been a row of methods for direct point cloud registration with deep learning.
Some of them replace parts of the standard strategies by deep learning methods and some try to replace the full pipeline.
A large number of different approaches, correspondence-based and correspondence-free, are reviewed in \cite{zhang2020deep} and \cite{villena2020deep}.

Nevertheless, an automatic and light resistant flow-based pose-estimation method, that works correspondence-free, and takes geometrical, textural and coherent scene motion into account has never been addressed before.
\begin{figure}[b]
    \centering
    \includegraphics[width=.47\textwidth]{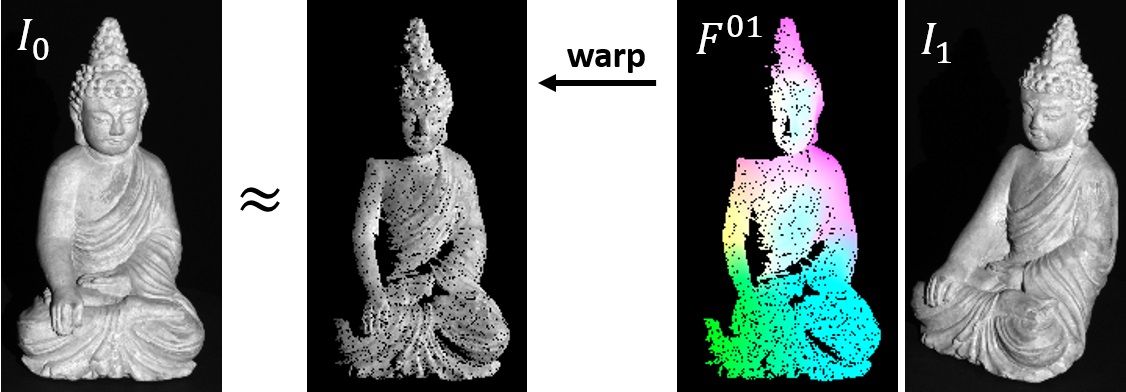}
    \caption{Image $I_0$ in comparison to image $I_1$ that has been warped by optical flow $F^{01}$.
    Assuming consistent brightness, these should be identical (ignoring masked pixels due to the semi-dense optical flow from real data).
    In case of strong rotations of the object the shading changes dramatically, which violates this assumption.
    }
    \label{warpedTexture}
\end{figure}
\begin{figure*}[t]
    \includegraphics[width=.99\textwidth]{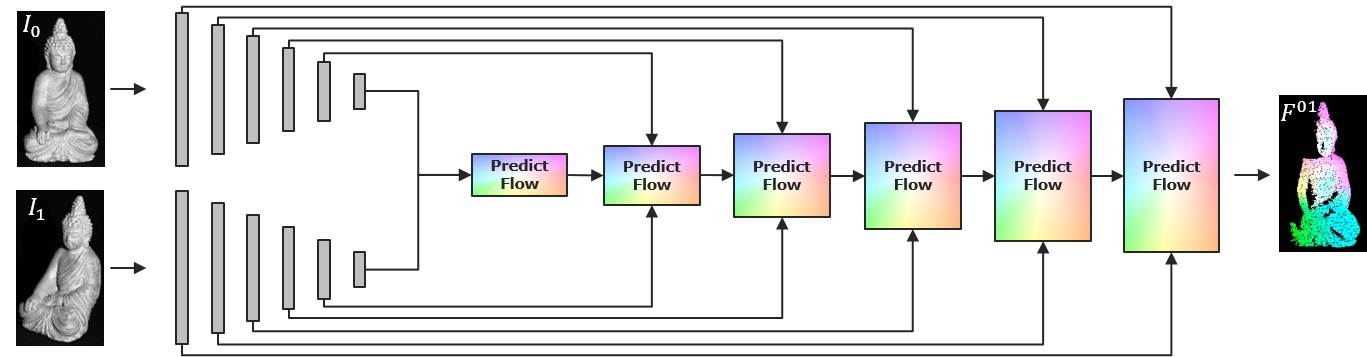}
    \caption{Sketch of the \textit{PWC-Net} architecture. The input is convoluted through multiple layers and the optical flow is predicted starting from the lowest level upwards in a U-Net structure. In each level, the layers of $I_1$ are warped towards the layers of $I_0$ in order to provide initial flows from previous lower levels. With this pyramidal approach also large flows are predictable with quite small filter kernels.\vspace{-0.3cm}}
    \label{PWCNetArchitecture}
\end{figure*}

\section{Light-Resistant Optical Flow}
The \textit{optical flow} between two images is understood as the displacements of the individual pixels from one to the other image.
Determining the optical flow between images of a scene often serves the purpose of scene understanding, as it directly allows the analysis of a large amount of scene information:
\begin{itemize}
    \item The optical flow between calibrated camera images from different perspectives of the same static scene allows theoretically dense point correspondences and accompanying depth data.
    \item The optical flow between static camera images of a moving scene theoretically allows the analysis of scene motion and object tracking. 
    If depth data is additionally available, the \textit{scene flow}, i.e. the spatial movement of the points in the scene, can be calculated.
\end{itemize}
In the estimation of the optical flow between two consecutive images $I_0$ and $I_1$, a horizontal and a vertical displacement field ($F_x^{01}$, $F_y^{01}$) are calculated, mapping each pixel in image $I_0$ to its corresponding pixel in image $I_1$.
The usual basis of the estimation is the brightness assumption, which assumes that corresponding pixels have the same appearance in the different images: 
\begin{align}
    I_0(x,y) \approx I_1(x+F_x^{01},y+F_y^{01})\label{BrightnessAssumption}
\end{align}
Figure \ref{warpedTexture} shows image $I_0$ and besides image $I_1$, which has been warped by the optical flow $F^{01}$.
Since the used optical flow has been computed from real data, the flow field is semi-dense and contains some masked pixels.
Such errors will be addressed later on, where we will also show how to adopt filters to sparse, semi-sparse and mixed data.
Instead of looking for exactly the same values between $I_0$ and $I_1$, filtered values are considered in a regional context in order to robustify the matching. 
Deep neural networks have proven to be extremely effective for this purpose. 
The current state of the art is given by \textit{PWC-Net}, which will be briefly introduced in the following to serve as a basis for the subsequently presented light-resistant variant.

\subsection{PWC-Net}
\textit{PWC-Net} combines classical techniques such as a pyramidal approach, warping and correlation to create a highly effective network for optical flow estimation. 
The input images are passed through a pyramid of convolutions which extract rotation- and translation-invariant features at different levels of the receptive field.
From the lowest level, cost volumes based on extracted features are established from which the optical flow is effectively predicted.
These flows are refined upwards with each level, incorporating new features of the current level and the flows and more global features from previous levels.
By warping the data using the previous flow, the search space is significantly reduced and even large displacements can be treated and predicted with this comparatively small network. 
Figure \ref{PWCNetArchitecture} depicts the architecture of the network.
Each prediction block consists of a cost volume for flow prediction and is fed with the corresponding layer in a U-Net structure, in order to predict a flow field in full resolution.
Note that the standard network presented by Sun \textit{et al.} in \cite{sun2018pwc} predicts the optical flow up to the second last level and afterwards refines the resulting flow by a context-network as post processing.
This results in a final optical flow whose resolution is only $\frac{1}{16}$ of the input images' resolution.
Instead of upsampling by variational methods, we go for two additional texture-guided upsampling steps within the network, in order to provide full resolution optical flows within a single training routine.
\begin{figure*}[t]
    \centering
    \includegraphics[width=.95\textwidth]{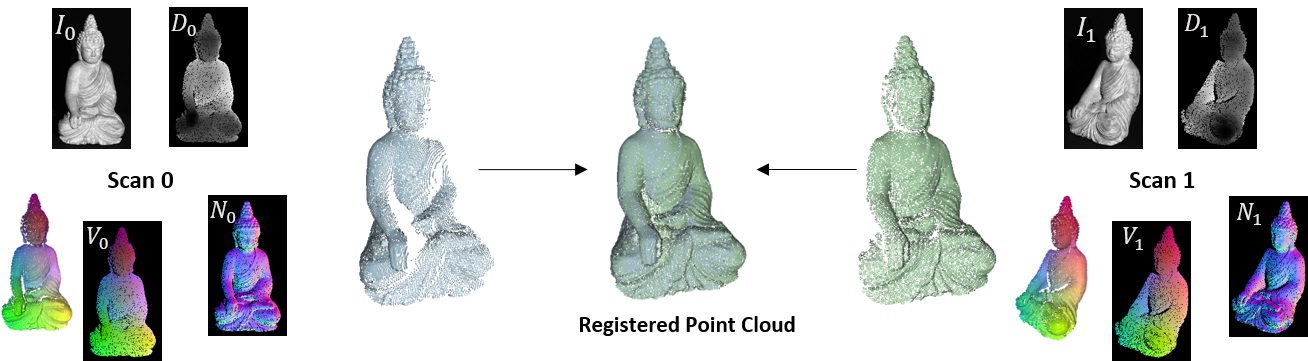}
    \caption{Possible input that is available to the task of light resistant optical flow estimation and subsequent pose prediction. In addition to texture images, there are depth maps, vertex maps, point clouds and normal maps available. As well the depth maps as the vertex maps contain geometrical information. Since the vertex maps are independent of the calibration it is the preferable choice for the presented method.\vspace{-0.3cm}}
    \label{PossibleInput}
\end{figure*}
\begin{figure}[b!]
    \includegraphics[width=0.48\textwidth]{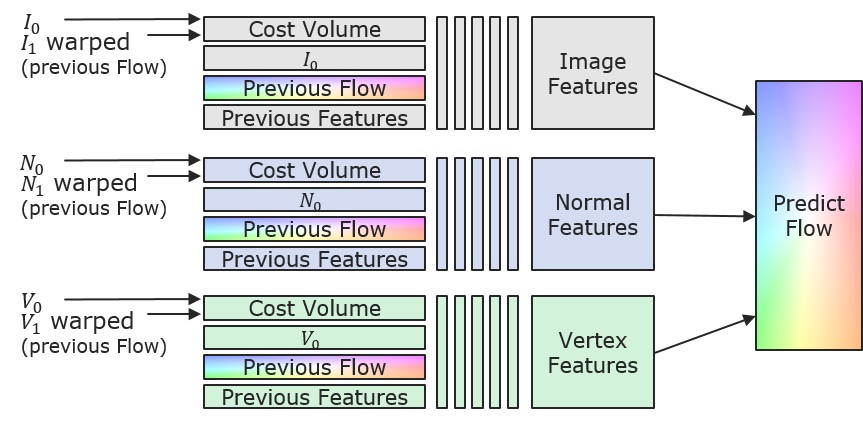}
    \caption{Flow prediction architecture in each layer (except first one). Features of images (texture), normals (shading) and vertices (geometry) are extracted separately and jointly fed to the prediction module.}
    \label{PredictFlow}
\end{figure}

\subsection{INV-Net using Images, Normals and Vertices}\label{Sec:INVNet}
\begin{figure*}[t]
    \centering
    \includegraphics[width=.95\textwidth]{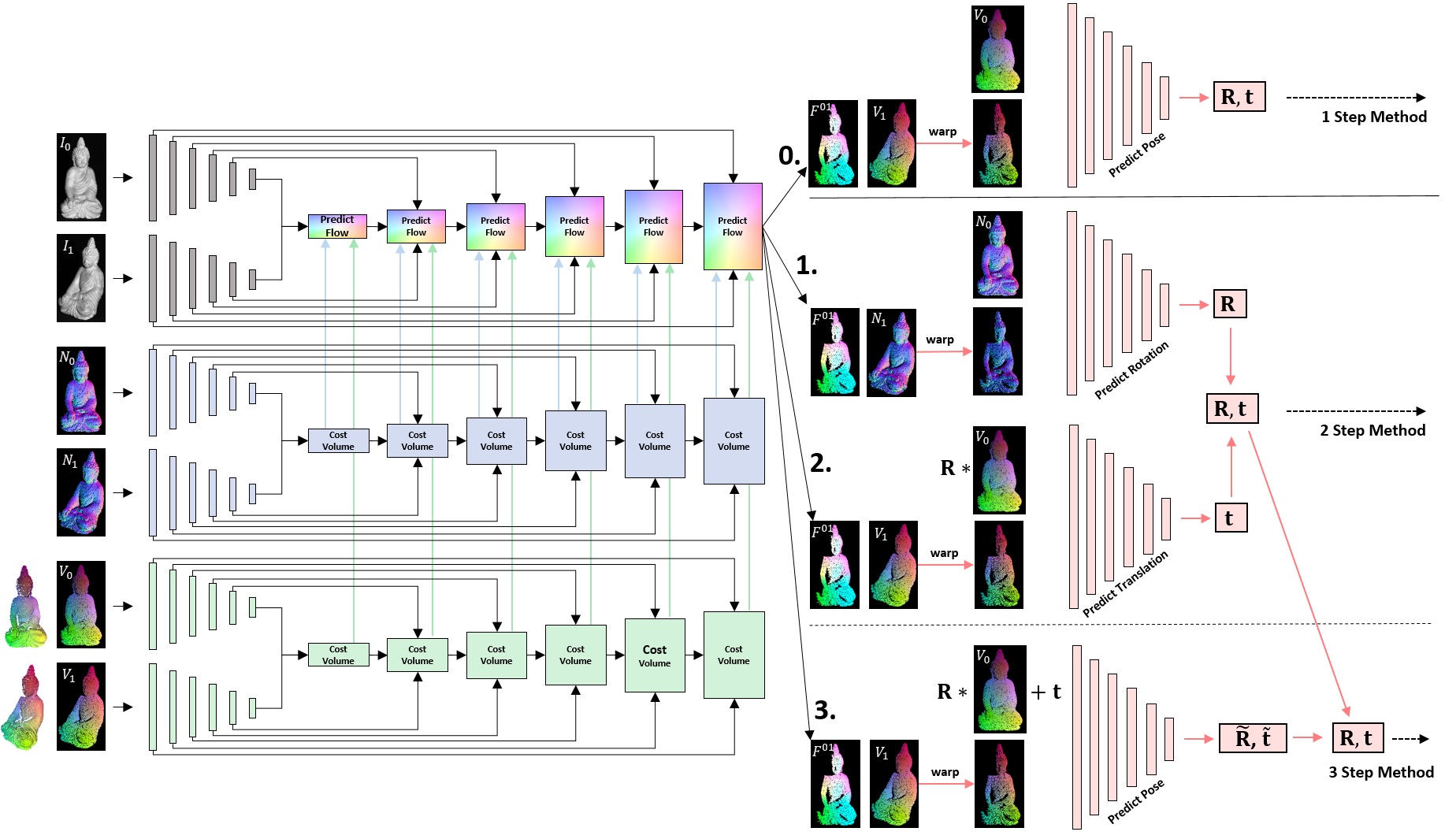}
    \caption{Architecture of \textit{Flow2PoseNet}. The left part of the network aims to predict accurate optical flow from images, normal- and vertex-maps, using textural features from images, shading features from normals and geometrical features from vertices in order to predict accurate and light resistant flow fields. The pose of the rigid scene is computed in three steps from the warped normal- and vertex-maps. The first step predicts the normals from the warped normal-maps. The second step predicts the translation from the warped and rotated vertex-maps. The third step predicts a correction transformation to refine the predicted rotation and translation incrementally.\vspace{-0.2cm}
    }
    \label{Flow2PoseNet}
\end{figure*}

Classical \textit{PWC-Net} uses texture images only.
Unfortunately, for the investigated use case these texture images may be disturbed due to shading changes, resulting from rotations of the object or light changes, which would make the network likely to fail due to a violated brightness assumption (see Figure \ref{Matches_Overlap}).
In many situations, where depth data is available, a lot of additional information can be provided to the network, that is invariant under the shading effects related to light changes or object rotations:
\begin{itemize}
    \item Texture images $I_0$ and $I_1$ that underlay shading effects. Nevertheless, they provide full and dense data, which can deliver local context.
    \item Depth maps $D_0$ and $D_1$ that store the relative geometrical information of the scene, light- and shading-invariant, with respect to the camera center.
    Due to measuring errors there may be outliers or data-less pixels, resulting in semi-dense depth maps.
    \item Vertex maps $V_0$ and $V_1$ that store the spatial information of the scene, light- and shading-invariant in three channels of a map in image resolution. 
    They are computed from the depth maps and the available camera calibration in order to store the geometrical information calibration independent.
    Therefore, they are similar to the depth maps semi-dense maps with masked pixels.
    Moreover, they are structured representations of point clouds, that allow to perform neighboring operations on 3D data in 2D space, which yields large advantages in the following approach.
    \item Normal maps $N_0$ and $N_1$ that store spatial information of the surfaces in the scene. 
    They are related to partial derivatives of the 3D vertices and do not underlay scaling and translation bias. 
    They are in a specific range and responsible for a large amount of shading features of a scene (where standard methods based on fulfilled brightness assumption get a large amount of information from), without being disturbed by the light changes. 
    They can be directly computed from the vertex maps, using the topological information given by the image grid (see \cite{holzer2012adaptive}).
    Unfortunately, they thus also inherit the semi-density from underlying vertex maps.
\end{itemize}
Figure \ref{PossibleInput} sketches the basic problem of finding a light resistant pose estimation from all the available input.
The first task is to find a light resistant high quality optical flow from this large amount of input data.
As well depth maps as vertex maps store the spatial information of reconstructed surface points.
Since they are somehow interchangeable we use the vertex maps only.
This way the method becomes independent of the intrinsic calibration to the cost of a higher amount of data that needs to be processed.
Figure \ref{Flow2PoseNet} (left part) sketches the basic network, that takes features from \textit{images} (textural features), \textit{normal maps} (shading features) and \textit{vertex maps} (geometrical features). 
Thereby we follow the basic principal of \textit{PWC-Net} but run the different input through separate feature pipelines and set up independent cost volumes, that contribute to the flow prediction.
All features are processed as in \cite{sun2018pwc} and fed to the pose prediction in each layer.
This way the network learns to treat the feature appropriate and to get advantages from all.
Figure \ref{PredictFlow} depicts the prediction procedure in each layer, except the first one, where only the cost volumes are used for initialization of the flow.

\paragraph{Normalized Convolutions}
In order to take into account the semi-density of the vertex maps and the normal maps, the convolutions, leading to the first layer are replaced by  \textit{normalized convolutions} as introduced by Eldesokey \textit{et al.} in \cite{eldesokey2018propagating}.
Using the following slightly changed convolution procedure, the known masks can be used to ensure that data-less pixels do not contribute to the convolution with respect to neighbored pixels.
Suppose, we are given a signal $\textbf{A}$ to be convolved with a filter kernel $\textbf{K}$.
Further assume that the measurements of the signal $\textbf{A}$ are of varying quality with a confidence measure $\textbf{W}$ of the same size as $\textbf{A}$ having values between 0 and 1 to describe these uncertainties.
It is desired to use the confidence measure as a weighting of the entries of $\textbf{A}$ during convolution to ensure that reliable measurements have a higher influence on the convolution signal than inferior measurements or missing data for certain points.
For this purpose, each summand within the convolution is weighted accordingly and divided by the sum of the weights to ensure the normalized character of the convolution.
In detail, the normalized convolution of signal $\textbf{A}$, convolved with kernel $\textbf{K}$ and weighted by confidence $\textbf{W}$ around data point $[n]$ is given by
\begin{align}
\begin{split}
    \big(\textbf{K}*\textbf{A}\big)_\textbf{W}[n]&= \frac{\sum_m\textbf{K}[m]\cdot\textbf{W}[n-m]\cdot\textbf{A}[n-m]}{\sum_m\textbf{K}[m]\cdot\textbf{W}[n-m]}\\
    &= \frac{\big(\textbf{K}*(\textbf{W}\odot\textbf{A})\big)[n]}{\big(\textbf{K}*\textbf{W}\big)[n]}\ ,
\end{split}
\end{align}
where $\odot$ denotes the element-wise \textit{Hadamard-Product}.
In order to avoid influence of missing pixels, a binary mask, that contains zeros in case of missing data and ones otherwise, can be fed to the convolutions as confidence $\textbf{W}$ .

\begin{figure}[h]
    \centering
    \includegraphics[width=.47\textwidth]{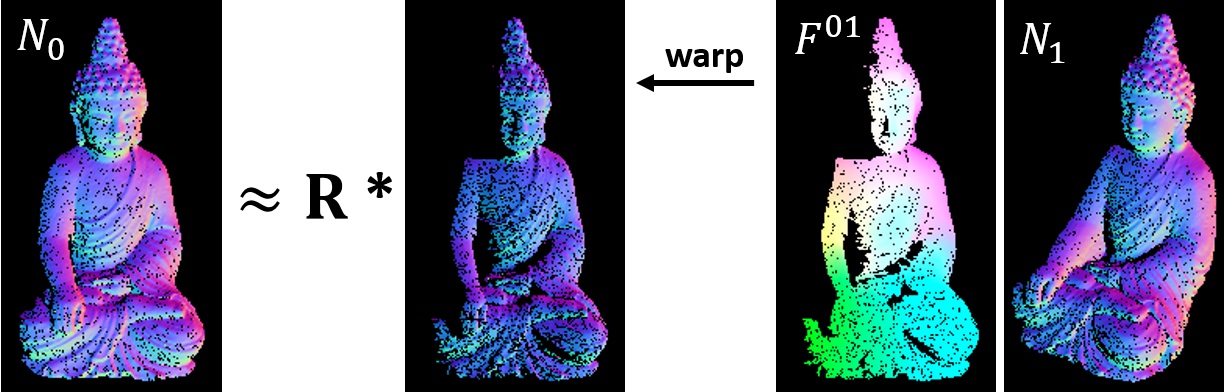}\\
    \includegraphics[width=.47\textwidth]{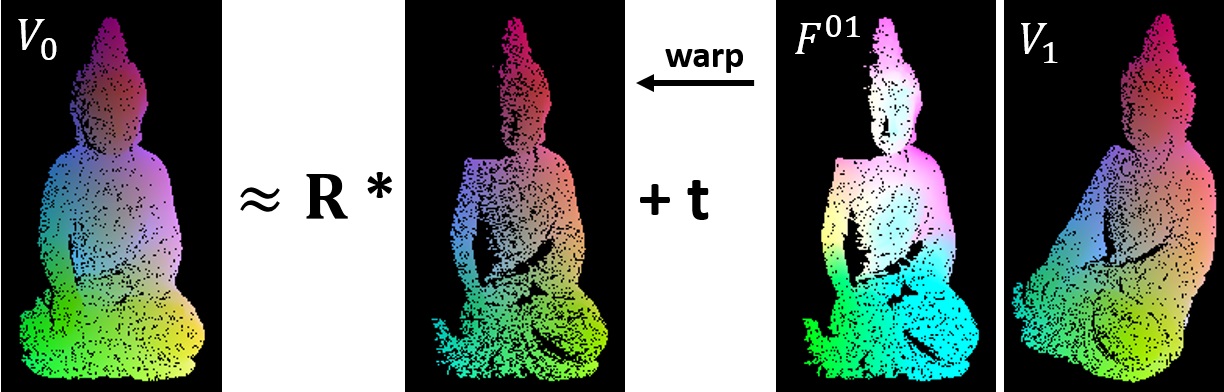}
    \caption{Normal maps and vertex maps that have been warped by optical flow $F^{01}$.
    Assuming rigid scenes, normals should be identical up to a rotation, vertices up to a rotation and a translation.\vspace{-0.2cm}
    }
    \label{warpedNormalsVertices}
\end{figure}

\paragraph{Consistency Assumptions}
Similar to the brightness assumption given in Equation (\ref{BrightnessAssumption}) the following consistency assumptions hold true for normals and vertices of rigid scenes:
\begin{align}
    &N_0(x,y) \approx \textbf{R}N_1(x+F_x^{01}, y+F_y^{01})\label{NormalConsistency}\\
    &V_0(x,y) \approx \textbf{R}V_1(x+F_x^{01}, y+F_y^{01})+\textbf{t}\label{VertexConsistency}
\end{align}
Figure \ref{warpedNormalsVertices} visualizes the consistency relations for normals and vertices.
While the pixels of the warped normal map coincide with the reference normals up to a rotation matrix $\textbf{R}$, the vertices coincide up to rotation $\textbf{R}$ and a translation vector $\textbf{t}$.
These relations will be essential later on, in order to extract the rigid pose from the given optical flow.
A very important result of our research is, that features, computed from filtered normal and vertex maps allow for computation of accurate optical flows.
This means, the standard approach for feature extraction from images (as used in \textit{PWC-Net}), is suitable to compute rotation- and transformation-invariant features from normal and vertex maps, as well.
\section{Pose from Warped Normals and Vertices}\label{Sec:PoseFromFlow}
Several research works have already shown that it is possible to predict the relative pose of two views of a scene using neural networks.
Usually, features are detected, matched, outliers are rejected and then passed through a series of layers in order to get representative feature vectors. 
Finally, as introduced in \cite{kendall2015posenet}, relative translation and rotation are predicted jointly using at least two subsequent fully connected layers.

In the previous section, a light-resistant optical flow has been computed by \textit{INV-Net}.
Based on this, it is not necessary to search for matches in the entire image.
Considering images, normal maps and vertex maps from the first view and the ones from the second view, that have been warped towards the first one with the computed optical flow, the data at each pixel-position theoretically matches densely.
Of course, there are also many erroneous and inaccurate regions in the flow field, especially in feature-poor areas, where the flow is mainly interpolated.
Previous work has shown that in general more accurate poses are estimated when only a few good features are used for the calculation, instead of many less good ones.
This can also be done by feature extraction from the warped normal and vertex maps. 
It should be noted that in areas where good features for the pose estimation can be found, also good optical flows are available. 
In a way, both the optical flow and the subsequently calculated pose are based on these same good features.
Nevertheless, in the case of low quality features, as is the case with texture-poor and smooth surfaces, or even many false features due to light changes, we benefit from the more general information of the dense flow field.

In order to obtain best poses from the warped vertex and normal maps, we investigated two different approaches (\textit{1 Step Method} and \textit{2 Step Method}), that have led to the conclusion that a combination \textit{3 Step Method} performs best, as it compensates for the weaknesses of each method.
\vspace{-0.2cm}
\paragraph{1 Step Method}
This approach uses the concatenated warped vertex maps to extract jointly rotation matrix $\textbf{R}$ and translation vector $\textbf{t}$ that align the vertex maps rigidly.
The relation is based on consistency assumption (\ref{VertexConsistency}).
Note that after warping, the matching vertices are theoretically placed at the same location in the concatenated input.
Due to convolutional layers the network is able to extract reliable locations, where a more accurate optical flow has been provided.
The basic structure is shown in Figure \ref{Flow2PoseNet} at branch \textbf{0.} on the right.
\vspace{-0.2cm}
\paragraph{2 Step Method}
This approach uses two steps to predict rotation and translation individually by two separate networks. 
Following the consistency property of Equation \ref{NormalConsistency}, the warped normal map $N_1$ and the reference normal map $N_0$ are related by a rotation matrix $\textbf{R}$ only. 
In a first step, this relative rotation $\textbf{R}$ is predicted by stacking $N_0$ and the warped $N_1$ and processing them through several convolutional layers, followed by two fully connected layers in order to predict optimal rotation with respect to the normals.

Based on the third consistency property of rigid transformations, given in Equation \ref{VertexConsistency}, the translation $\textbf{t}$ is predicted from the warped vertex map $V_1$, that has been rotated by matrix $\textbf{R}$ and the reference vertex map $V_0$.
Rotation matrix $\textbf{R}$, from the previous step, has been applied in order to get dependency on the translation vector $\textbf{t}$ for this inference step only.
The structure is again shown in Figure \ref{Flow2PoseNet} at branches \textbf{1.} and \textbf{2.} on the right.
\vspace{-0.2cm}
\paragraph{3 Step Method}
Rotation and translation are two fundamentally different operations that have a strong influence on each other.
The smaller a rotation, the better it can be approximated linearly.
Unfortunately, the joint extraction as in \textit{1 Step Method} may yield inaccuracies in case of large rotations.
In these situations, it may be beneficial to extract them separately like in the \textit{2 Step Method}.
Nevertheless, small rotational errors, from the first step of this approach influence the predicted translation from the second step.
\begin{figure}[b!]
    \centering
    \includegraphics[width=.4\textwidth]{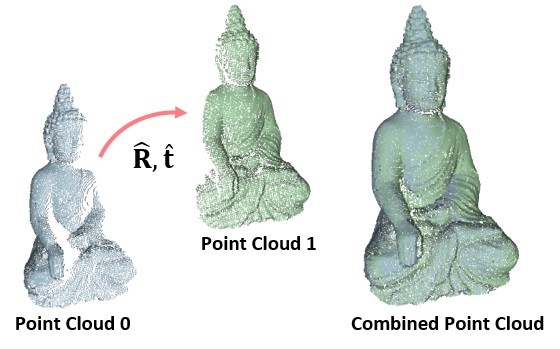}
    \caption{Point clouds of the two exemplary views. The resulting transformation $P=(\hat{\textbf{R}},\hat{\textbf{t}})$ aligns the point cloud of the first view to the one of the second view. The registered combined point cloud is shown besides.
    }
    \label{Alignment}
\end{figure}

The idea of the \textit{3 Step Method} is to first apply the \textit{2 Step Method} to pre-align the vertex maps.
In a third step a correctional rotation matrix $\tilde{\textbf{R}}$ and a correctional translation vector $\tilde{\textbf{t}}$ are jointly predicted from the warped and pre-transformed vertex map $\textbf{R}V_1+\textbf{t}$ and reference vertex map $V_0$.
The final pose $P=(\hat{\textbf{R}},\hat{\textbf{t}})$ is then given by:
\begin{align}
    \hat{\textbf{R}}=\tilde{\textbf{R}}\textbf{R},\qquad\hat{\textbf{t}}=\tilde{\textbf{R}}\textbf{t}+\tilde{\textbf{t}}
\end{align}
For extracting this correctional transformation the \textit{1 Step Method} is used. 
This is beneficial, since the correctional rotations are usually small, which makes it possible to predict the rotation and the translation jointly in order to avoid weaknesses of successive prediction as in the \textit{2 Step Method}.
The structure is again depicted in Figure \ref{Flow2PoseNet} at branches \textbf{1.}, \textbf{2.} and \textbf{3.} on the right.

\begin{figure*}[t!]
    \centering
    \subfloat[Models of synthetic training data]{
    \includegraphics[scale=0.33]{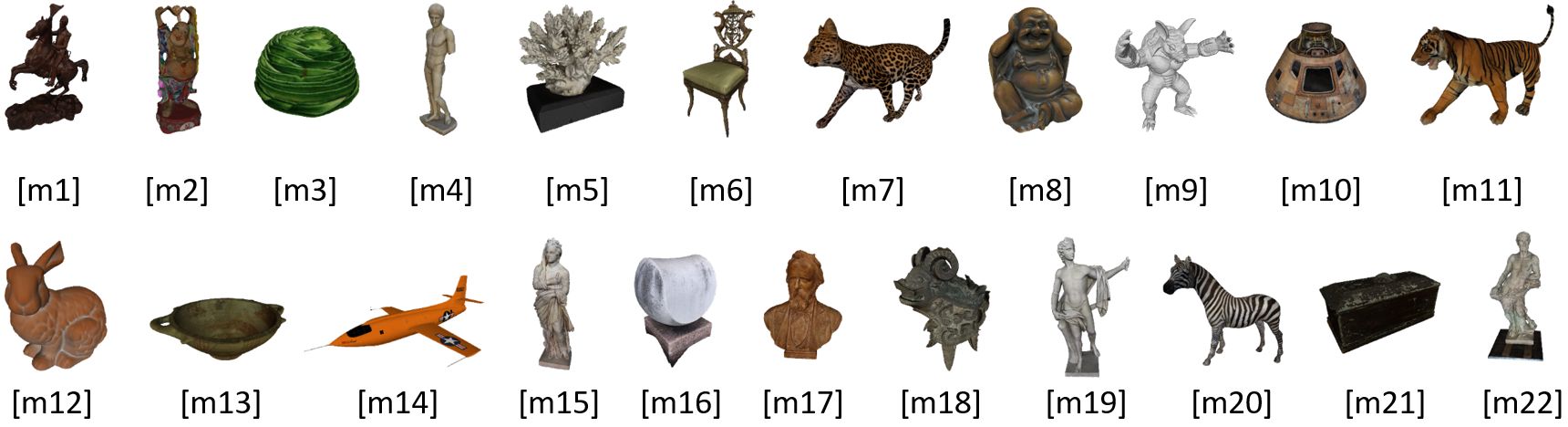}}\\
    \subfloat[Models of synthetic test data]{
    \includegraphics[scale=0.33]{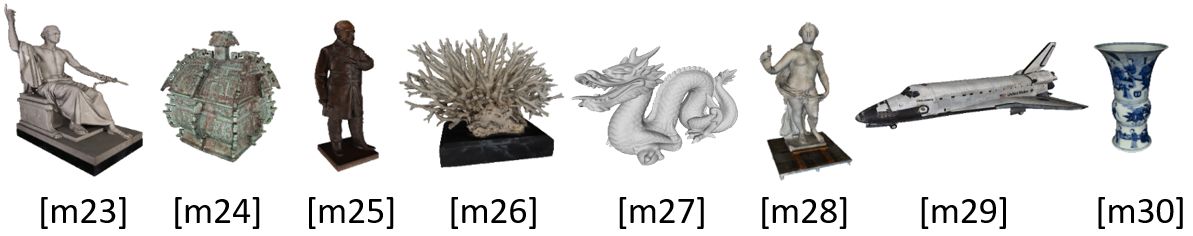}}\qquad
    \subfloat[Models of real test data]{
    \includegraphics[scale=0.33]{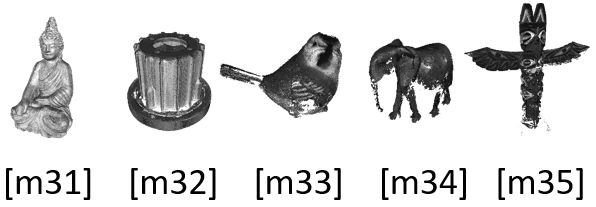}}
    \caption{3D models that have been used to create the synthetic and real datasets. (a) shows the models on which the synthetic training scenes are based on, (b) shows the models of the synthetic test scenes (c) shows the models, that result from the captured real data.
    }
    \label{DatasetsModels}
    \vspace{0.5cm}
\end{figure*}

\section{Data Sets and Data-Processing}
There are already a number of public datasets, as well for optical flow estimation (\textit{Flying Chairs}, \textit{Sintel}, \textit{Kitti}, \textit{Flying Things3D}) as for pose estimation and odometry (\textit{Kitty Odomety}, \textit{3D Match}, \textit{ModelNet14}, \textit{ShapeNet}). 
Unfortunately, only datasets that provide both images and depth data are suitable for the proposed investigations.
Given the depth map and the camera calibration, the required normal maps can be approximated by practical methods, such as \cite{holzer2012adaptive} and are thus no prerequisite.
Therefore, for the evaluation of the estimated flow fields and the inferred poses in comparison to state-of-the-art techniques, the established \textit{Kitty Odomety} dataset will be used later on.
Even if it does not reflect the main application area for the development of the method, since it involves quite small rotations that barely show shading differences due to movement of the camera instead of the scene, it allows comparison with previously existing methods.

\begin{figure*}[t]
    \centering
    \subfloat[Rendered images]{
    \includegraphics[height=0.115\textwidth]{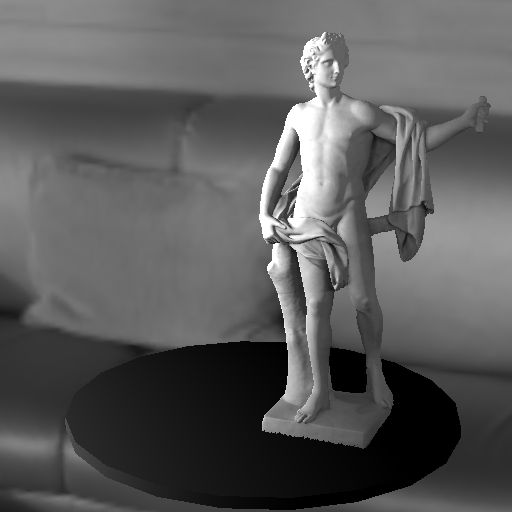}
    \includegraphics[height=0.115\textwidth]{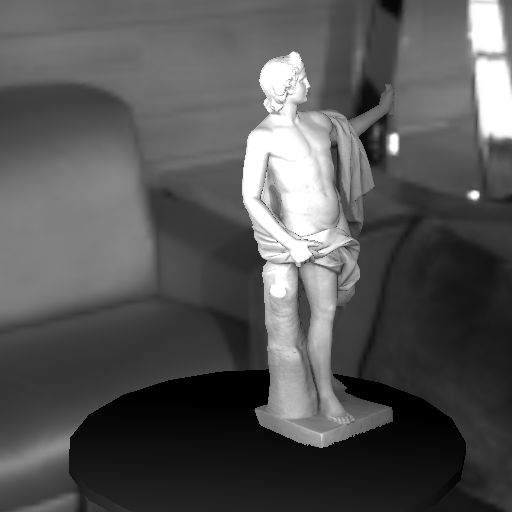}}\ 
    \subfloat[Rendered depth maps]{
    \includegraphics[height=0.115\textwidth]{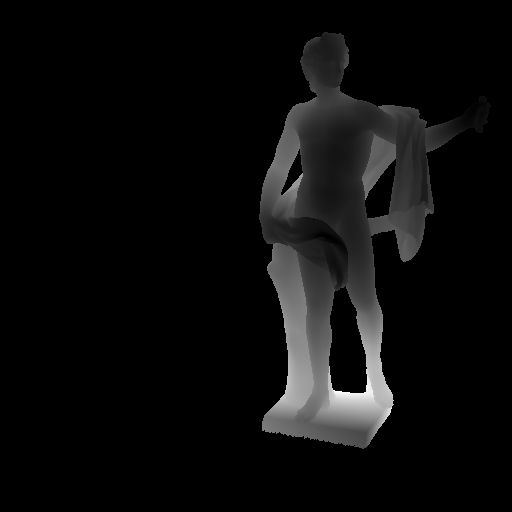}
    \includegraphics[height=0.115\textwidth]{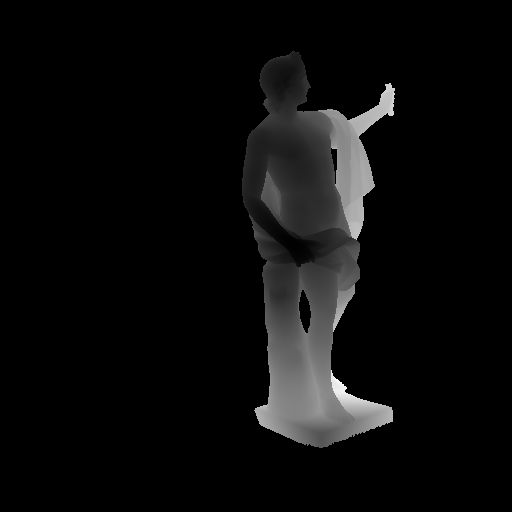}}\ 
    \subfloat[Rendered normal maps]{
    \includegraphics[height=0.115\textwidth]{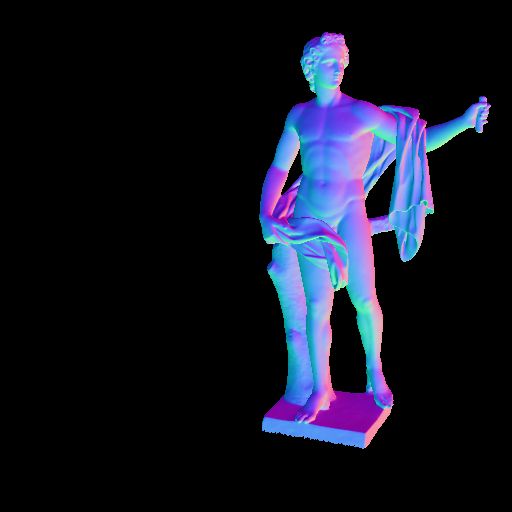}
    \includegraphics[height=0.115\textwidth]{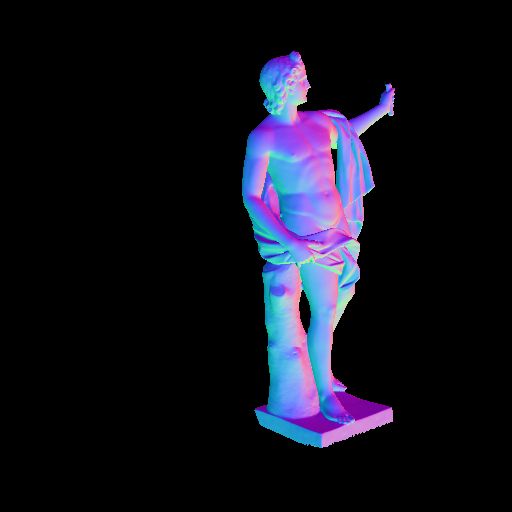}}\ 
    \subfloat[Rendered flow fields]{
    \includegraphics[height=0.115\textwidth]{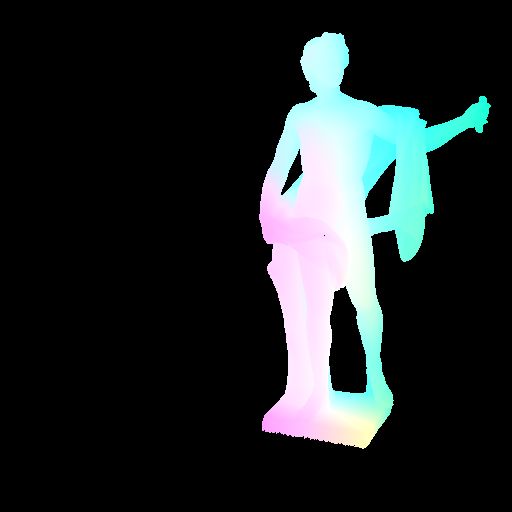}
    \includegraphics[height=0.115\textwidth]{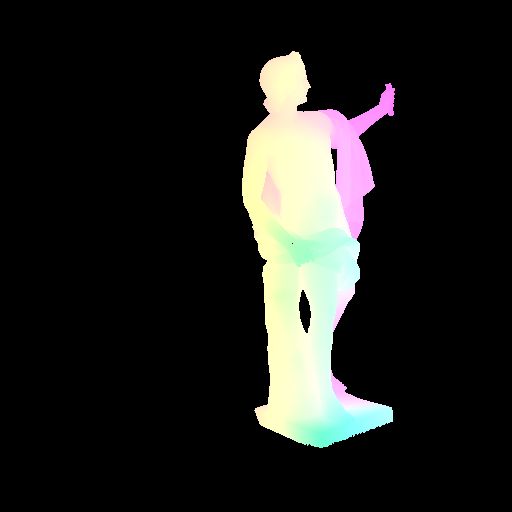}}\\
    \subfloat[Captured images]{
    \includegraphics[height=0.115\textwidth]{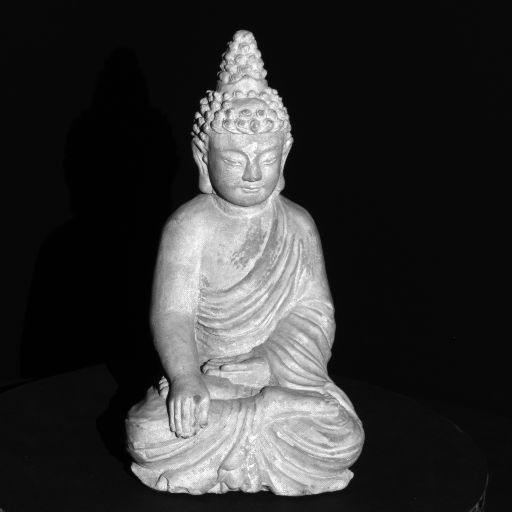}
    \includegraphics[height=0.115\textwidth]{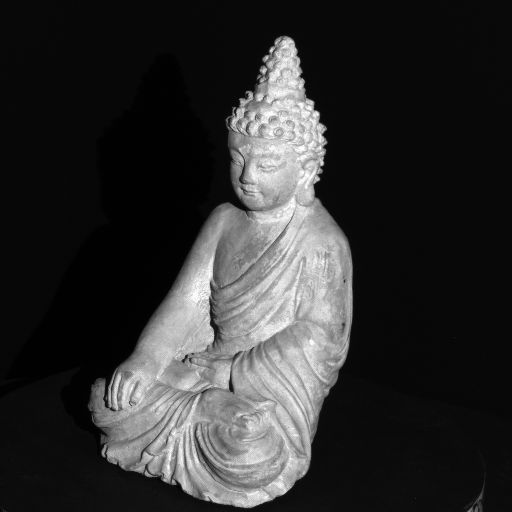}}\ 
    \subfloat[Reconstructed depth maps]{
    \includegraphics[height=0.115\textwidth]{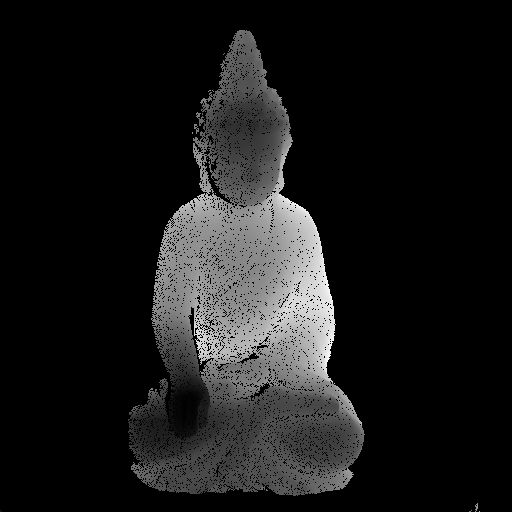}
    \includegraphics[height=0.115\textwidth]{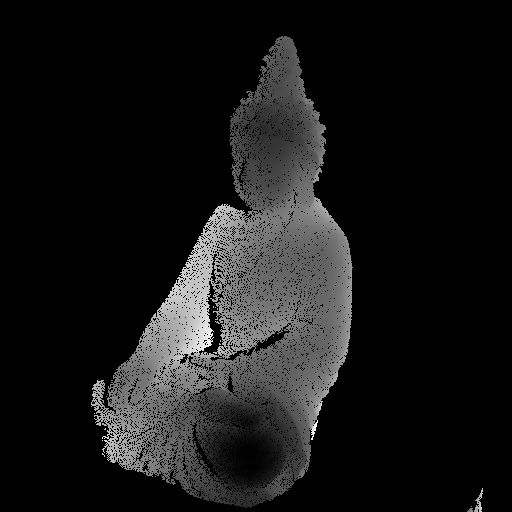}}\ 
    \subfloat[Computed normal maps]{
    \includegraphics[height=0.115\textwidth]{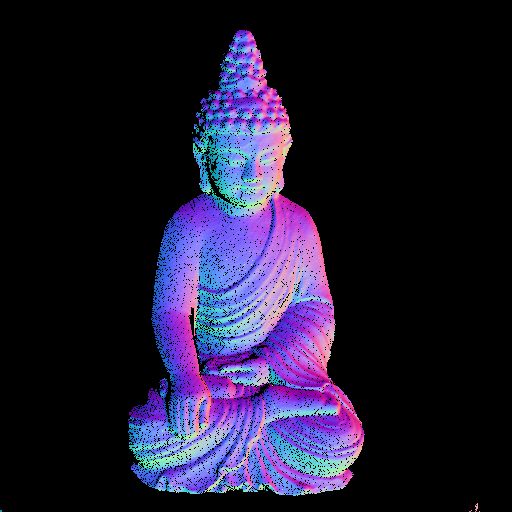}
    \includegraphics[height=0.115\textwidth]{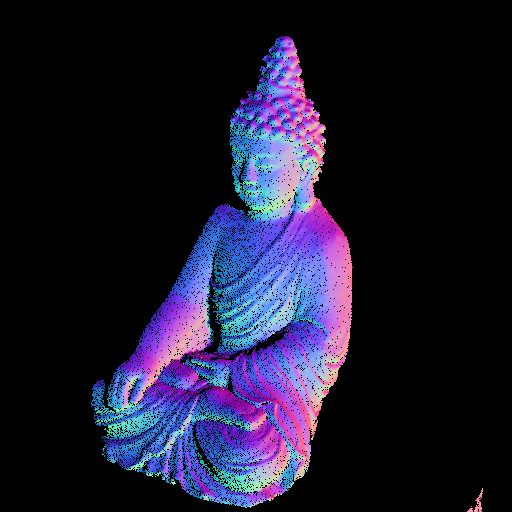}}\ 
    \subfloat[Computed flow fields]{
    \includegraphics[height=0.115\textwidth]{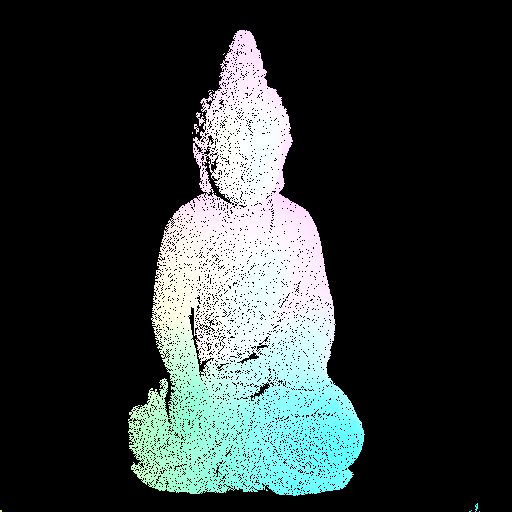}
    \includegraphics[height=0.115\textwidth]{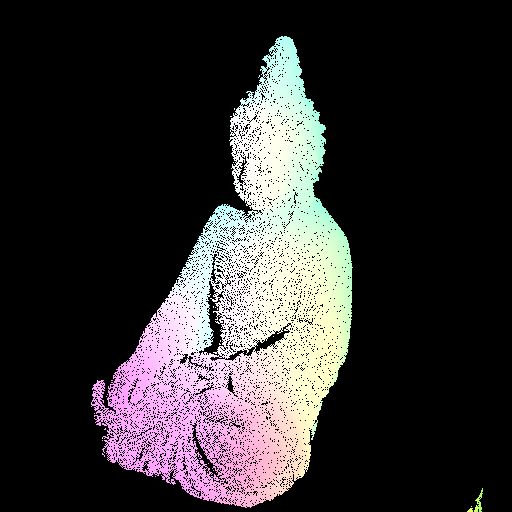}}
    \caption{Example scene of the synthetic (top row) and real (bottom row) datasets. 
    Each scene contains images, depth maps, normal maps and flow fields of two different camera views.
    In addition a data file for each camera is stored, that contains calibration information, camera position, light source position and minimal/maximal values of flows and depths in order to allow memory efficient saving of the data.\vspace{-0.2cm}
    }
    \label{DatasetsExamples}
\end{figure*}

Nevertheless, for the task of rotating objects, ground truth data of both, optical flow and scene pose are required for training the presented network.
Also, it is advantageous to be able to use absolutely correct normal, depth and calibration data to avoid the influence of computational errors on the training.
To the best of our knowledge, no such dataset exists. 
Also, a general dataset for object orientation in the context of 3D reconstruction is not available to our knowledge.
Therefore, several datasets are published together with this publication (\href{https://www.dfki.uni-kl.de/~fetzer/flow2pose.html}{https://www.dfki.uni-kl.de/$\sim$fetzer/flow2pose.html}). 
Among them are two synthetic datasets with rendered images, normals, depth maps and ground truths of camera calibration, optical flow and camera positions.
One of them contains scenes with consistent scene illumination (\textit{ConsistentLight}) of both camera views.
The other one contains scenes with inconsistent illumination (\textit{InConsistentLight}), where the position of the light source changes significantly between the views.
This simulates the difficult case, where, for example, the object rotates, which may dramatically change the angle of incidence of the light (violated brightness assumption).
The scenes of the synthetic data sets were created and rendered using \textit{Unity} \cite{Unity}.
To avoid dependencies on the background, 75 spherical backgrounds were added to the scenes randomly.
The gray scale images, depth maps, normal maps and optical flows were rendered for random scenes each from two random camera perspectives. 
The calibration information, the camera positions and the position of the illuminating point light are also provided.
For both synthetic datasets, each a training subset and a test subset were created.
The training sets contain 20.000 random scenes in which objects were randomly placed in the scene.
The test sets contain 1.000 random scenes in which other objects, that have not been used in the training sets were chosen.
The 22 models used for the training sets are shown in Figure \ref{DatasetsModels} (a) and the 8 models used for the test sets are shown in \ref{DatasetsModels} (b).
Figure \ref{DatasetsExamples} (a), (b), (c) and (d) shows the rendered data for an exemplary scene.

In a similar format, a real dataset (\textit{BuddhaBirdRealData}) is delivered, which consists of captured data from 5 different objects, shown in Figure \ref{DatasetsModels} (c).
The images are captured by monochrome cameras.
The depth data has been reconstructed by a structured light approach.
The normal maps are computed from the geometry, defined by the depth data and the calibration information.
After manually aligning the data, the semi-dense flow fields have been computed and stored.
The scenes were illuminated by a projector, that has been calibrated jointly with the cameras and thus also delivers the light position in the scenes.
Each model has been captured from 8 positions with two different cameras each.
Flow and pose data is available for each of the camera combinations of ancient positions, which yields ground truth data for 40 combinations per object.
This results in 200 ground truth scenes of the real data, that can be used for testing the models in real scenarios.
Thereby, the first 40 pairs represent the scans within one scan head (consistent light) with 8 reconstructions per object. 
The last 160 pairs represent the inconsistent light case with 
combinations of camera views between adjacent scans (that use different projectors).
Similar to the synthetic case Figure \ref{DatasetsExamples} (e), (f), (g), (h) shows the captured and estimated data for an exemplary real scene. 

\subsection{Data Sources and Data Formats}
The 3D Models that have been used to create the data sets are taken from different sources and free to use.
Models [m9, m12, m27] were taken from the Stanford 3D Scanning Repository \cite{StanfordScanRep}.
Models [m2, m7, m8, m11, m20] were taken from \cite{zhou2005texturemontage}.
Models [m1, m3, m5, m6, m10, m13, m14, m17, m18, m21, m23, m24, m25, m26, m29, m30] were taken from the Smithsonian 3D Digitization page \cite{Smithsonian} that collected a large amount of 3D data from several museums and archives, from which many are free to use.
Models [m4, m15, m16, m19, m22, m28, m31, m32, m33, m34, m35] resulted from own research and are released with this work.

\noindent Each scene of the datasets, no matter if real or synthetic, consists of the following data parts:
\begin{itemize}
    \item \texttt{image0} and \texttt{image1} contain the 8-bit integer grayscale images of the two camera views.
    \item \texttt{data0} and \texttt{data1} are .json files that contain the intrinsic calibration matrices $\textbf{K}$, camera rotation $\textbf{R}$ and translation $\textbf{t}$, the minimal and maximal depth values $minDepth$ and $maxDepth$, the minimal and maximals values of the horizontal and vertical optical flows $minFlowX$, $maxFlowX$, $minFlowY$ and $maxFlowY$ and the coordinates if the light source $lightPos$.
    \item \texttt{depth0} and \texttt{depth1} are 16-bit integer grayscale images that need to be scaled after loading using minimal and maximal depth values from the data files:
    \begin{align*}
        D = D\cdot\frac{maxDepth-minDepth}{65535}+minDepth
    \end{align*}
    \item \texttt{normal0} and \texttt{normal1} are 24-bit integer RGB images in tangent space that can be re-transformed to spatial space by:
    \begin{align*}
        \textbf{n}=(\frac{2\textbf{n}_1}{255}-1,\         \frac{2\textbf{n}_2}{255}-1,\ 1-\frac{2\textbf{n}_3}{255})
    \end{align*}
    \item \texttt{flow0} and \texttt{flow1} contain the horizontal and vertical displacements of the respective flow fields between the views. The flows are stored as 16-bit integers in three channel images (\textit{flowX, flowY, zeros}) and are scaled similar to the depth files.
\end{itemize}
Note that missing/masked pixels for which no depth information is available contain zeros in the depth, flow and normal files.
After rescaling and shifting these files, the mask should be applied again to keep the masking information with values of zero.

The presented network uses vertex maps instead of depth maps.
These can be computed from the depth data and the given calibration information by applying the following operation to each image pixel $(x,y)$:
\begin{align}
    V(x,y) = \frac{\textbf{K}^{-1}(x\ y\ 1)^\mathsf{T}}{\|\textbf{K}^{-1}(x\ y\ 1)^\mathsf{T}\|_2}\cdot D(x,y)
    \label{Depth2Vertex}
\end{align}
\subsection{Camera Pose and Scene Pose}
The given depth, vertex and normal maps are independent of any camera pose, as these are usually not available beforehand and need to be computed by the procedure.
In order to use them to triangulate point clouds with respect to the given pose, the vertex maps (or point clouds) and normal maps can be transformed in the following way.
Given a camera pose $P=(\textbf{R}$, $\textbf{t})$, the 3D point with respect to a complete camera matrix $\textbf{P}=\textbf{K}[\textbf{R}|\textbf{t}]$ is given by:
\begin{align}
    V(x,y) = -\textbf{R}^\mathsf{T}\textbf{t} +\textbf{R}^\mathsf{T} V(x,y)
\end{align}
and the normals of the respective 3D points are given by:
\begin{align}
    N(x,y) = \textbf{R}^\mathsf{T} N(x,y)
\end{align}
For completeness, we remind, that the camera itself is located in $\textbf{R}^\mathsf{T}\textbf{t}$.

In the usual case of unknown camera poses, only the relative transformation between two vertex maps / point clouds can be estimated from the given data by a procedure as introduced in the previous section.
In order to use the provided data, to deliver relative ground truth transformations between two views, the absolute poses need to be transferred to relative ones.
If we are given the camera extrinsics of two views $\textbf{R}_0$, $\textbf{t}_0$ and $\textbf{R}_1$, $\textbf{t}_1$, the relative pose between vertex map $V_0$ and vertex map $V_1$ is given by
\begin{align}
    \textbf{R}_{01}=\textbf{R}_1\textbf{R}_0^\mathsf{T},\quad \textbf{t}_{01}=\textbf{t}_1-\textbf{R}_1\textbf{R}_0^\mathsf{T}\textbf{t}_0
\end{align}
where vertex map $V_0$ is mapped to vertex map $V_1$ by applying the transformation as:
\begin{align}
    V_1=\textbf{R}_{01}V_0+\textbf{t}_{01}\ .
\end{align}
Example code on how to read, transform and also visualize the given data can be found together with the datasets (\href{https://www.dfki.uni-kl.de/~fetzer/flow2pose.html}{https://www.dfki.uni-kl.de/$\sim$fetzer/flow2pose.html}).

\subsection{Pre- and Post-Processing of Data}
Point clouds that need to be aligned may theoretically be of arbitrary scale.
Neural network based approaches, like the presented one, need to extract meaningful features within the given vertex maps to find corresponding points from which the desired transformation can be predicted.
For this purpose, the network is adapted to the specific task with fixed weights that have been optimally determined during a data based training.
For different absolute values of the 3D positions it is not possible to extract meaningful features within the vertex maps with always the same weights.
Especially, learned thresholds for activations within the network may not be applicable.

A practical way around is to scale and move the point clouds, or equivalently the 3D data in the vertex maps, approximately towards the unit cube, which is located at the world origin.
Within this working volume, the neural network can work effectively and perform the alignment. 
The calculated pose is then combined with the previous transformation towards the unit cube and thus provides the desired operation on the raw data.

In a first step, the point clouds are moved to the origin by subtracting the centroids.
In a second step the point clouds are scaled to fit approximately into the unit cube.
Note, that the method presented assumes the point clouds to be of similar scale, as it appears from usual depth data coming up from the same sensor. 
Therefore, the scaling factor $s$ towards the unit cube should also be chosen similar for both point clouds that are processed.

Let be given the two point clouds $X_0=\{x^{(0)}_1,...,x^{(0)}_M\}$ and $X_1=\{x^{(1)}_1,...,x^{(1)}_N\}$ that need to be aligned.
The centered point clouds at the origin are given by:
\begin{align}
\begin{split}
    X_0-\mu_0=\{x^{(0)}_n-\mu_0\ |\ x^{(0)}_n\in X_0\},\ \mu_0=\sum_{m=1}^Mx^{(1)}_m\\
    X_1-\mu_1=\{x^{(1)}_m-\mu_1\ |\ x^{(1)}_m\in X_1\},\ \mu_1=\sum_{n=1}^Nx^{(1)}_n
\end{split}
\end{align}
$X_0-\mu_0$ and $X_1-\mu_1$ are then scaled jointly and robustly in order to ensure that 90\% of the point clouds map into the according subspace of the unit cube ($[-0.45,0.45]^3\subset\mathbb{R}^3$), that is located at the origin.
This robustifies the scaling and reduces the negative effect of outliers dramatically.
Note, that in general it can be assumed that at least 90\% of a point cloud should contain usable data.
Let be given the set of values with maximal absolute coordinates of both centered point sets, $Y = \{\max(|x|)\ |\ x\in(X_0-\mu_0)\cup(X_1-\mu_1)\}.$\\
Having sorted the values $y_n\in Y$ in ascending order ${y_1\le...\le y_{M+N}}$, the scaling factor, that ensures 90\% of both point clouds being mapped into the cube, defined above is given by $s=1/y_{\lfloor0.45(M+N)\rfloor}$, where $\lfloor\cdot\rfloor$ denotes floor rounding to integer values.
The scaled, centered point clouds, that are ready to be fed to the network, are finally given by:
\begin{align}
    \tilde{X}_0 = s(X_0-\mu_0),\qquad
    \tilde{X}_1 = s(X_1-\mu_1)
\end{align}
Having computed a pose $\tilde{P}=(\tilde{\textbf{R}},\tilde{\textbf{t}})$ using the neural network, that aligns the scaled point clouds by
\begin{align}
    \tilde{\textbf{R}}\tilde{X}_0+\tilde{t}\approx\tilde{X}_1,
\end{align}
the final transformation $P=(\textbf{R},\textbf{t})$, that aligns the raw point clouds $X_0$ and $X_1$ is given by
\begin{align}
    \textbf{R}=\tilde{\textbf{R}},\quad \textbf{t}=\frac{1}{s}\tilde{\textbf{t}}+\mu_1-\tilde{\textbf{R}}\mu_0
\end{align}

\section{Coherent Learning of INV-Flow2PoseNet}
The goal is to train the network to estimate the best possible optical flow that will enable stable extraction of the pose.
Therefore, to get an end-to-end trainable network, we define a joint loss function that penalizes both the ground truth flow and the extracted pose under given flow. 

The \textit{PWC-Net} structure predicts flows $F^{(l)}$ of different levels $l=0,...,L$.

The \textit{Flow2PoseNet} moreover uses these flows in order to predict the relative rotation $\textbf{R}$ and translation $\textbf{t}$.
Let according ground truth be given by $F^{(l)}_\text{GT}$, $\textbf{R}_\text{GT}$ and $\textbf{t}_\text{GT}$.

\subsection{Multiscale Endpoint Error}
The multiscale endpoint error (EPE) penalizes the different levels of the flow calculation with different hardness, provided by the respective weighting parameters $\alpha_l$:
\begin{align}
    \mathcal{L}_\text{EPE}(F^{(0)},...,F^{(L)}) = \sum_{l=0}^L\alpha_l\|F^{(l)}-F^{(l)}_\text{GT}\|_\mathcal{F}
\end{align}
with suitable level weights $\alpha_l$, $l=0,...,L$ and Frobenius matrix norm $\|\cdot\|_\mathcal{F}$. 
In case of sparse data the differences inside the norm are masked in order to take the sparsity into account.

Note that the higher levels, which describe the rather coarse flow, are more important than the lower levels, which get the higher levels as input.
However, since the higher levels have a lower resolution, the flow errors in absolute numbers are smaller than those of the lower levels.
As a rule of thumb, because of the pooling between each level, the weighting should be at least halved each time to account for the resolution discrepancy.
The weights that have been used for the proposed network are $\{\alpha_0,...,\alpha_6\}=\{0.001, 0.0025, 0.005, 0.01, 0.02, 0.08, 0.32\}.$

\subsection{Alignment Error}
A measure that treats both rotation and translation together is the well-known alignment error. 
It models the mean Euclidean distance of all point correspondences given by the groundtruth flow:
\begin{align}
    \begin{split}
    \mathcal{L}_\text{AE}(\textbf{R},\textbf{t})=\sum\|\textbf{R}&V_0(x,y)+\textbf{t}\\&-V_1(x+F^{01}_x,y+F^{01}_y)\|_\mathcal{F}
    \end{split}
\end{align}
This measure best describes the problem to be solved. 
It has the advantage that it weights the impact of rotation against the translation.
Note that it is important to mask errors that contain invalid pixels either of $V_0$ or of warped $V_1$, in order to ensure that only locations are taken into account, where matching vertices in both views are available.

Note that this error alone might erroneously interchange rotations and translation effects in order to receive a minimal alignment error. 
These interchanges can be prevented by adding some direct translational and rotational error terms to the overall loss function. 
These additional terms act as a regularization to enforce a better decomposition into translation and rotation.

\subsection{Translational and Rotational Errors}
The error of the predicted translation is given directly as the Eucliden distance towards the ground truth translation:
\begin{align}
    \mathcal{L}_\text{TRANS}(\textbf{t}) = \|\textbf{t}-\textbf{t}_\text{GT}\|_2
\end{align}
Special attention is required for the rotation error.
A suitable differentiable error between two rotation matrices $\textbf{R}$ and $\textbf{R}_\text{GT}$ is given by the angular error, which is defined by the absolute value of the rotation angle $\theta$ of the relative rotation $\textbf{R}_\text{rel} = \textbf{R}\textbf{R}^\mathsf{T}_\text{GT}$.
Having a look at the conversion towards the axis angle representation there are basically two ways to compute the rotation angle.
The first relation is given with the trace of the rotation matrix:
\begin{align}
    \text{Tr}(\textbf{R}_\text{rel})= 1 + 2\cos(\theta)
    \label{RotAngle1}
\end{align}
Another way is to calculate the rotation angle from the length of the extracted rotation axis.
Having an explicit rotation matrix, the rotation axis $\textbf{u}$ is given by:
\begin{align}
    \textbf{R}_\text{rel}=\begin{pmatrix}a&b&c\\d&e&f\\g&h&i\end{pmatrix}
    \quad\Rightarrow\quad\textbf{u}=\begin{pmatrix}h-f\\c-g\\d-b\end{pmatrix}
\end{align}
The rotation angle $\theta$ is related to the length of $\textbf{u}$ by:
\begin{align}
    \|\textbf{u}\|_2= 2\sin(\theta)
    \label{RotAngle2}
\end{align}
A direct computation of $\theta$ from one of equations (\ref{RotAngle1}) or (\ref{RotAngle2}) 
requires the use of one of the inverse trigonometric functions arcussinus or arcuscosinus.
These yield numeric problems due to singularities in case of angles close to $\pm\frac{\pi}{2}$ or $\pm\pi$, which is unsuitable for a general loss function that has to be differentiable.
A more stable way to achieve $\theta$ is to use the two-dimensional arcustangens atan2 with both arguments:
\begin{align}
    \mathcal{L}_\text{ROT}(\textbf{R}) = |\text{atan2}\big(\|\textbf{u}\|_2, 1-\text{Tr}(\textbf{R}_\text{rel})\big)|
\end{align}

\subsection{Joint Training Loss}
The joint loss function, is subsequently given by:
\begin{align}
\begin{split}
    \mathcal{L} = &\ \mathcal{L}_\text{EPE}(F^{(0)},...,F^{(L)}) + \mathcal{L}_\text{AE}(\textbf{R}_{1 Step}, \textbf{t}_{1 Step})\\
    &+\mathcal{L}_\text{AE}(\textbf{R}_{2 Step}, \textbf{t}_{2 Step})
    +\mathcal{L}_\text{AE}(\textbf{R}_{3 Step}, \textbf{t}_{3 Step})\\
    &+\mathcal{L}_\text{TRANS}(\textbf{t}_{3 Step})+\mathcal{L}_\text{ROT}(\textbf{R}_{3 Step})
    \end{split}
\end{align}
At the beginning of the training, the gradients of the computed optical flow are detached before backpropagating the alignment errors.

\subsection{Representation of Rotation}
In order to ensure the predicted rotation matrix to be a proper rotation, a minimal parameterization by \textit{Euler Angles} is chosen.
Therefore, three values $(\theta, \rho, \phi)$ are predicted by the network, defining the rotation angles around the $x,y$ and $z$ axes by the rotation matrices:
\begin{align}
    \textbf{R}_x = \begin{pmatrix}1&0&0\\0&\cos(\theta)&-\sin(\theta)\\0&\sin(\theta)&\cos(\theta)\end{pmatrix}\\
    \textbf{R}_y = \begin{pmatrix}\cos(\rho)&0&\sin(\rho)\\0&1&0\\-\sin(\rho)&0&\cos(\rho)\end{pmatrix}\\
    \textbf{R}_z = \begin{pmatrix}\cos(\phi)&-\sin(\phi)&0\\\sin(\phi)&\cos(\phi)&0\\0&0&1\end{pmatrix}
\end{align}
The total rotation is given by the consecutive execution of these rotations: $\textbf{R}=\textbf{R}_x\textbf{R}_y\textbf{R}_z$.

Vice versa, the respective \textit{Euler Angles} can be extracted from a given rotation matrix $\textbf{R}$ by:
\begin{align}
    \theta &= \operatorname{atan2}(-\textbf{R}_{23},\textbf{R}_{33})\\
    \rho &= \operatorname{atan2}(\textbf{R}_{13},\sqrt{\textbf{R}_{23}^2+\textbf{R}_{33}^2})\\
    \phi &= \operatorname{atan2}(-\textbf{R}_{12},\textbf{R}_{11})
\end{align}
This conversion is especially used to compute the \textit{Euler Angles} of the refined rotation matrix $\hat{\textbf{R}}$ in the \textit{3 Step Method} of Section \ref{Sec:PoseFromFlow}.

\begin{figure*}[t]
    \includegraphics[width=.99\textwidth]{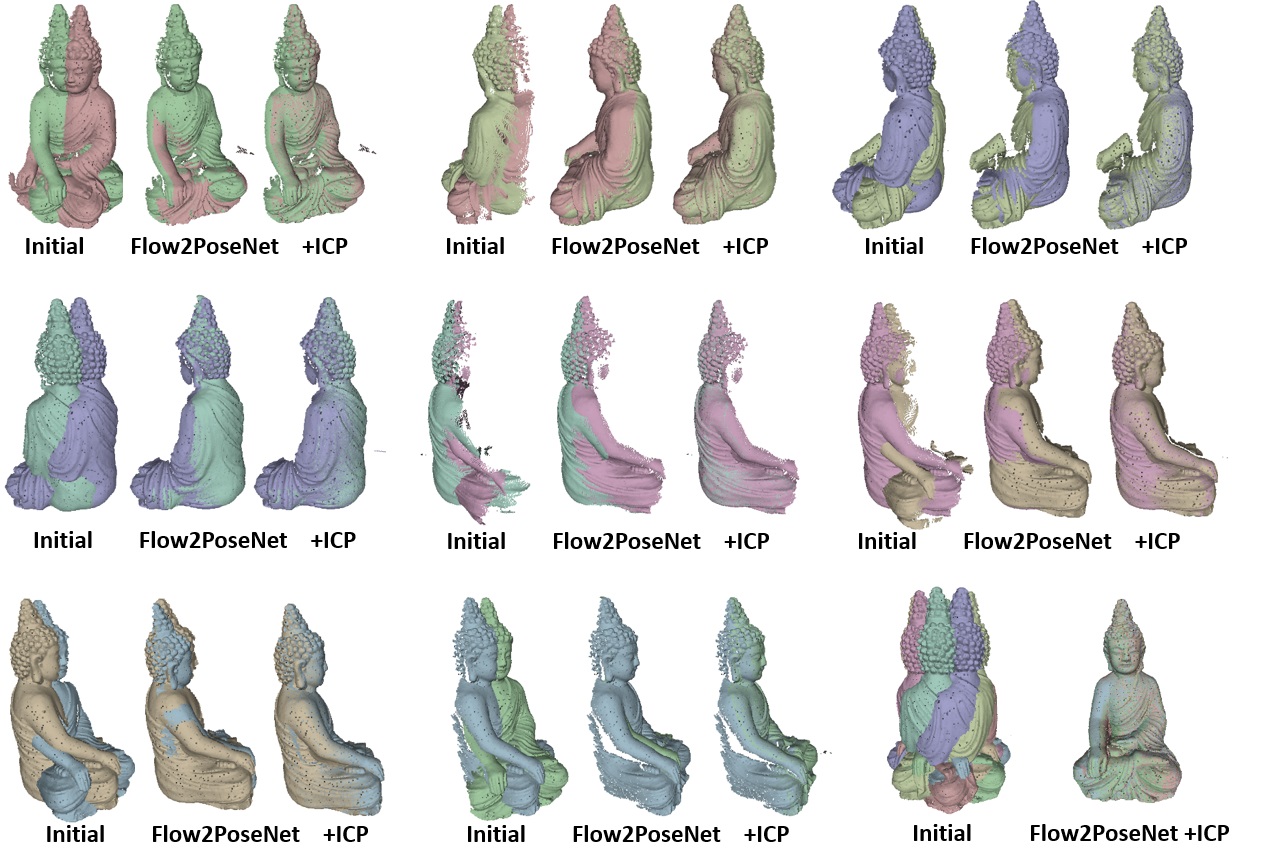}
    \caption{Application of the method to a full sequence of partial reconstructions of a the real Buddha object from the \textit{BuddhaBirdReal} dataset. Such sequences usually result from 3D scanners (as here from a structured light scanner). Since usually a turntable is used, strong rotations ($\approx45^\circ$) and shading changes disturb the data. After the pre-alignment a few iterations of the \textit{ICP} algorithm are applied to refine the alignment of the point clouds. The image on the bottom right shows the impressive result on the overall aligned full point cloud of the statue.\vspace{0.6cm}}
    \label{BuddhaAlignmentSteps}
\end{figure*}
\section{Evaluation}
For evaluation, we compare the calculated optical flows and registrations qualitatively on different synthetic and real data sets.
Highly accurate results visualize a good generalization without finetuning from synthetic training data to the difficult real test scenes.
Figure \ref{Figure:ConsistentLight} and Figure \ref{Figure:InconsistentLight} show the results for exemplary objects from the training (top 3 rows) and test datasets (bottom 3 rows) for the consistent and inconsistent light (moving light source) case. 
Thereby the first columns show the input data consisting of images, normals and depth maps (that are converted to vertex maps using the calibration information, as in Equation \ref{Depth2Vertex}).
The second column shows the resulting optical flow in comparison to the semi-dense ground truth optical flow in column 3.
Columns 4 and 5 finally show the initial and the registered point clouds using the proposed neural networks.
Special attention should be given to row 6 of Figure \ref{Figure:InconsistentLight}, which shows the performance of the neural network on a real test scene without finetuning.

Further positions of the real scene are shown in Figure \ref{BuddhaAlignmentSteps}. 
It visualizes the performance of  the method applied to 8 partial scans of the Buddha scene (from the \textit{BuddhaBirdReal} dataset), as it usually comes up from 3D scanners.
Using the alignment given by the neural network, a few iterations of \textit{Iterative Closest Points (ICP)} for refinement yield impressive results on the overall aligned point cloud of the object.

In particular, for comparability with other methods, we also consider a network trained on the popular training sequences of Kitti Odometry and evaluate it on the test data as shown in Figure \ref{Figure:Kitti}.
As the Kitti dataset has less strong rotations and less shading changes, it is not the typical use case for the proposed method. 
Nevertheless, the proposed method works reliable for this easier kind of situations as well.
\begin{table*}[]
    \centering
    \begin{tabular}{cccccc} 
        & Input & Optical Flow & Ground Truth & Initial Pose & Alignment\\\vspace{0.05cm}\\
        \begin{tabular}{c}\vspace{-2.8cm}\\Synthetic\\Train Data\\\#01003\end{tabular} &
        \includegraphics[height=0.16\textwidth]{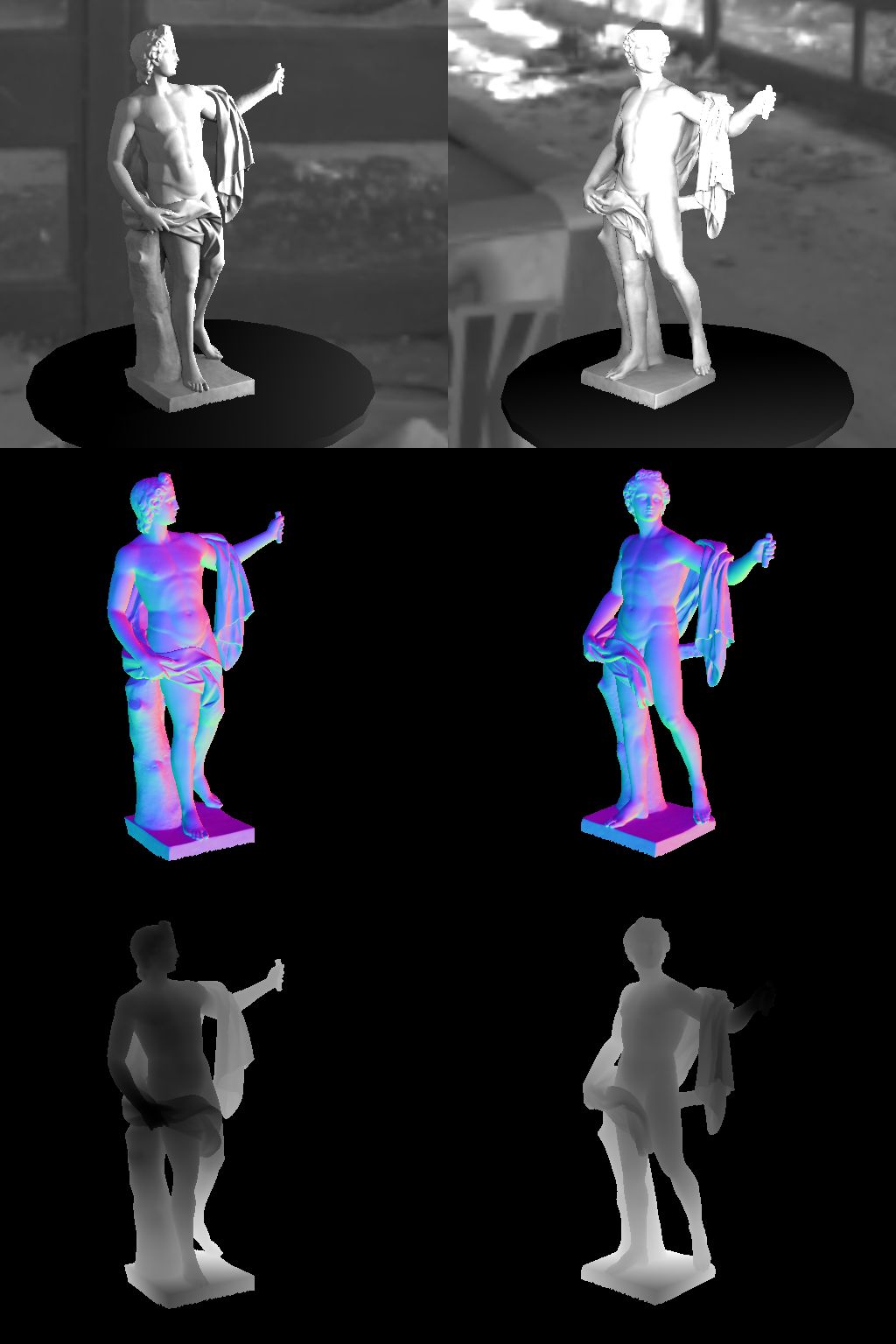} &
        \includegraphics[height=0.16\textwidth]{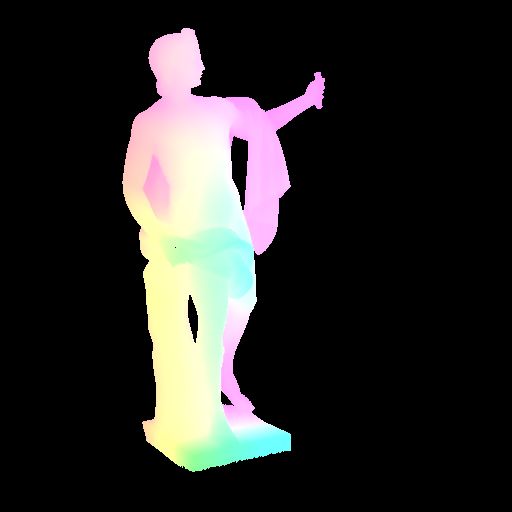} &
        \includegraphics[height=0.16\textwidth]{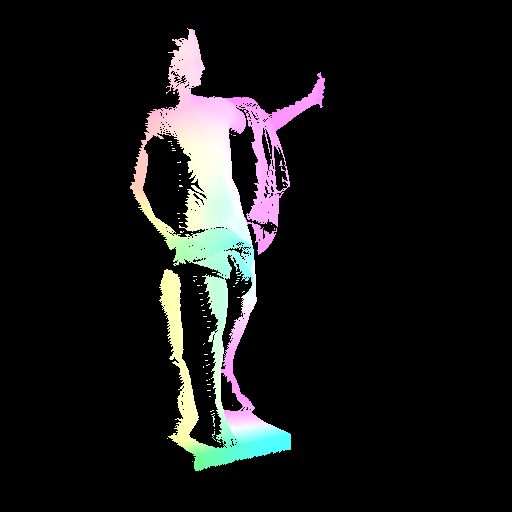} & 
        \includegraphics[height=0.16\textwidth]{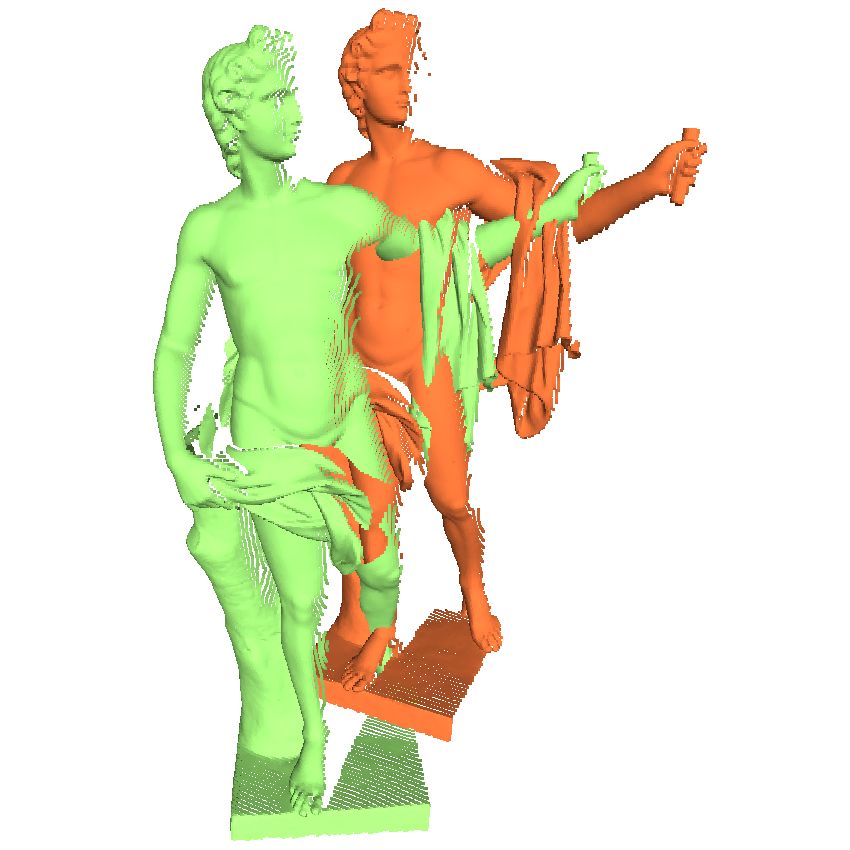} &
        \includegraphics[height=0.16\textwidth]{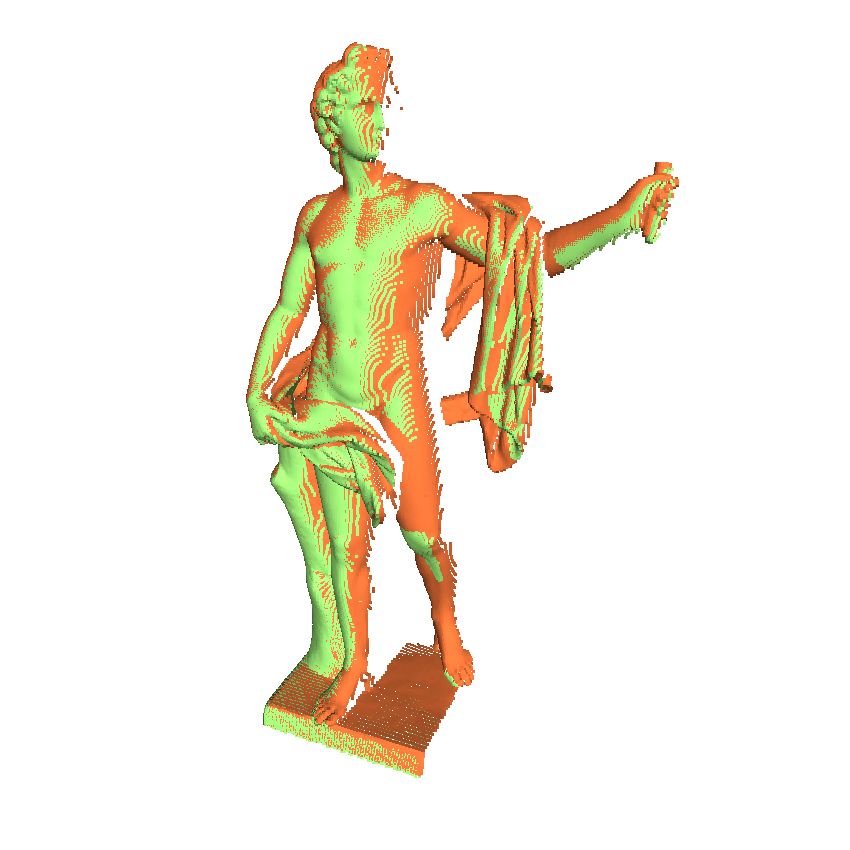}\vspace{0.2cm}\\
        \begin{tabular}{c}\vspace{-2.8cm}\\Synthetic\\Train Data\\\#07285\end{tabular} &
        \includegraphics[height=0.16\textwidth]{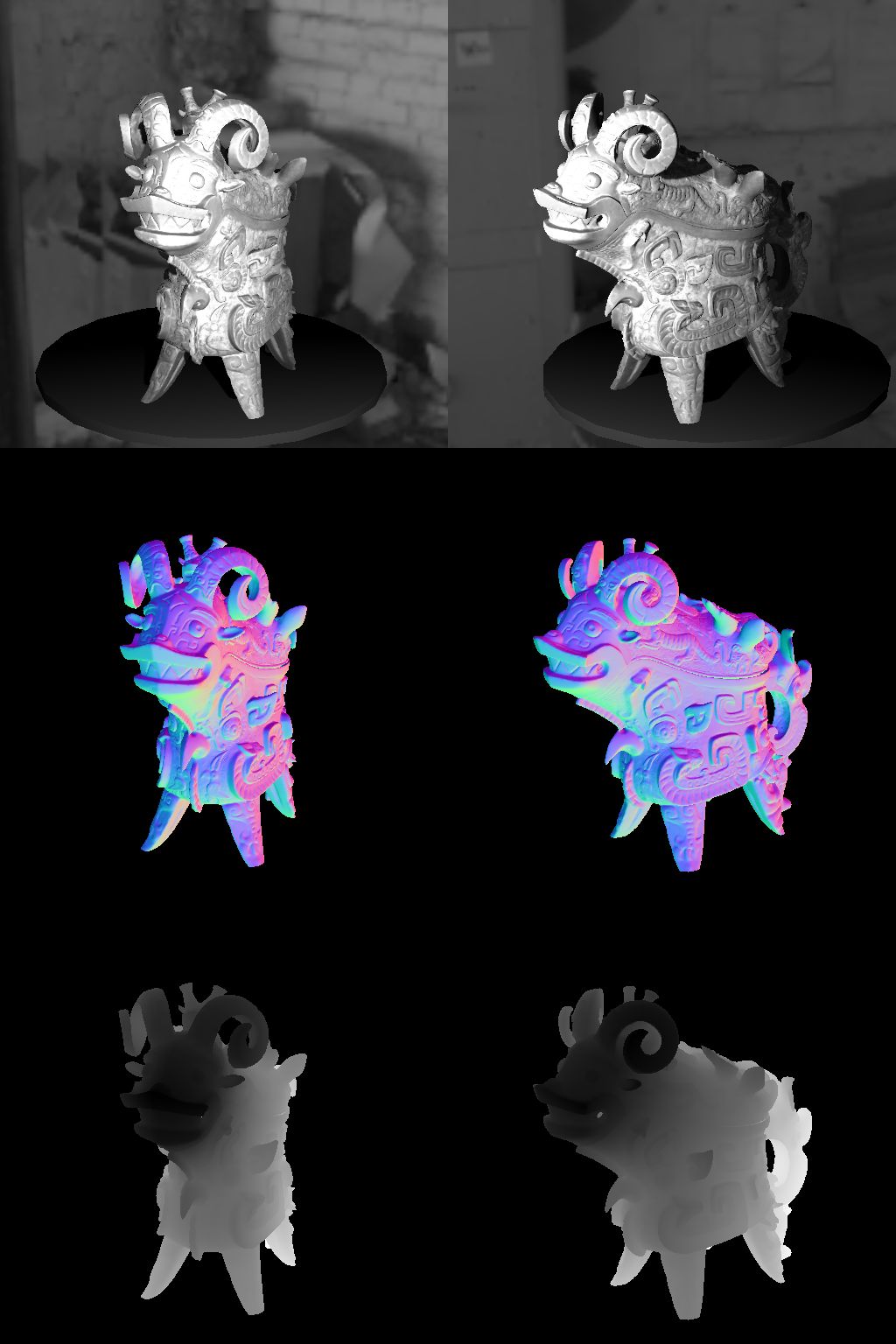} &
        \includegraphics[height=0.16\textwidth]{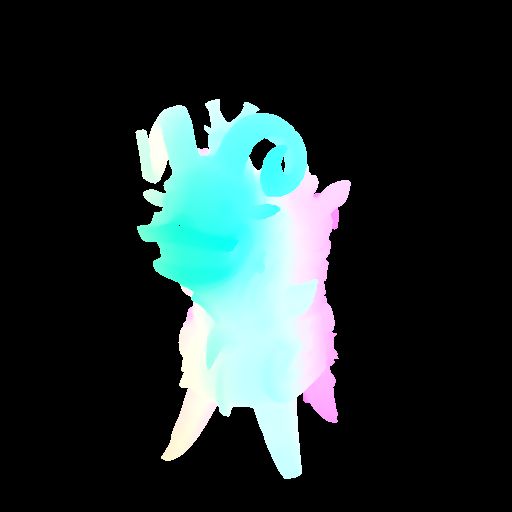} &
        \includegraphics[height=0.16\textwidth]{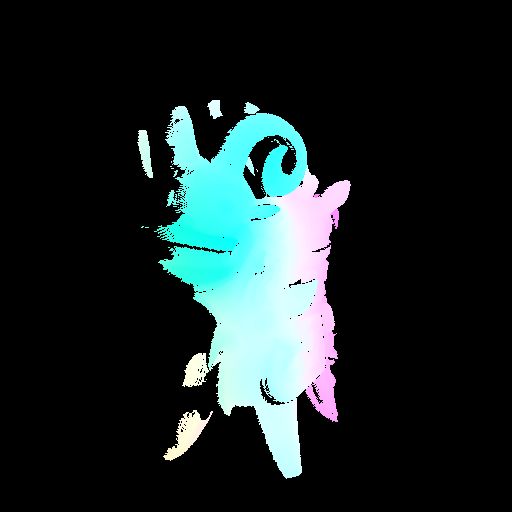} & 
        \includegraphics[height=0.16\textwidth]{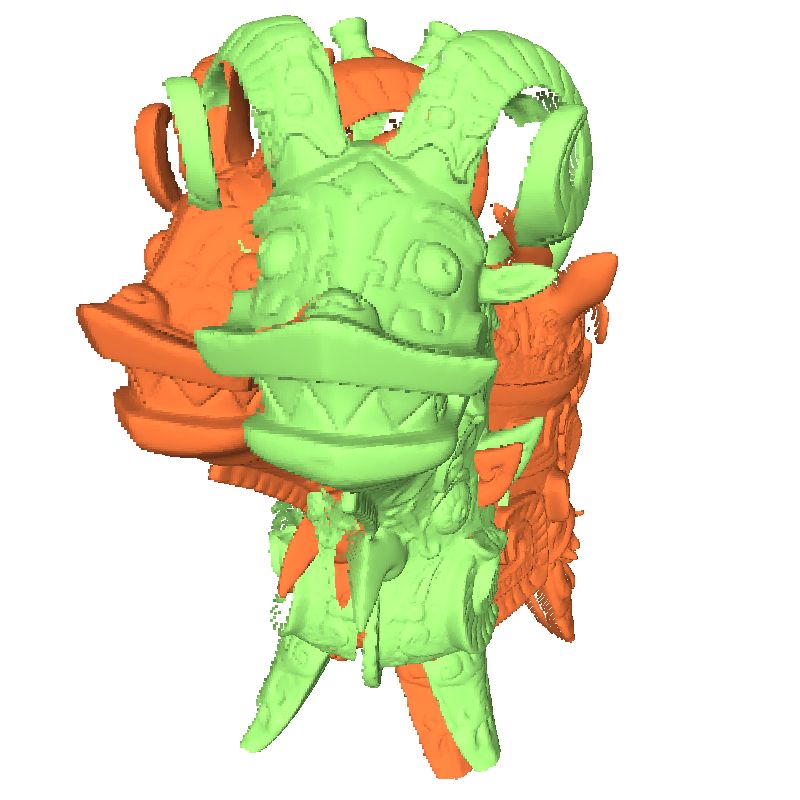} &
        \includegraphics[height=0.16\textwidth]{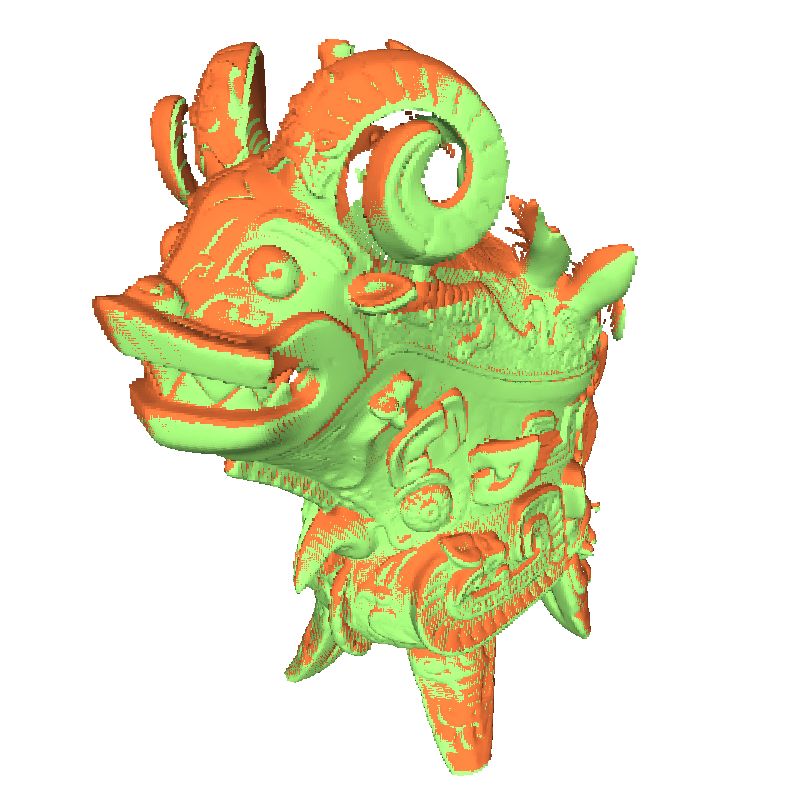}\vspace{0.2cm}\\
        \begin{tabular}{c}\vspace{-2.8cm}\\Synthetic\\Train Data\\\#07236\end{tabular} &
        \includegraphics[height=0.16\textwidth]{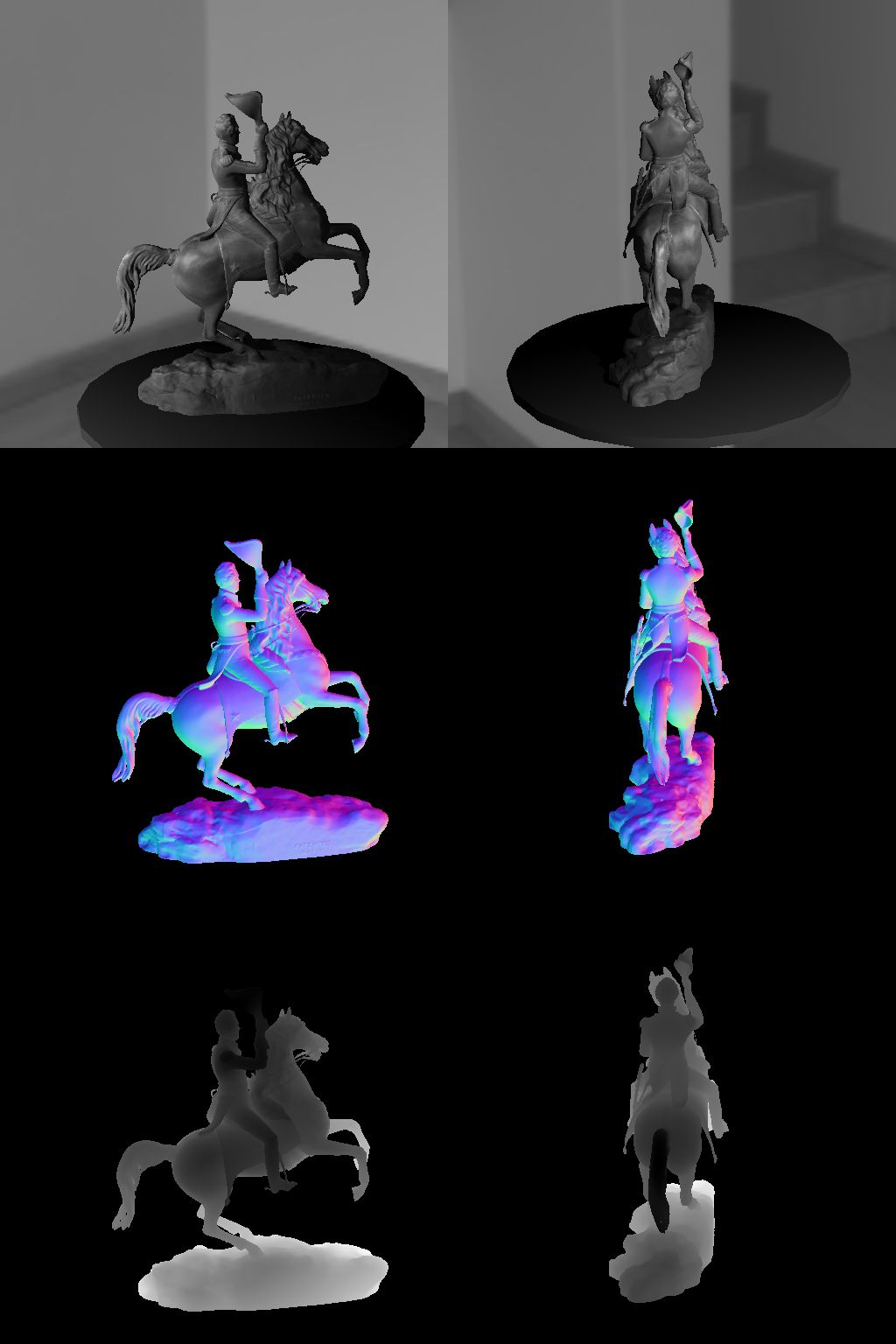} &
        \includegraphics[height=0.16\textwidth]{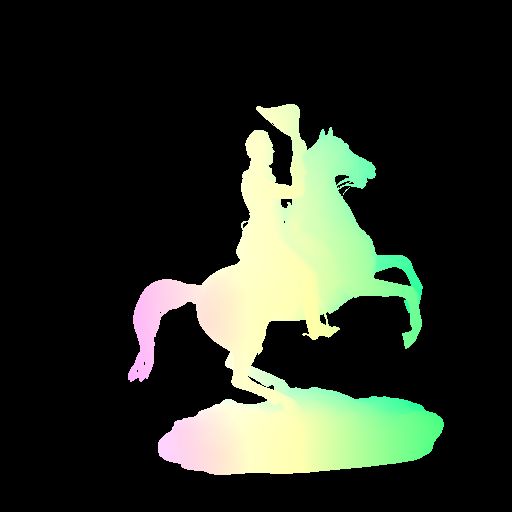} &
        \includegraphics[height=0.16\textwidth]{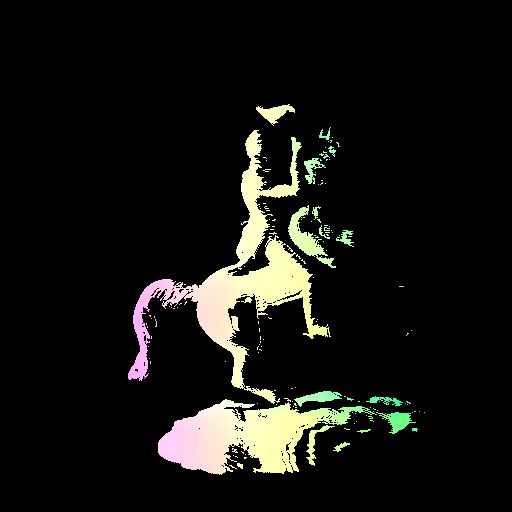} & 
        \includegraphics[height=0.16\textwidth]{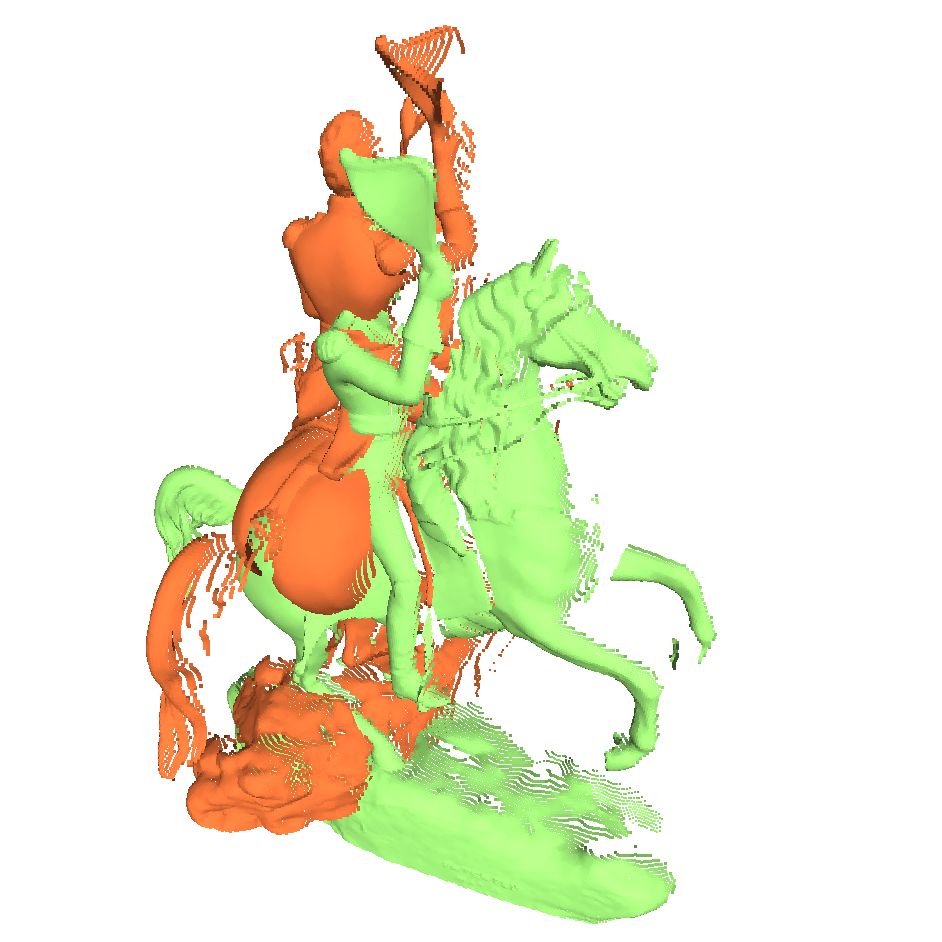} &
        \includegraphics[height=0.16\textwidth]{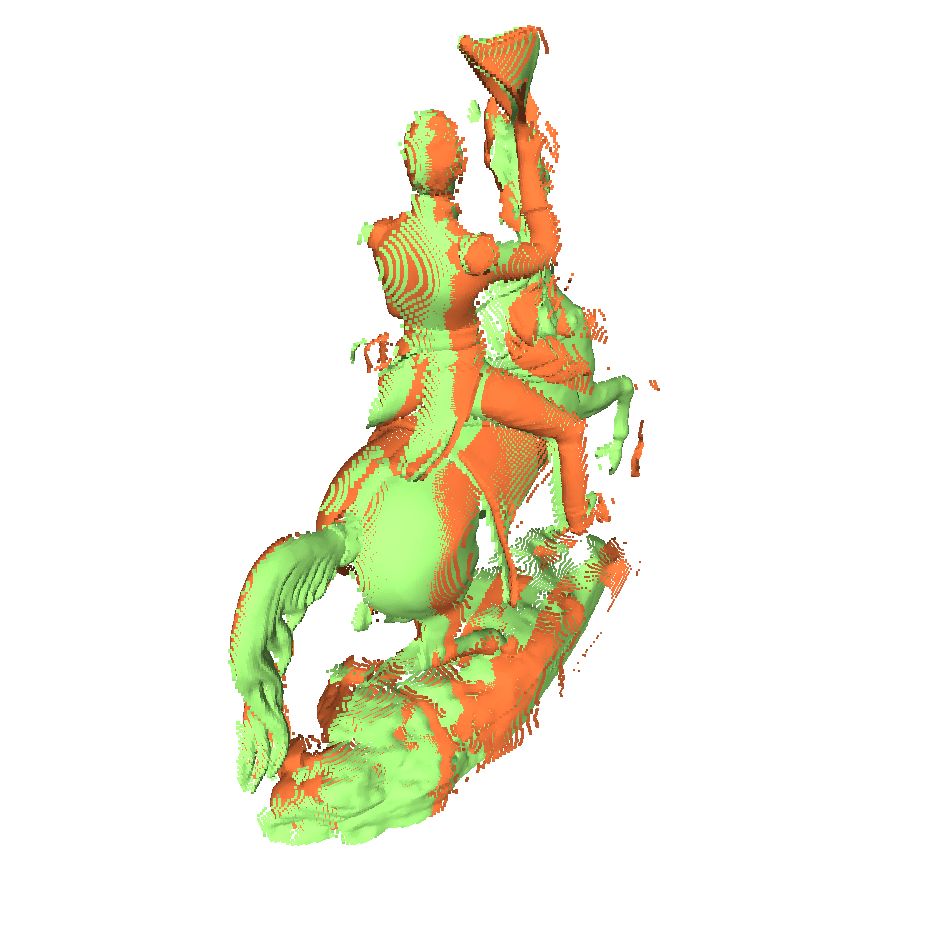}\vspace{1cm}\\
        \begin{tabular}{c}\vspace{-2.8cm}\\Synthetic\\Test Data\\\#00000\end{tabular} &
        \includegraphics[height=0.16\textwidth]{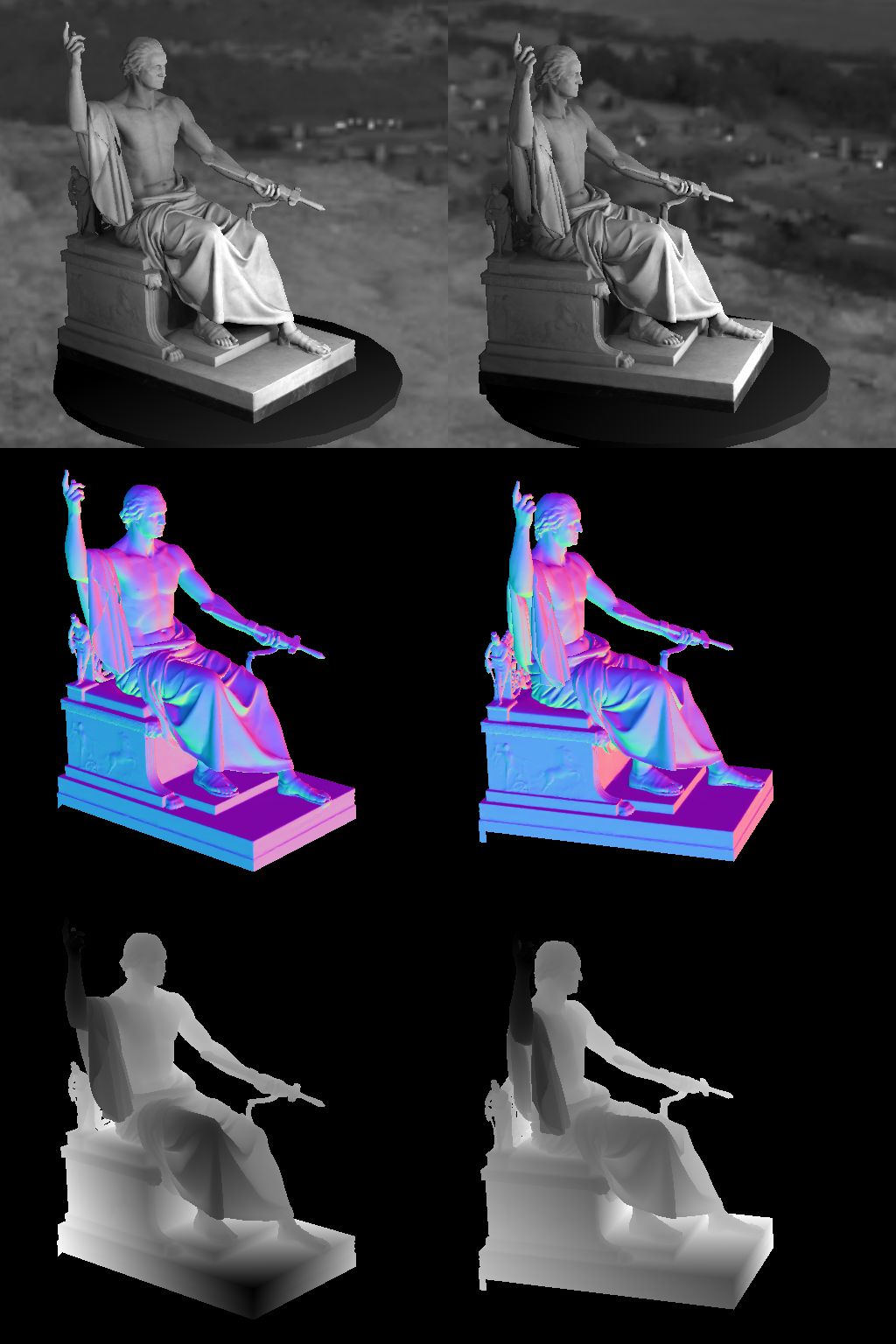} &
        \includegraphics[height=0.16\textwidth]{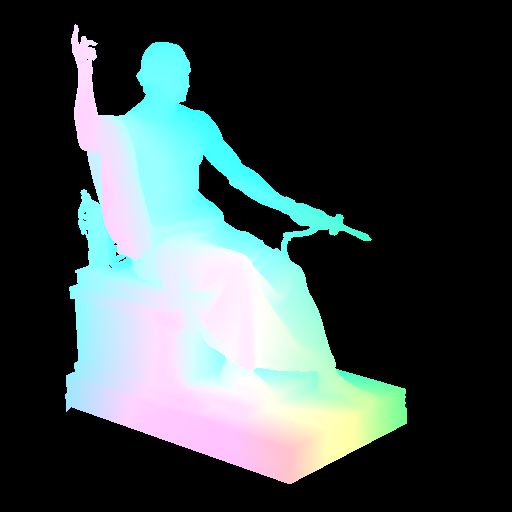} &
        \includegraphics[height=0.16\textwidth]{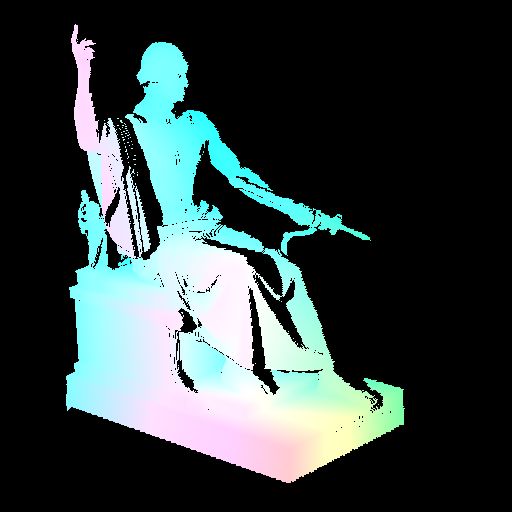} & 
        \includegraphics[height=0.16\textwidth]{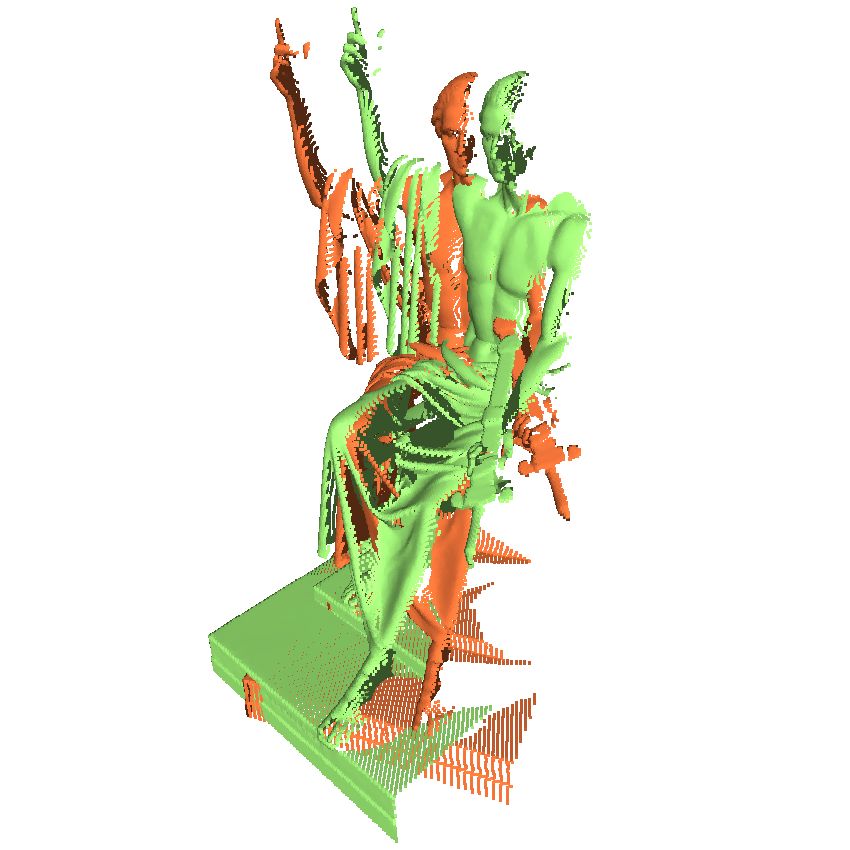} &
        \includegraphics[height=0.16\textwidth]{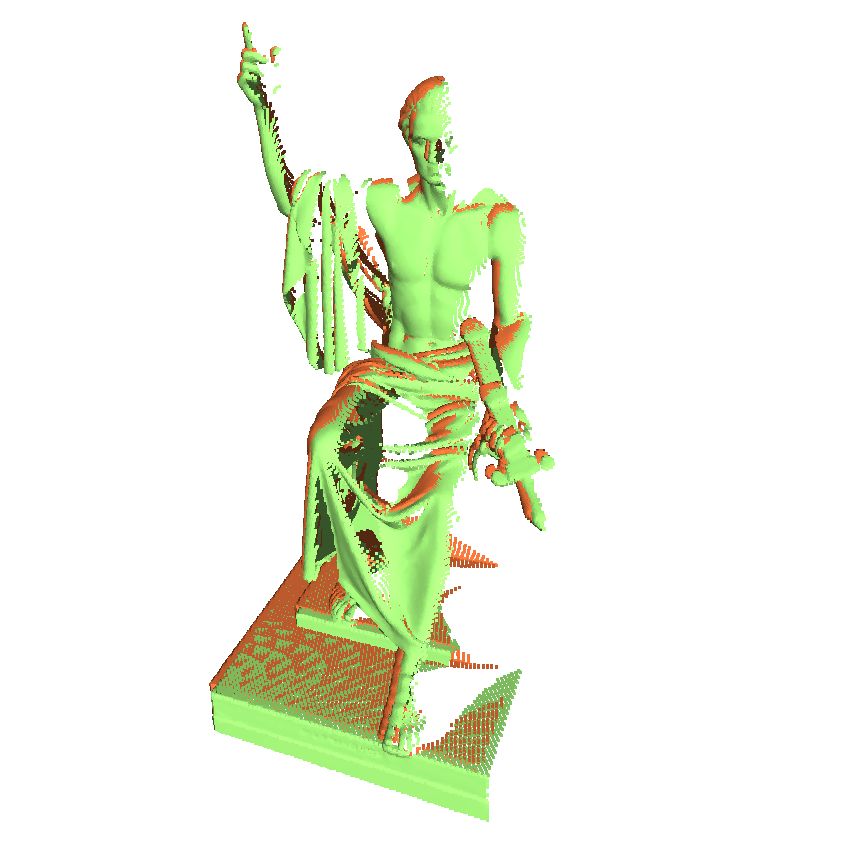}\vspace{0.2cm}\\
        \begin{tabular}{c}\vspace{-2.8cm}\\Synthetic\\Test Data\\\#00107\end{tabular} &
        \includegraphics[height=0.16\textwidth]{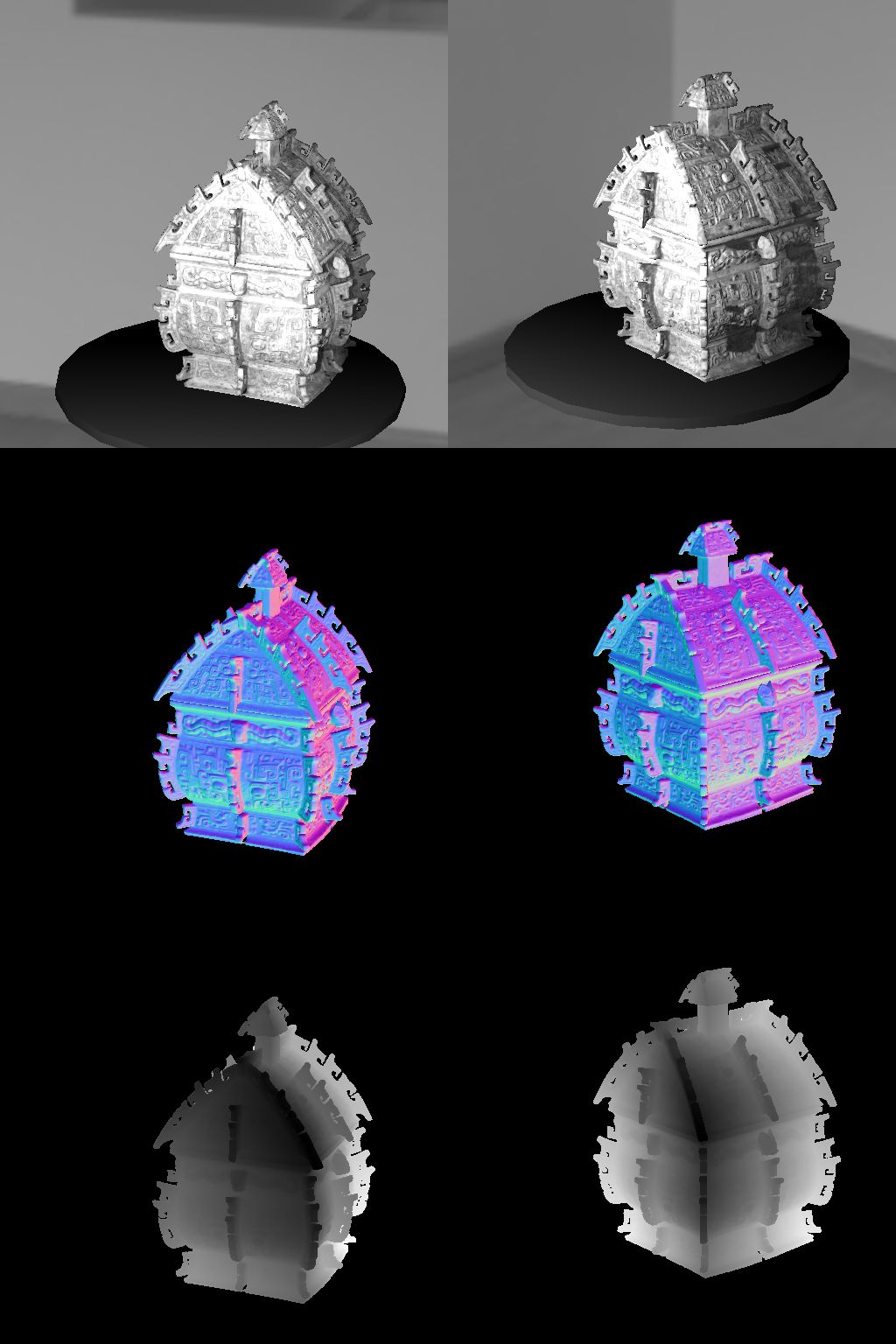} &
        \includegraphics[height=0.16\textwidth]{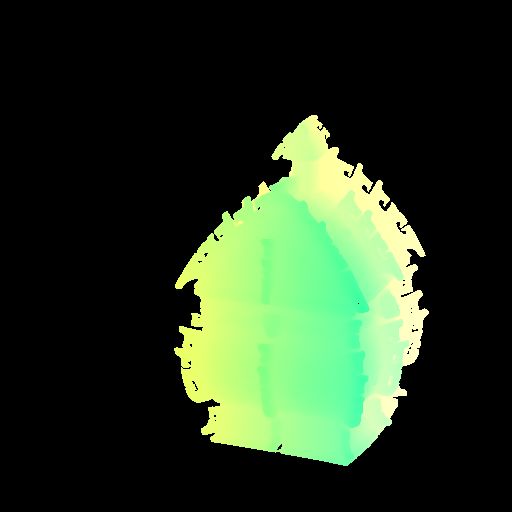} &
        \includegraphics[height=0.16\textwidth]{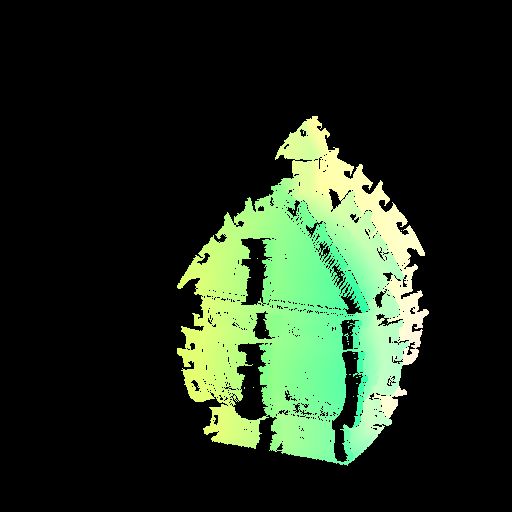} & 
        \includegraphics[height=0.16\textwidth]{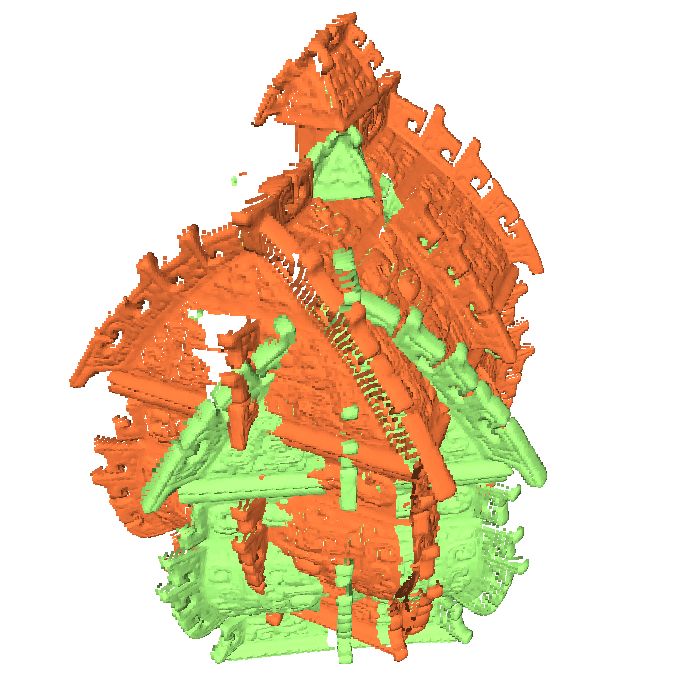} &
        \includegraphics[height=0.16\textwidth]{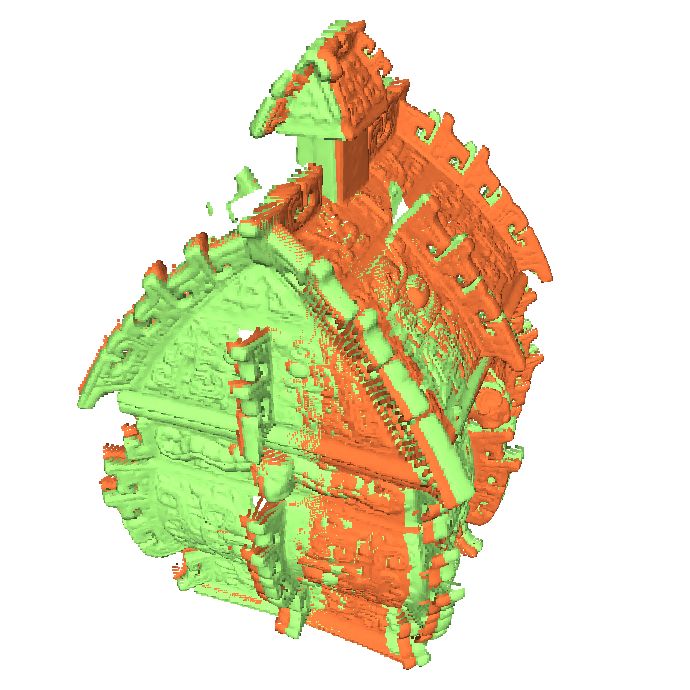}\vspace{0.2cm}\\
        \begin{tabular}{c}\vspace{-2.8cm}\\Synthetic\\Test Data\\\#00153\end{tabular} &
        \includegraphics[height=0.16\textwidth]{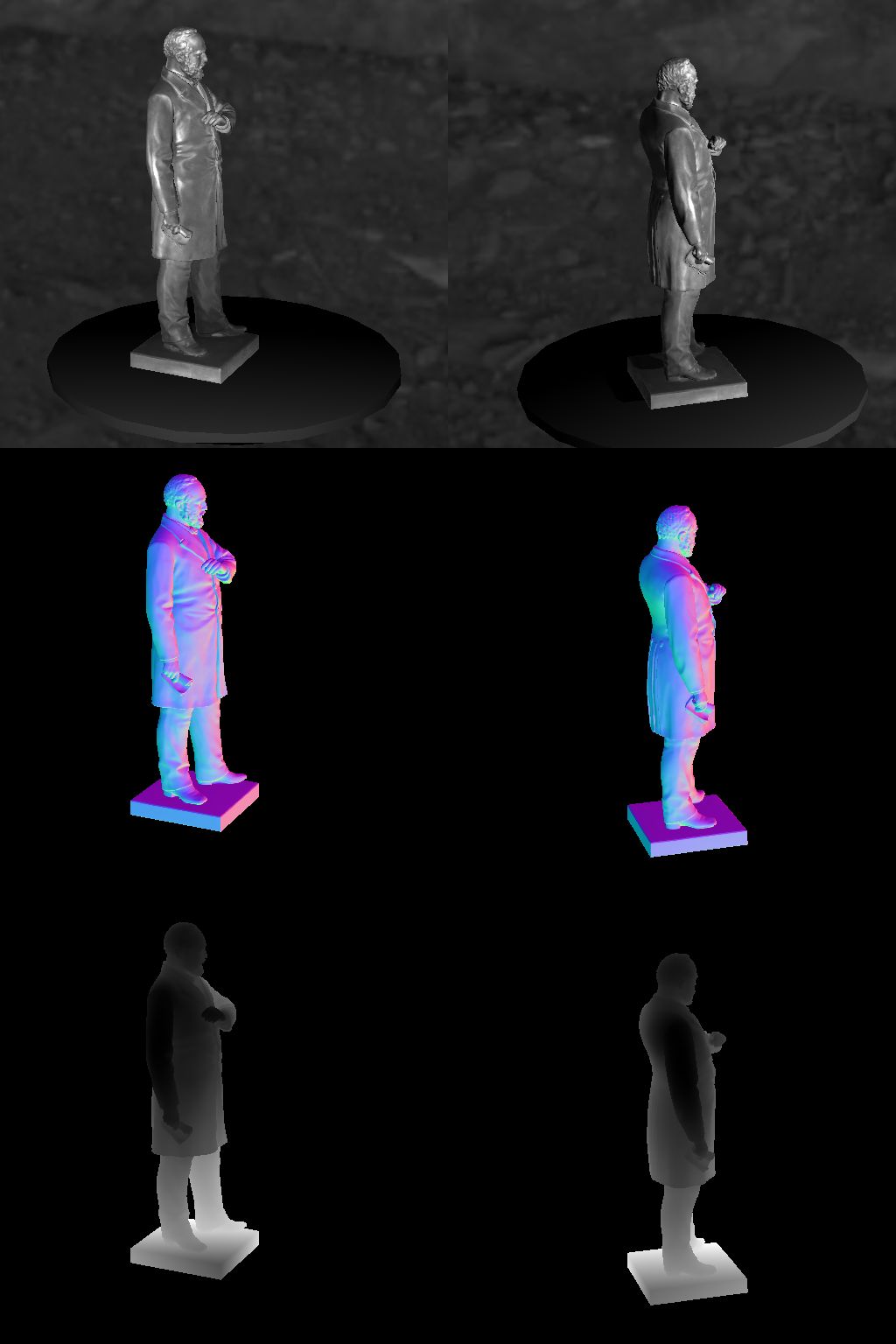} &
        \includegraphics[height=0.16\textwidth]{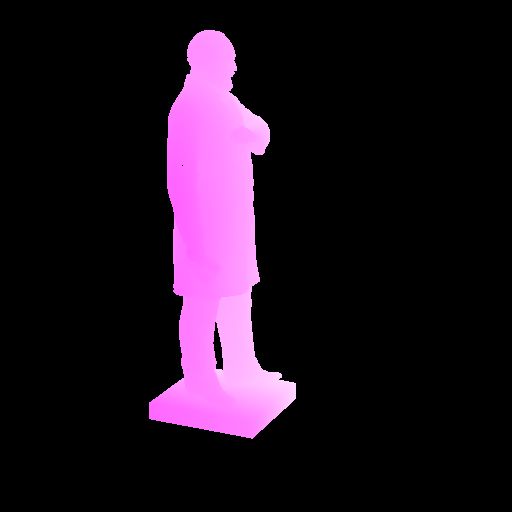} &
        \includegraphics[height=0.16\textwidth]{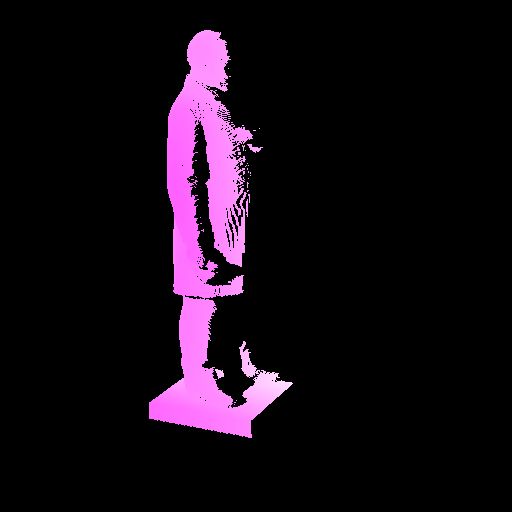} & 
        \includegraphics[height=0.16\textwidth]{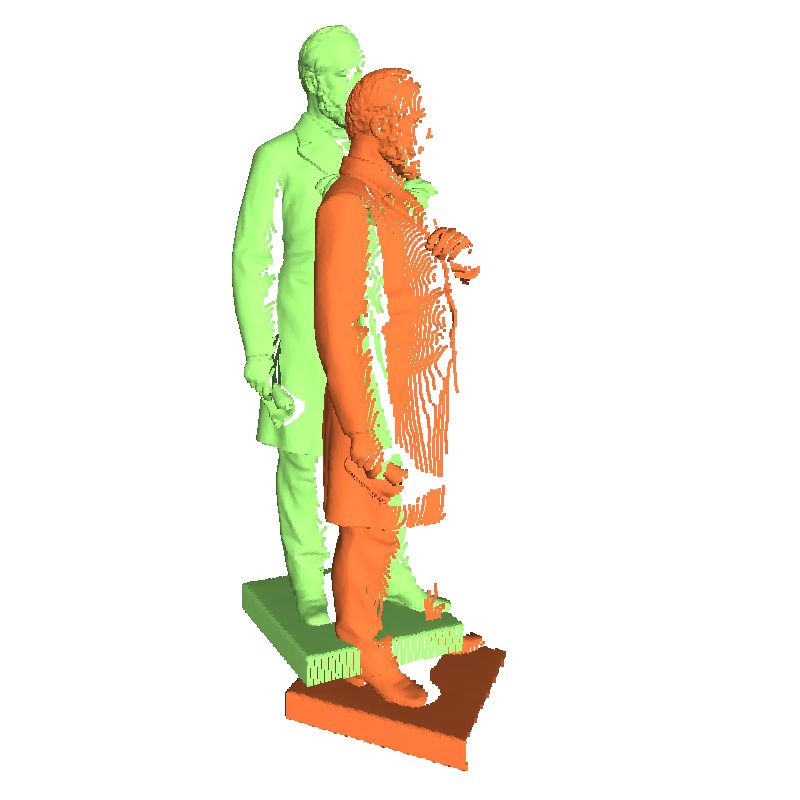} &
        \includegraphics[height=0.16\textwidth]{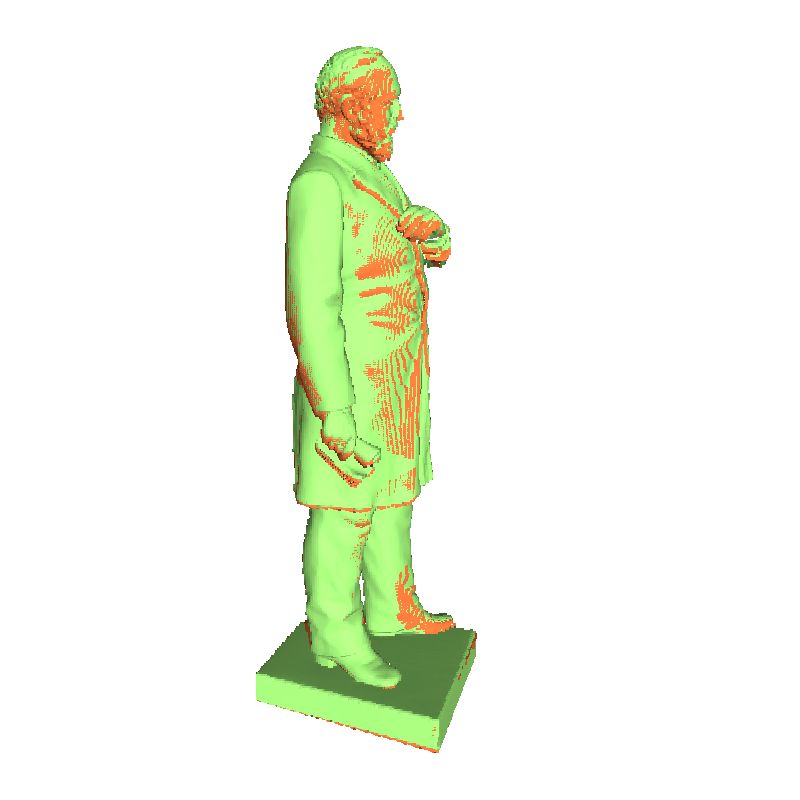}
    \end{tabular}
    \captionof{figure}{Qualitative results of the proposed method on training (top 3 rows) and test (bottom 3 rows) data of the synthetic \textit{consistent light} dataset. The situation of \textit{consistent light} represents the standard case, where for example the camera moves through a static scene with static light sources. The brightness assumption is usually not violated. The network generalizes well from known training to unknown test data.}
    \label{Figure:ConsistentLight}
\end{table*}
\begin{table*}[]
    \centering
    \begin{tabular}{cccccc} 
        & Input & Optical Flow & Ground Truth & Initial Pose & Alignment\\\vspace{0.05cm}\\
        \begin{tabular}{c}\vspace{-2.8cm}\\Synthetic\\Train Data\\\#00207\end{tabular} &
        \includegraphics[height=0.16\textwidth]{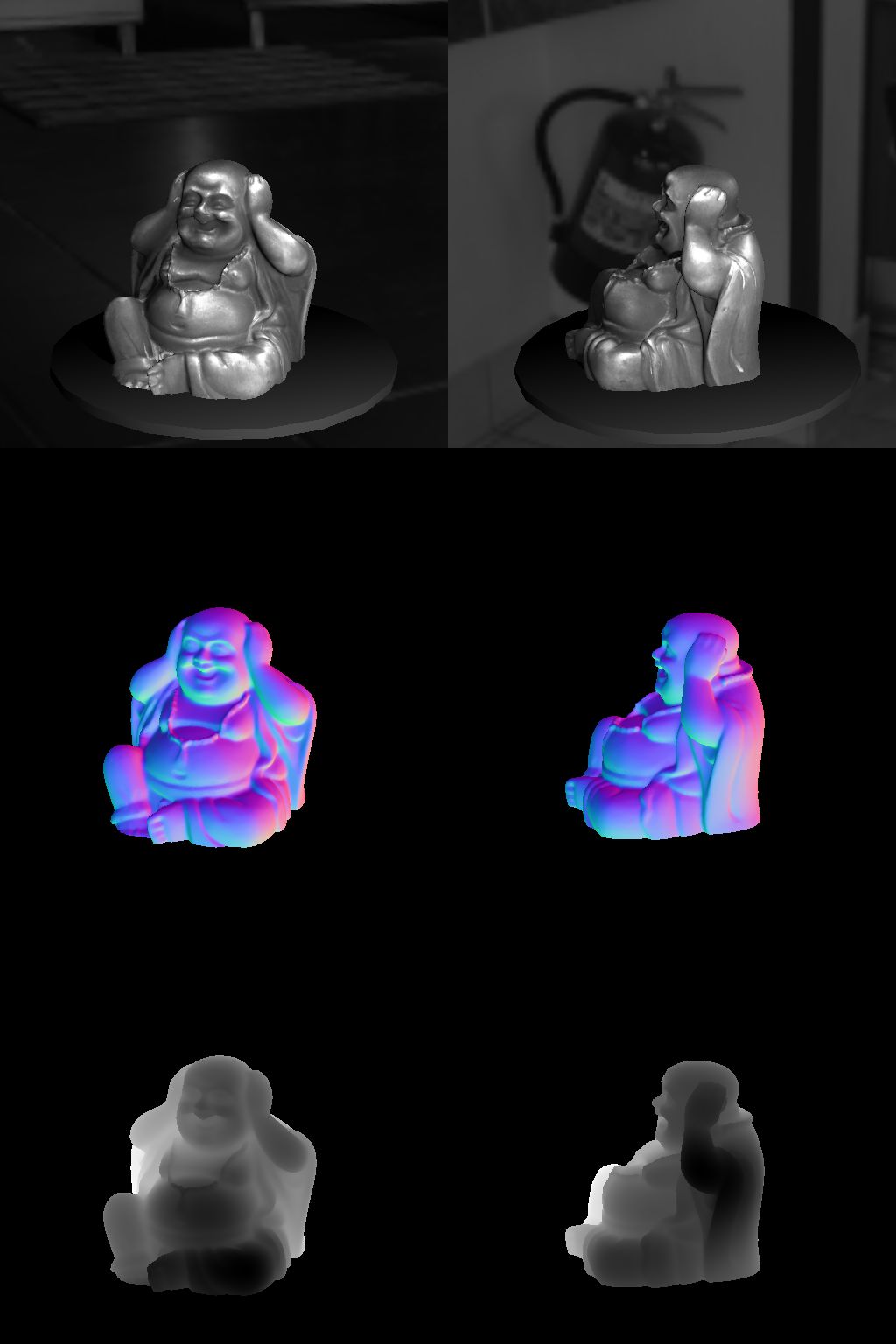} &
        \includegraphics[height=0.16\textwidth]{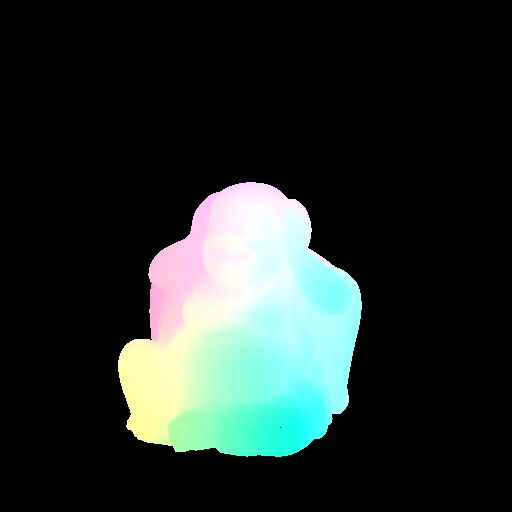} &
        \includegraphics[height=0.16\textwidth]{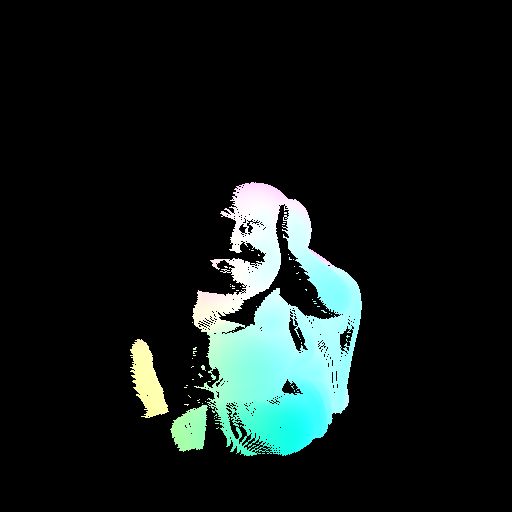} & 
        \includegraphics[height=0.16\textwidth]{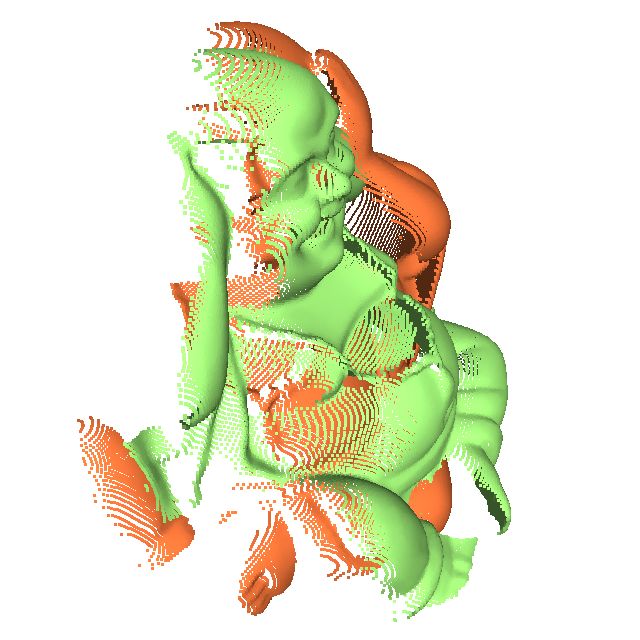} &
        \includegraphics[height=0.16\textwidth]{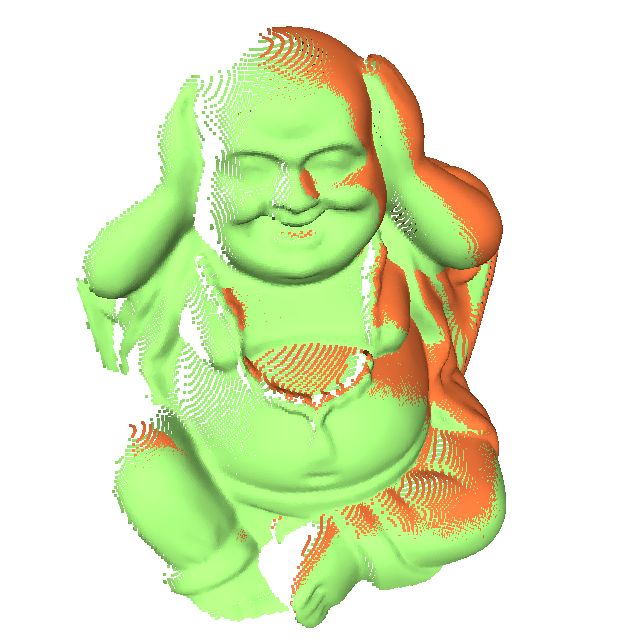}\vspace{0.2cm}\\
        \begin{tabular}{c}\vspace{-2.8cm}\\Synthetic\\Train Data\\\#00381\end{tabular} &
        \includegraphics[height=0.16\textwidth]{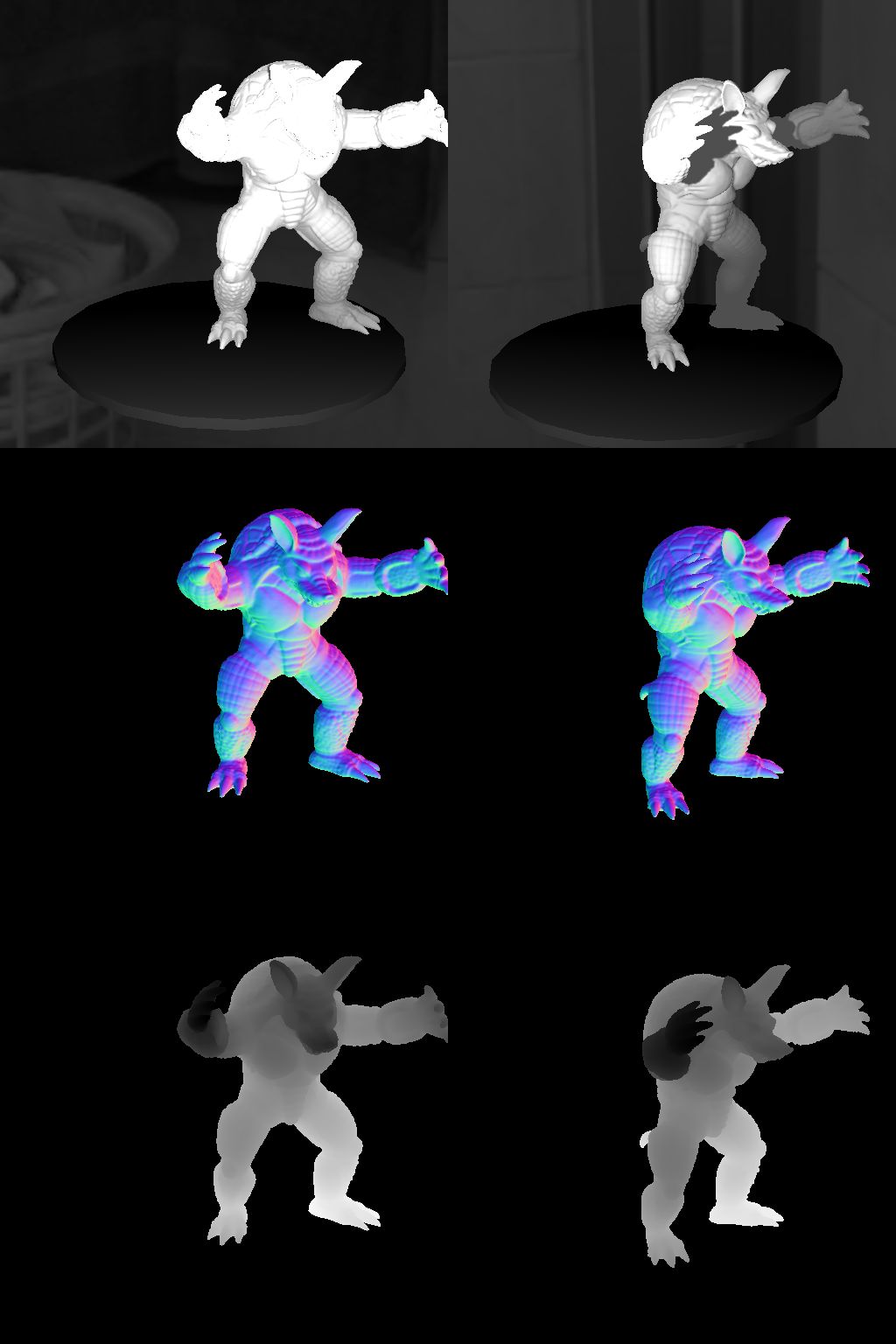} &
        \includegraphics[height=0.16\textwidth]{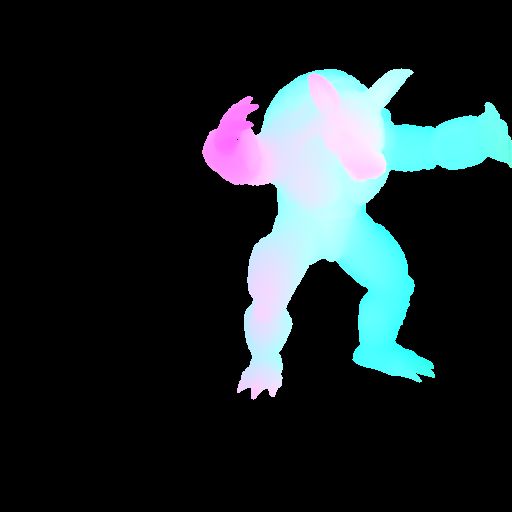} &
        \includegraphics[height=0.16\textwidth]{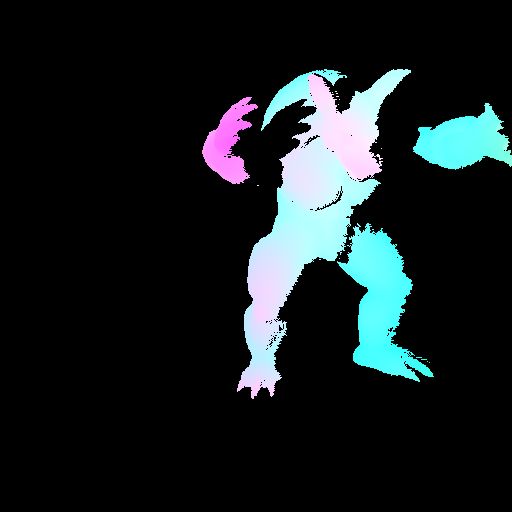} & 
        \includegraphics[height=0.16\textwidth]{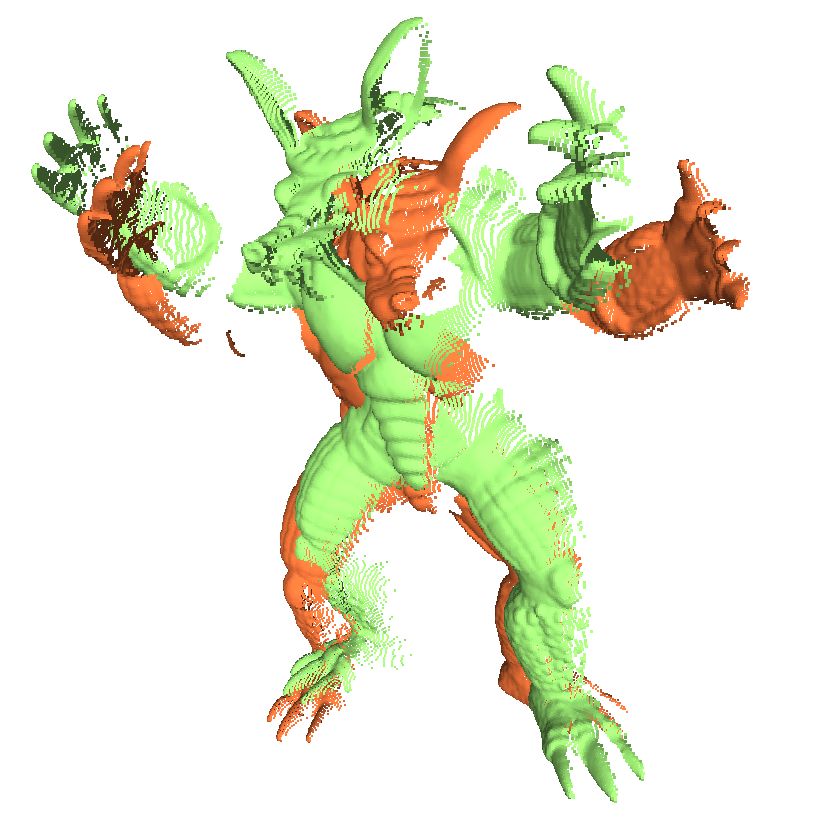} &
        \includegraphics[height=0.16\textwidth]{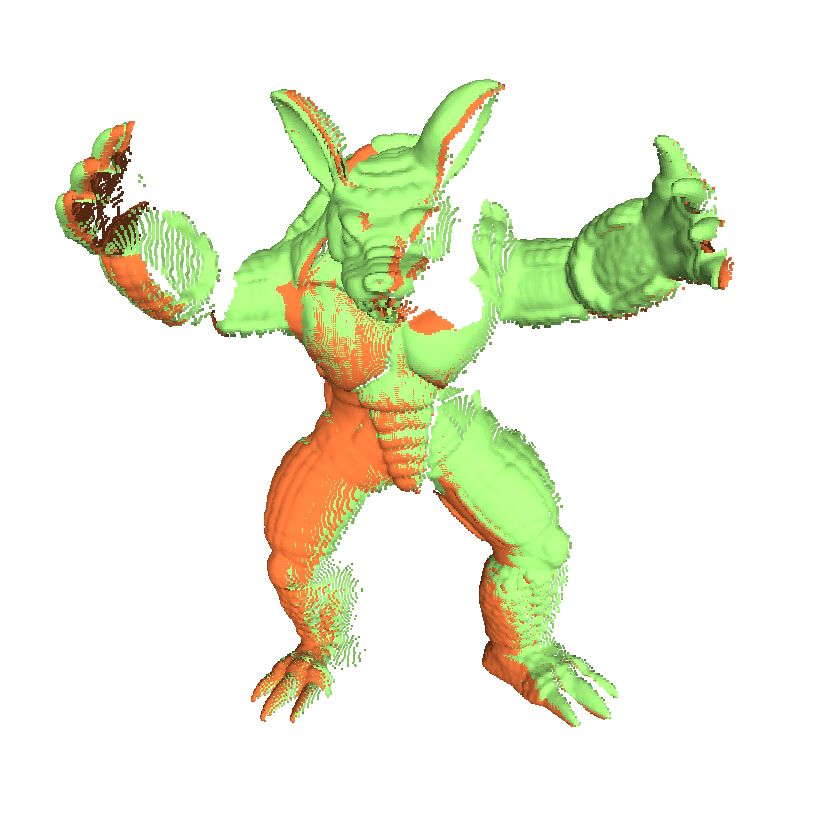}\vspace{0.2cm}\\
        \begin{tabular}{c}\vspace{-2.8cm}\\Synthetic\\Train Data\\\#00338\end{tabular} &
        \includegraphics[height=0.16\textwidth]{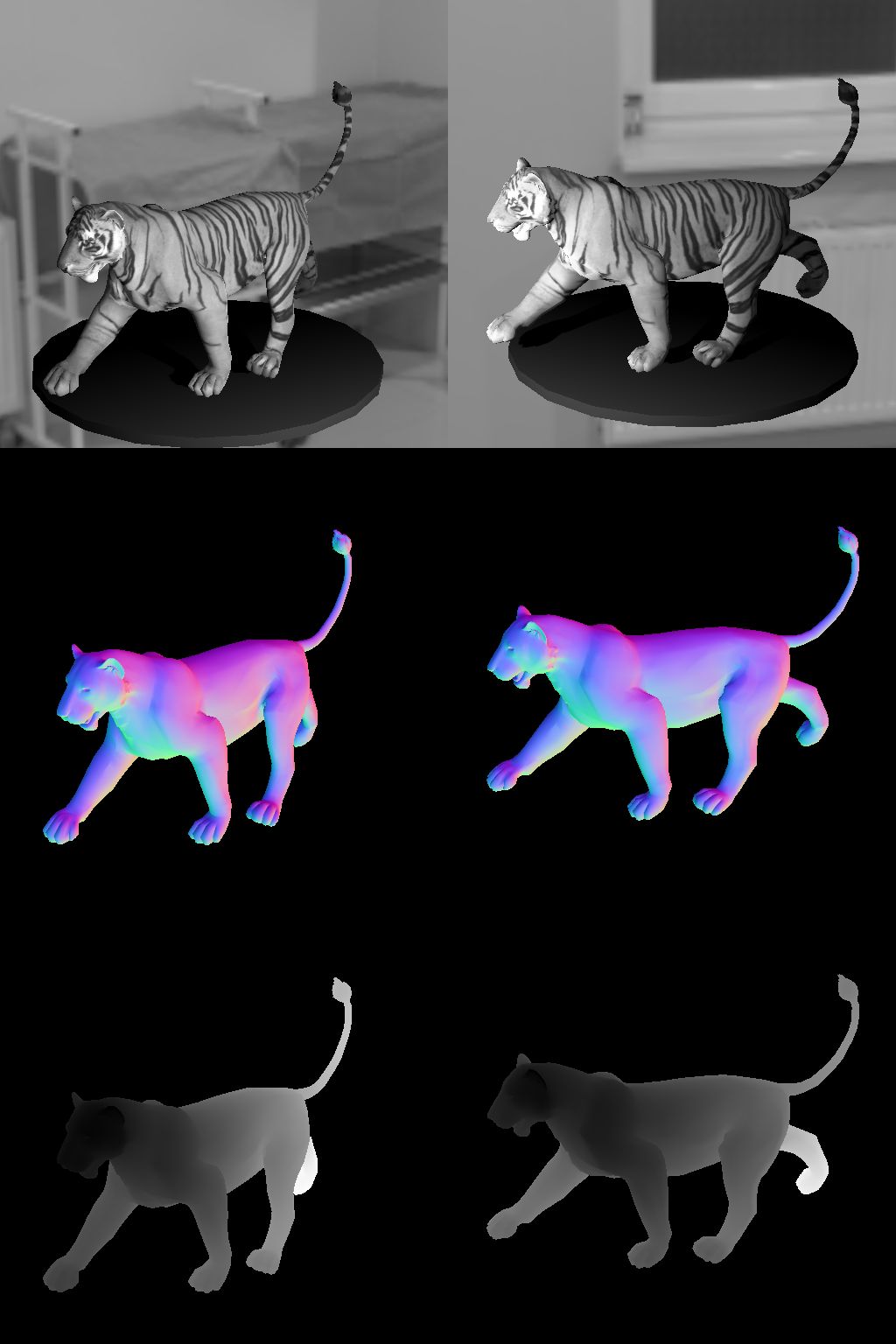} &
        \includegraphics[height=0.16\textwidth]{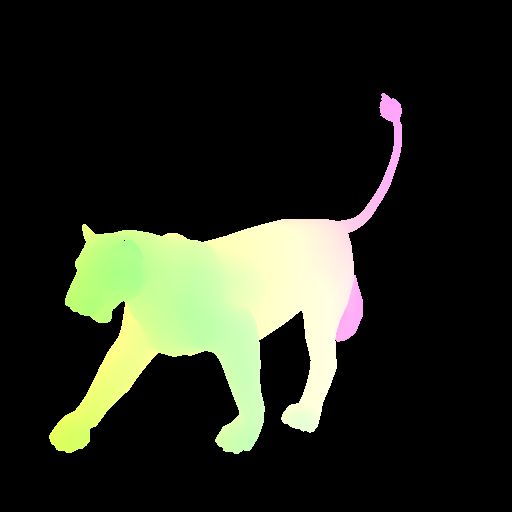} &
        \includegraphics[height=0.16\textwidth]{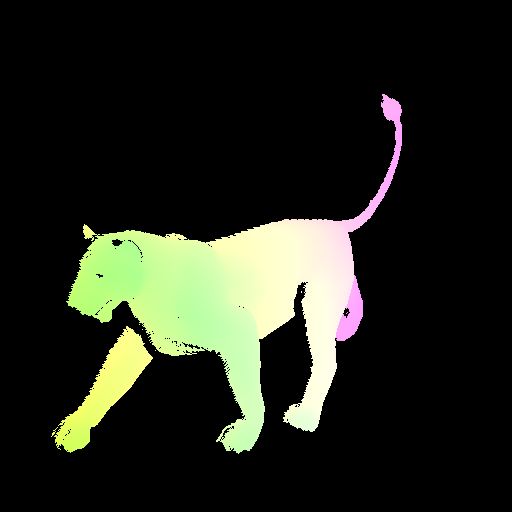} & 
        \includegraphics[height=0.16\textwidth]{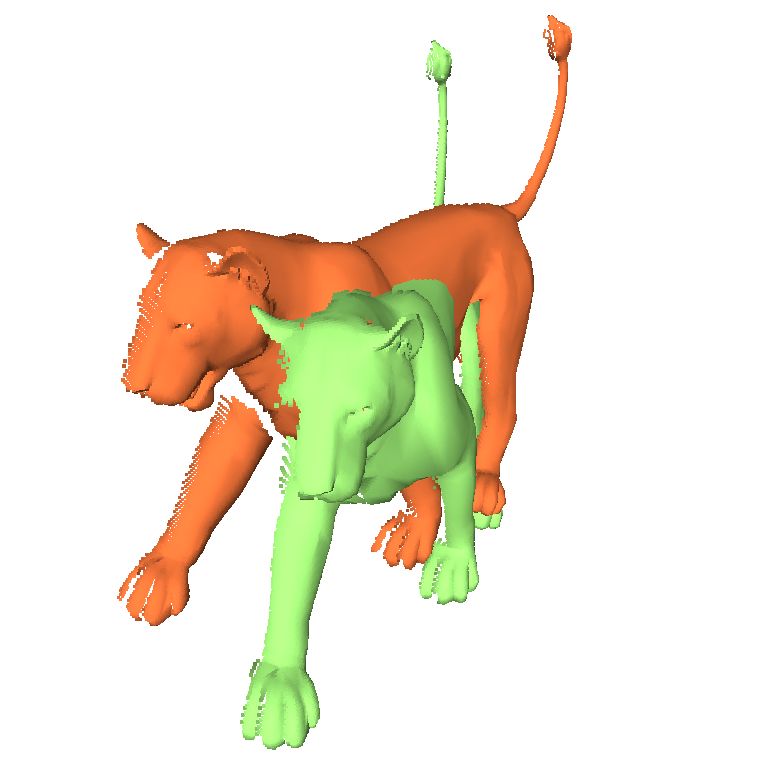} &
        \includegraphics[height=0.16\textwidth]{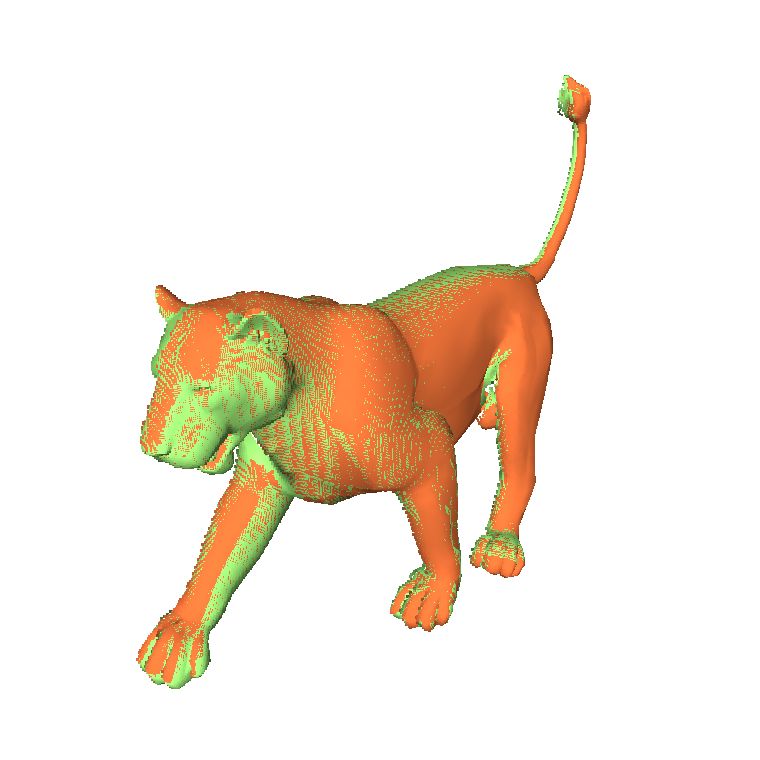}\vspace{1cm}\\
        \begin{tabular}{c}\vspace{-2.8cm}\\Synthetic\\Test Data\\\#00323\end{tabular} &
        \includegraphics[height=0.16\textwidth]{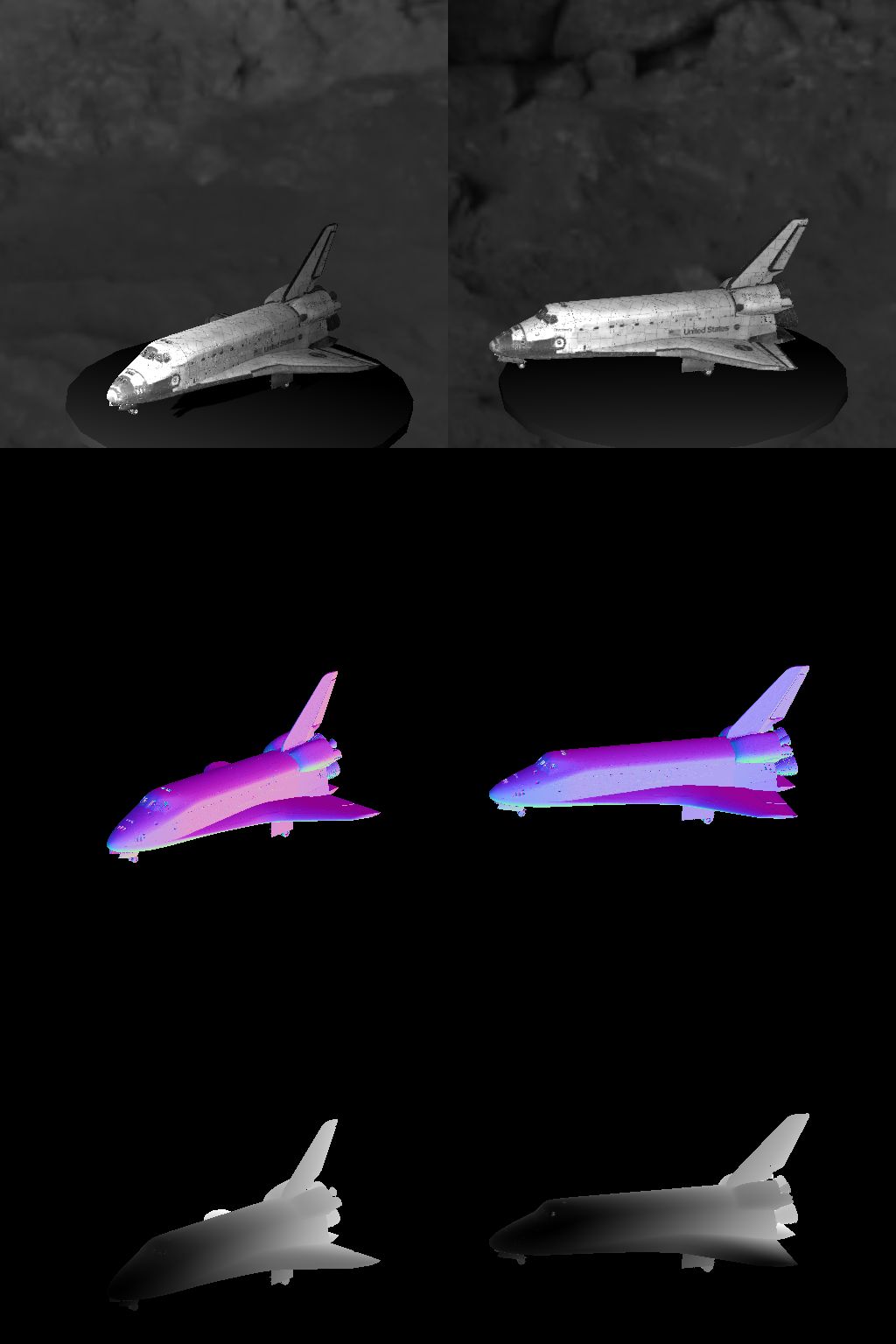} &
        \includegraphics[height=0.16\textwidth]{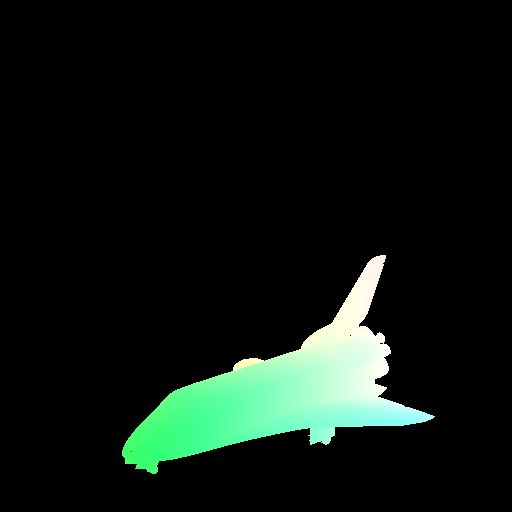} &
        \includegraphics[height=0.16\textwidth]{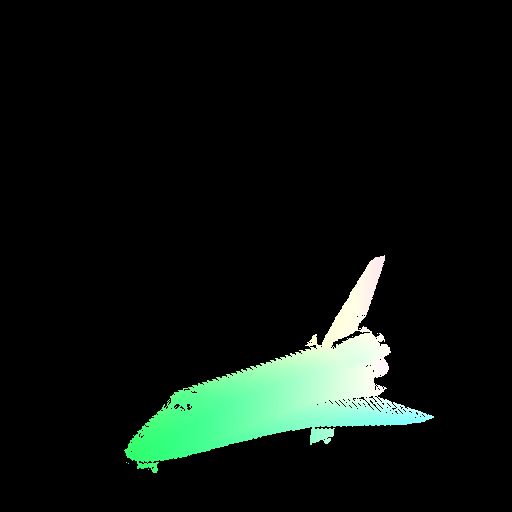} & 
        \includegraphics[height=0.16\textwidth]{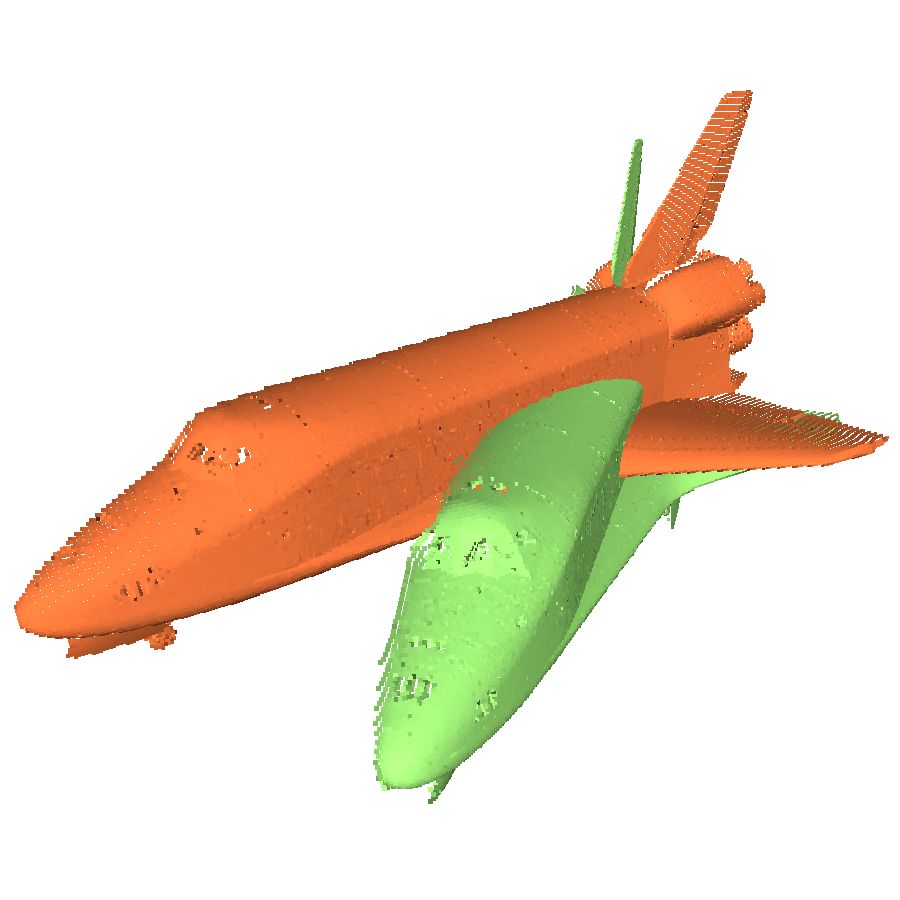} &
        \includegraphics[height=0.16\textwidth]{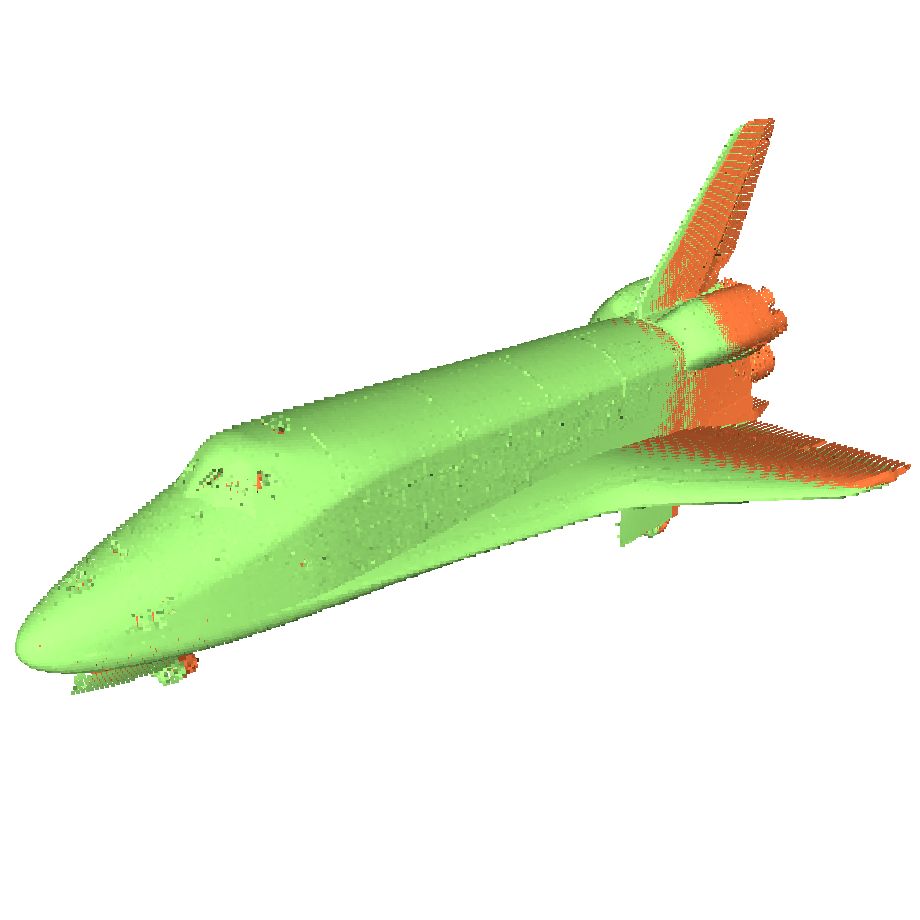}\vspace{0.2cm}\\
        \begin{tabular}{c}\vspace{-2.8cm}\\Synthetic\\Test Data\\\#00263\end{tabular} &
        \includegraphics[height=0.16\textwidth]{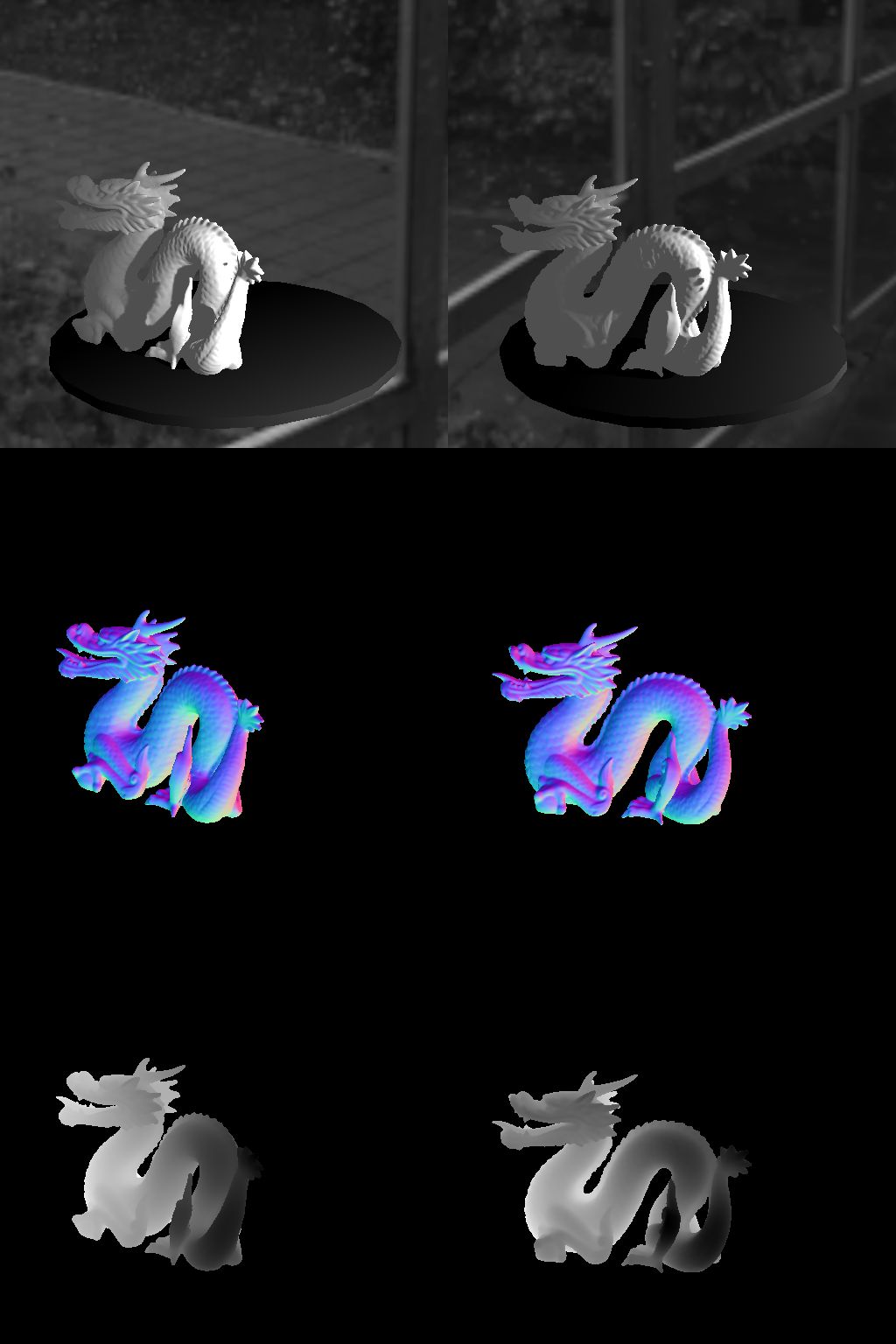} &
        \includegraphics[height=0.16\textwidth]{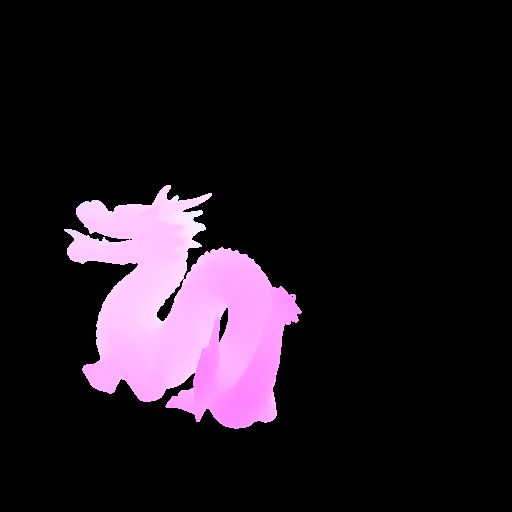} &
        \includegraphics[height=0.16\textwidth]{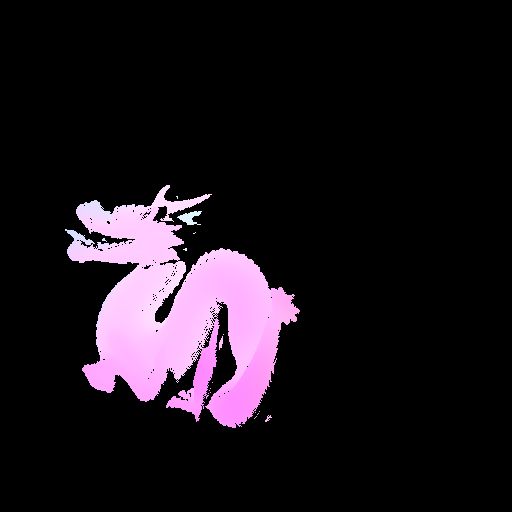} & 
        \includegraphics[height=0.16\textwidth]{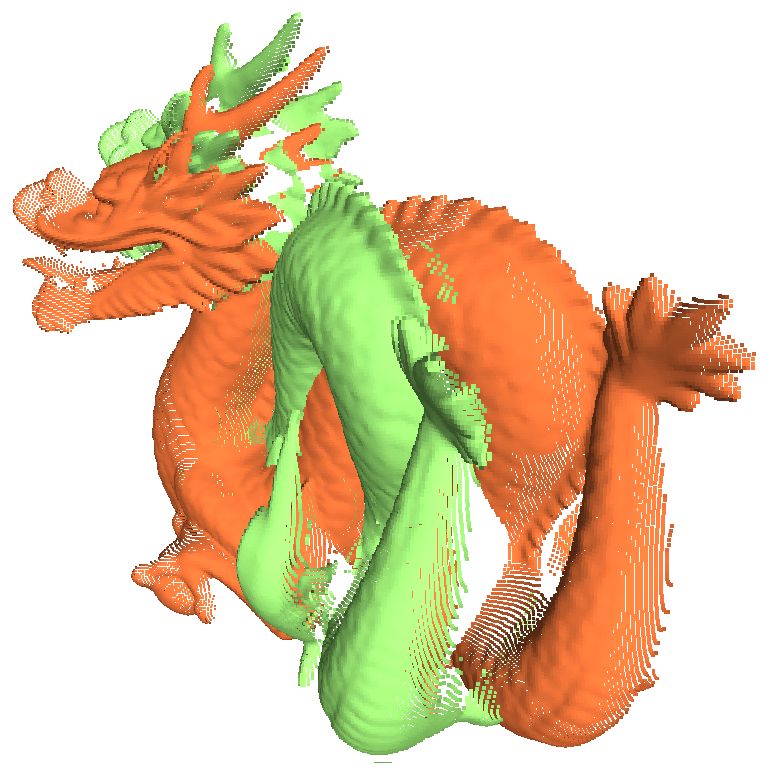} &
        \includegraphics[height=0.16\textwidth]{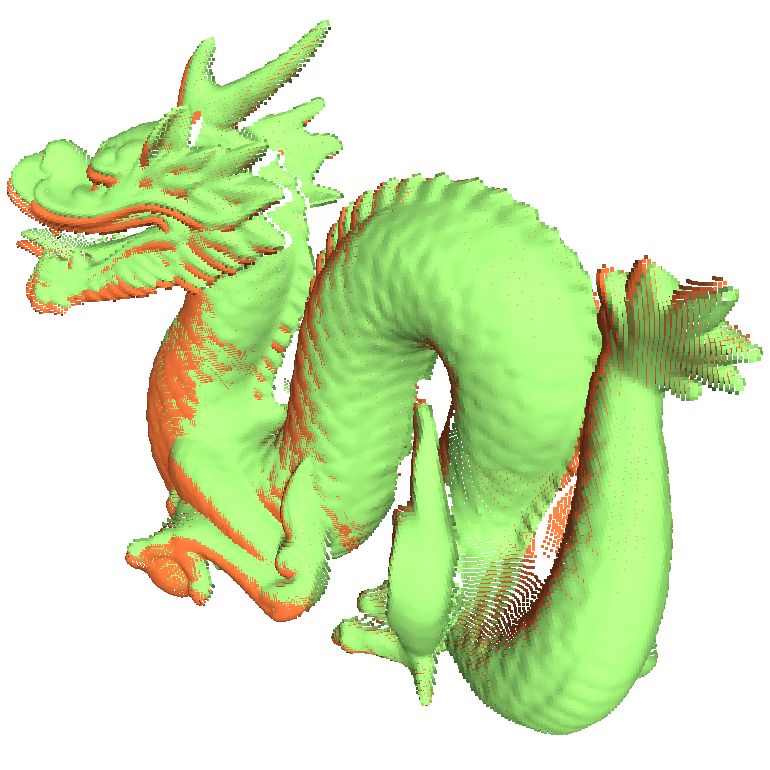}\vspace{0.2cm}\\
        \begin{tabular}{c}\vspace{-2.8cm}\\Real\\Test Data\\\#00075\end{tabular} &
        \includegraphics[height=0.16\textwidth]{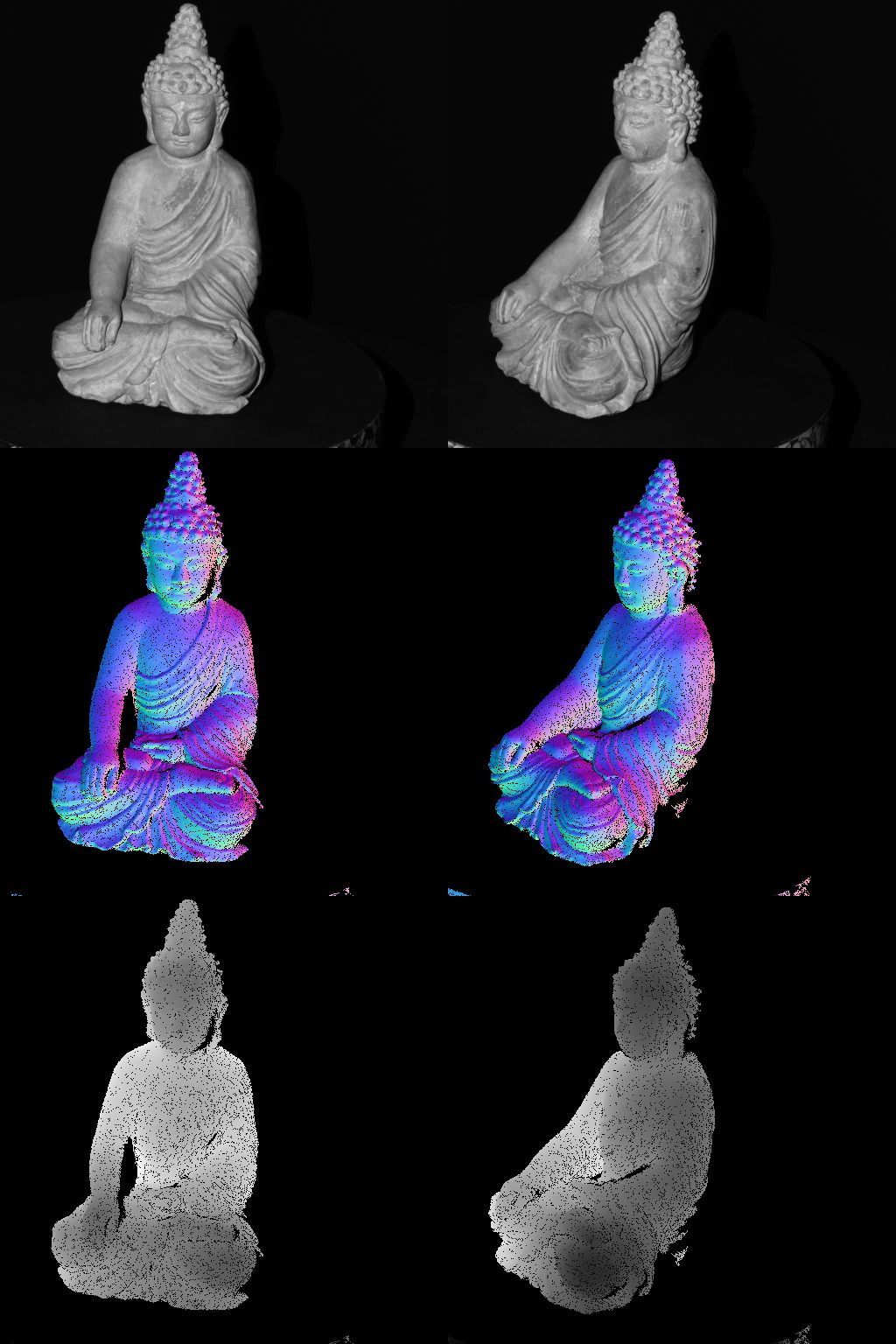} &
        \includegraphics[height=0.16\textwidth]{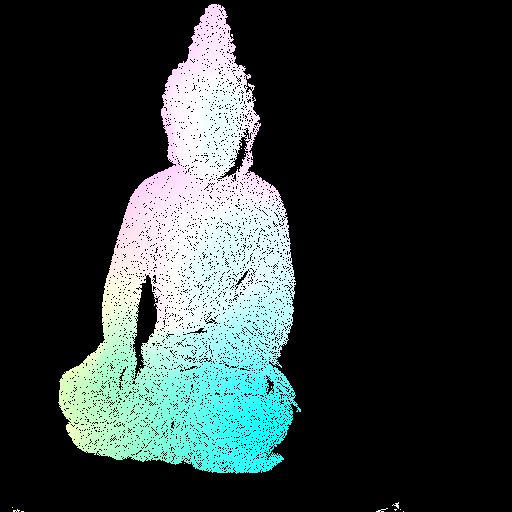} &
        \includegraphics[height=0.16\textwidth]{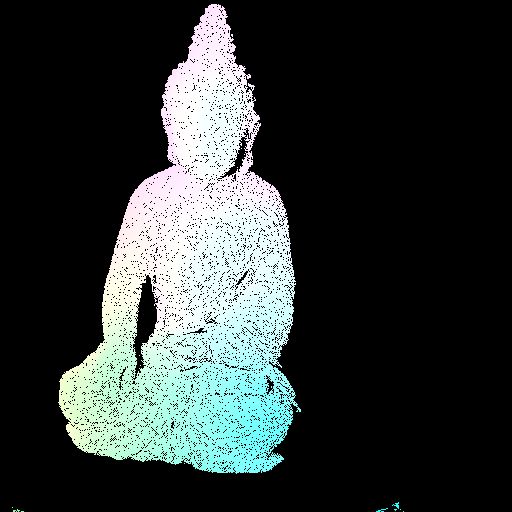} & 
        \includegraphics[height=0.16\textwidth]{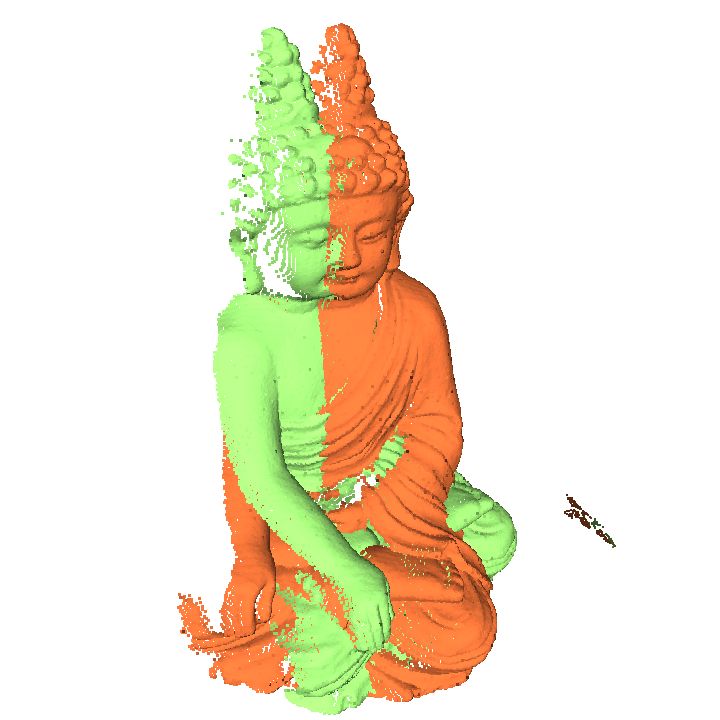} &
        \includegraphics[height=0.16\textwidth]{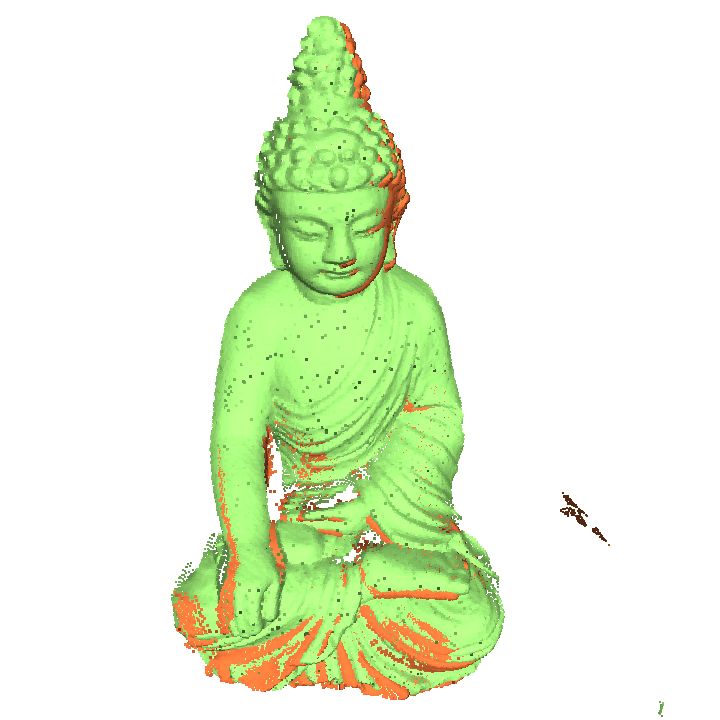}
    \end{tabular}
    \captionof{figure}{Qualitative results of the proposed method on training (top 3 rows) and test (bottom 3 rows) data of the synthetic \textit{inconsistent light} dataset as well as real test data. The situation of \textit{inconsistent light} represents the situation under investigation, motivating this paper, where the light sources or the objects in the scene move or rotate, yielding strong shading changes. The brightness assumption is dramatically violated. The network still generalizes well from known training to unknown test data. Even for real data without additional finetuning the results are impressive.}
    \label{Figure:InconsistentLight}
\end{table*}
\begin{table*}[]
    \centering
    \begin{tabular}{cccccc} 
        & Input & Optical Flow & Ground Truth & Initial Pose & Alignment\\\vspace{0.05cm}\\
        \begin{tabular}{c}\vspace{-2.8cm}\\Kitti\\Train Data\\\#00-105\end{tabular} &
        \includegraphics[height=0.16\textwidth]{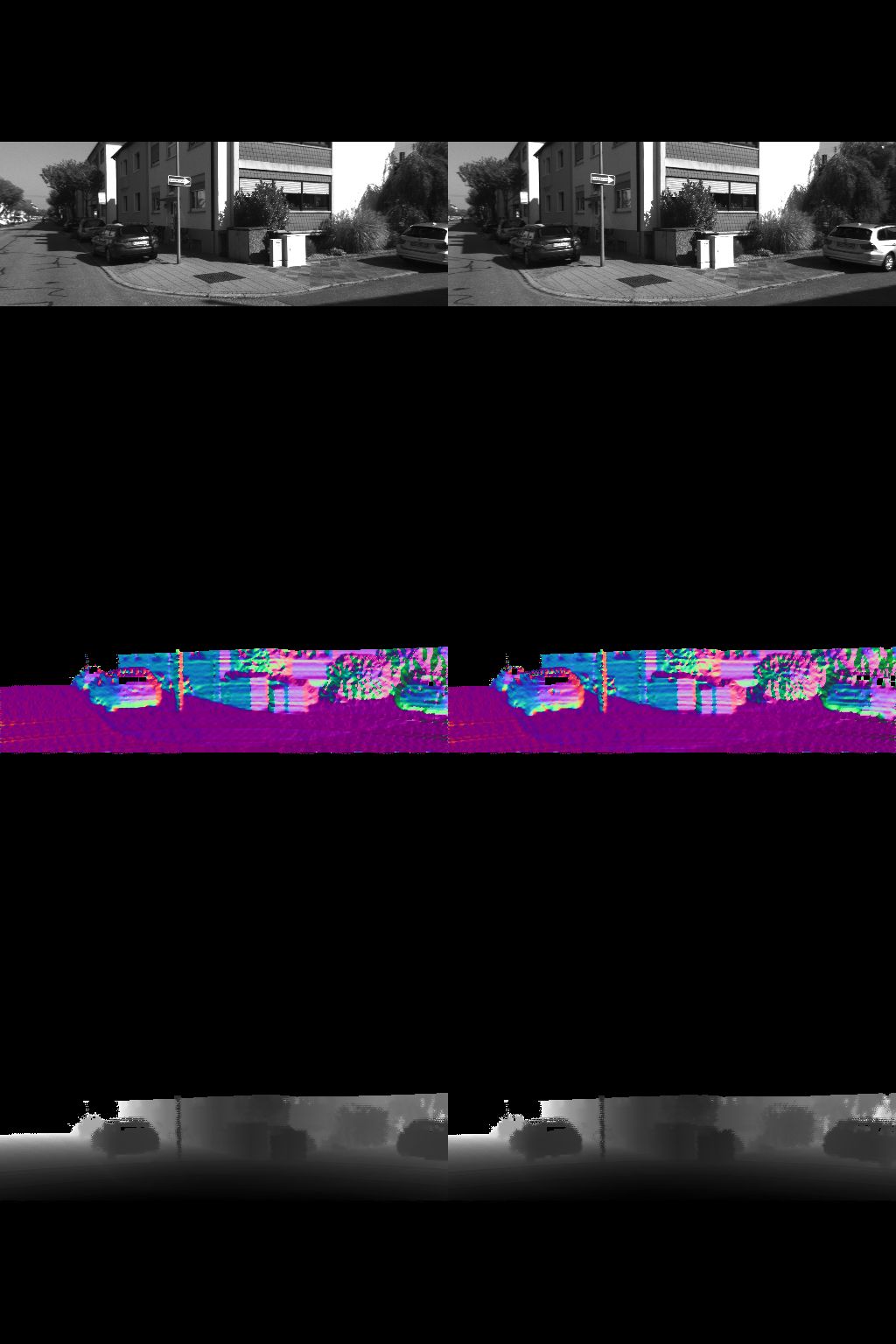} &
        \includegraphics[height=0.16\textwidth]{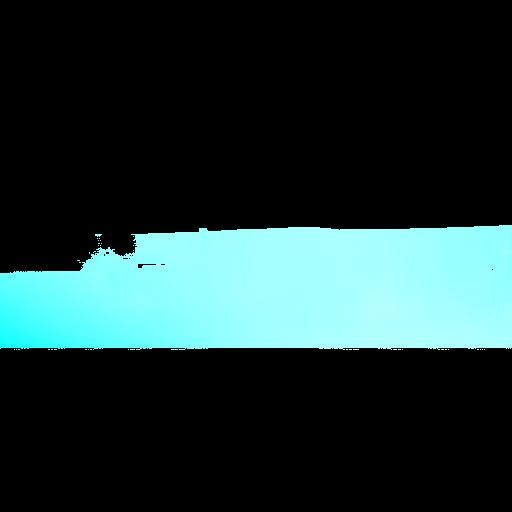} &
        \includegraphics[height=0.16\textwidth]{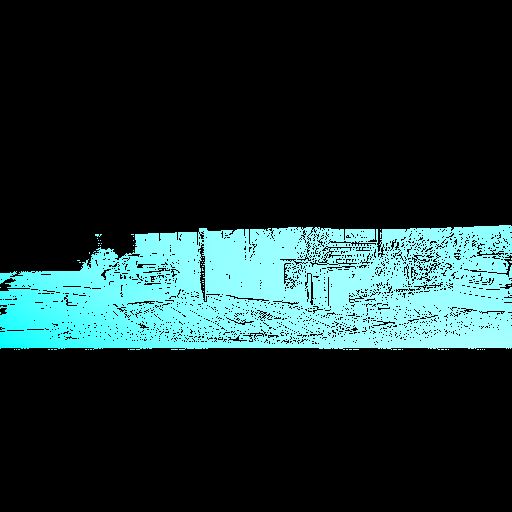} & 
        \includegraphics[height=0.16\textwidth]{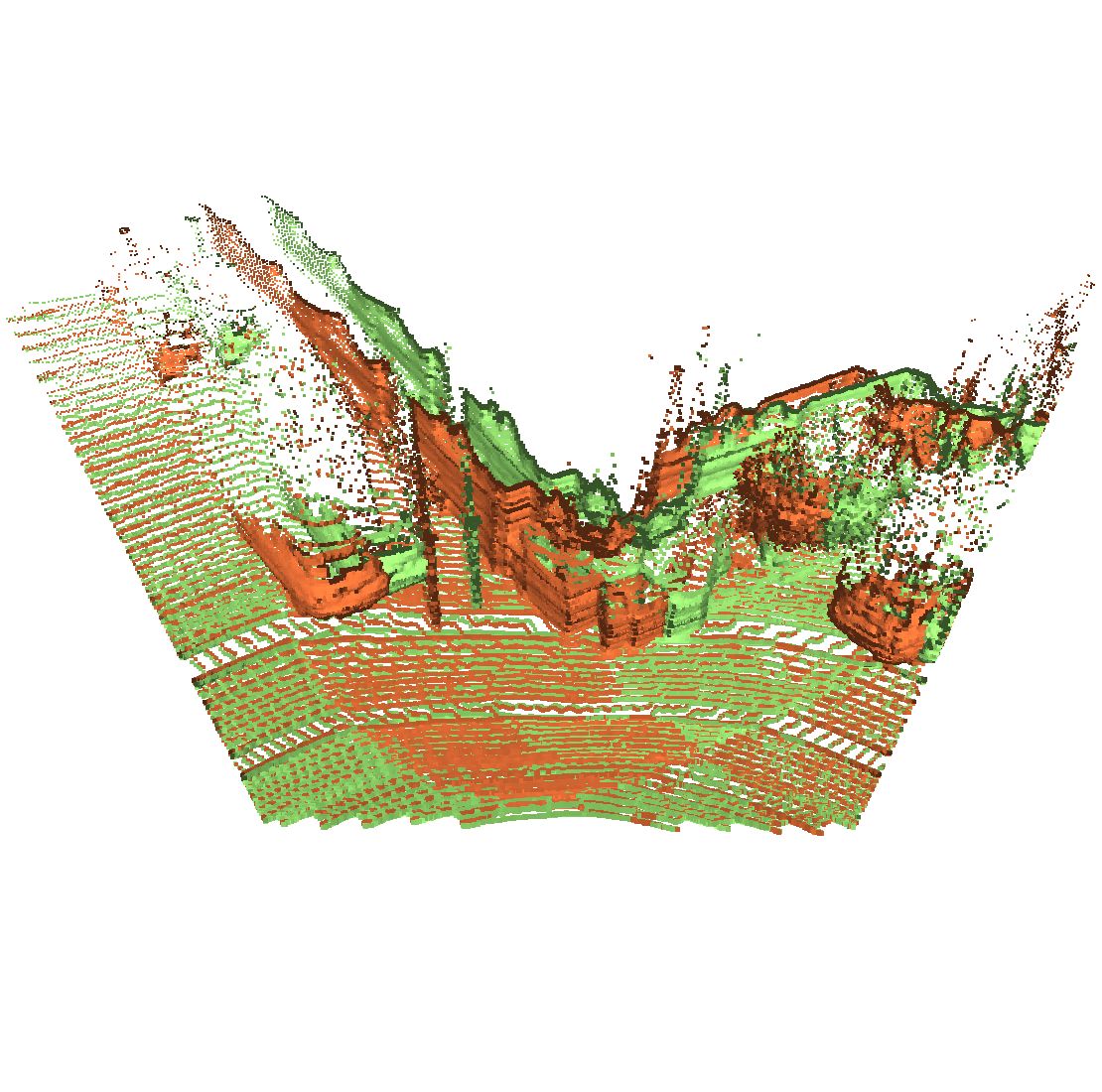} &
        \includegraphics[height=0.16\textwidth]{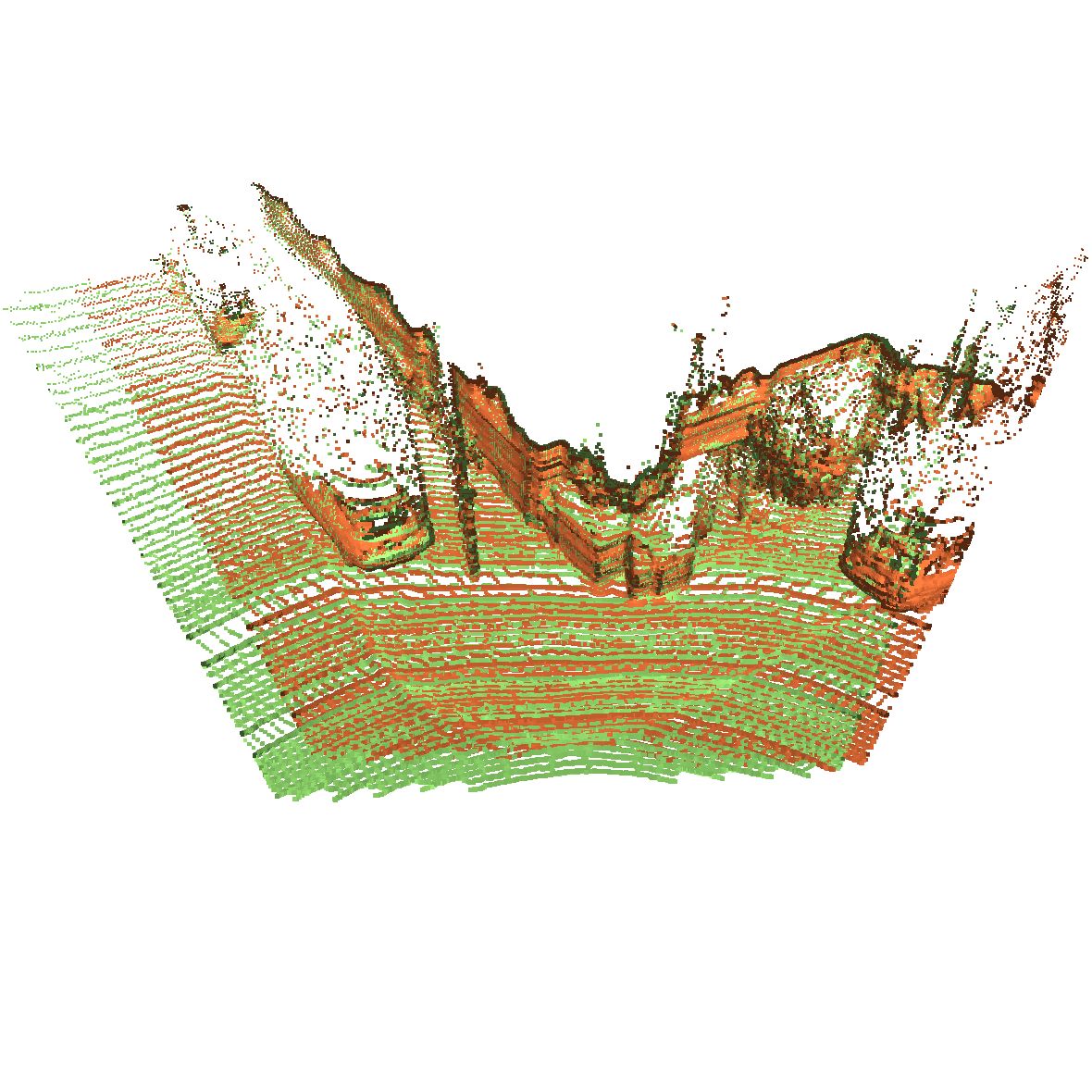}\vspace{0.2cm}\\
        \begin{tabular}{c}\vspace{-2.8cm}\\Kitti\\Train Data\\\#03-93\end{tabular} &
        \includegraphics[height=0.16\textwidth]{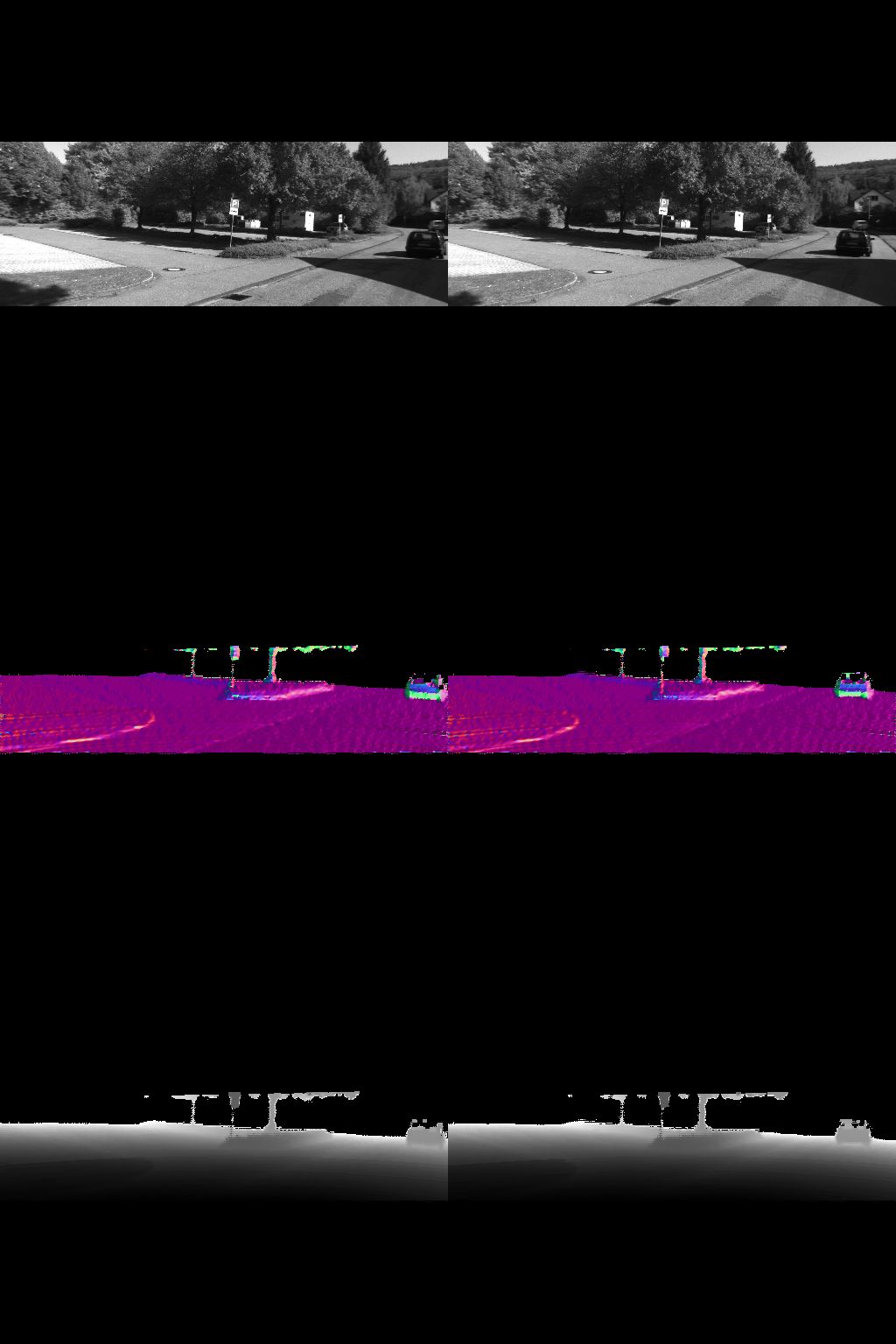} &
        \includegraphics[height=0.16\textwidth]{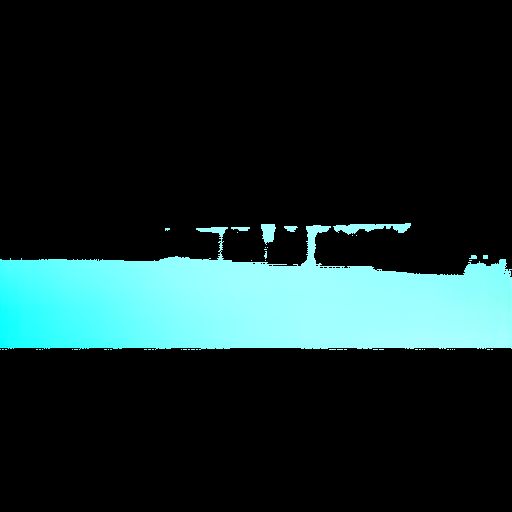} &
        \includegraphics[height=0.16\textwidth]{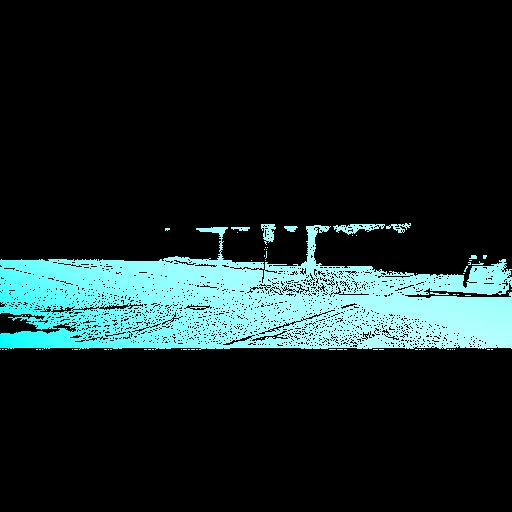} & 
        \includegraphics[height=0.16\textwidth]{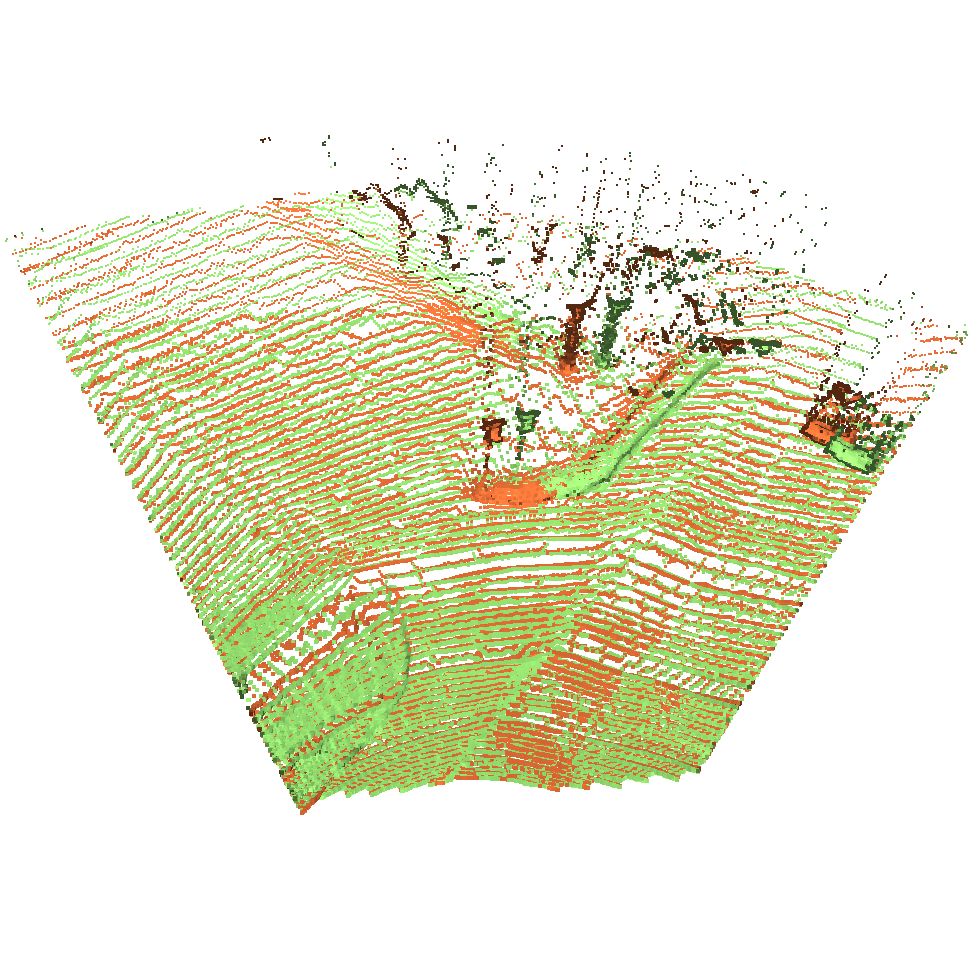} &
        \includegraphics[height=0.16\textwidth]{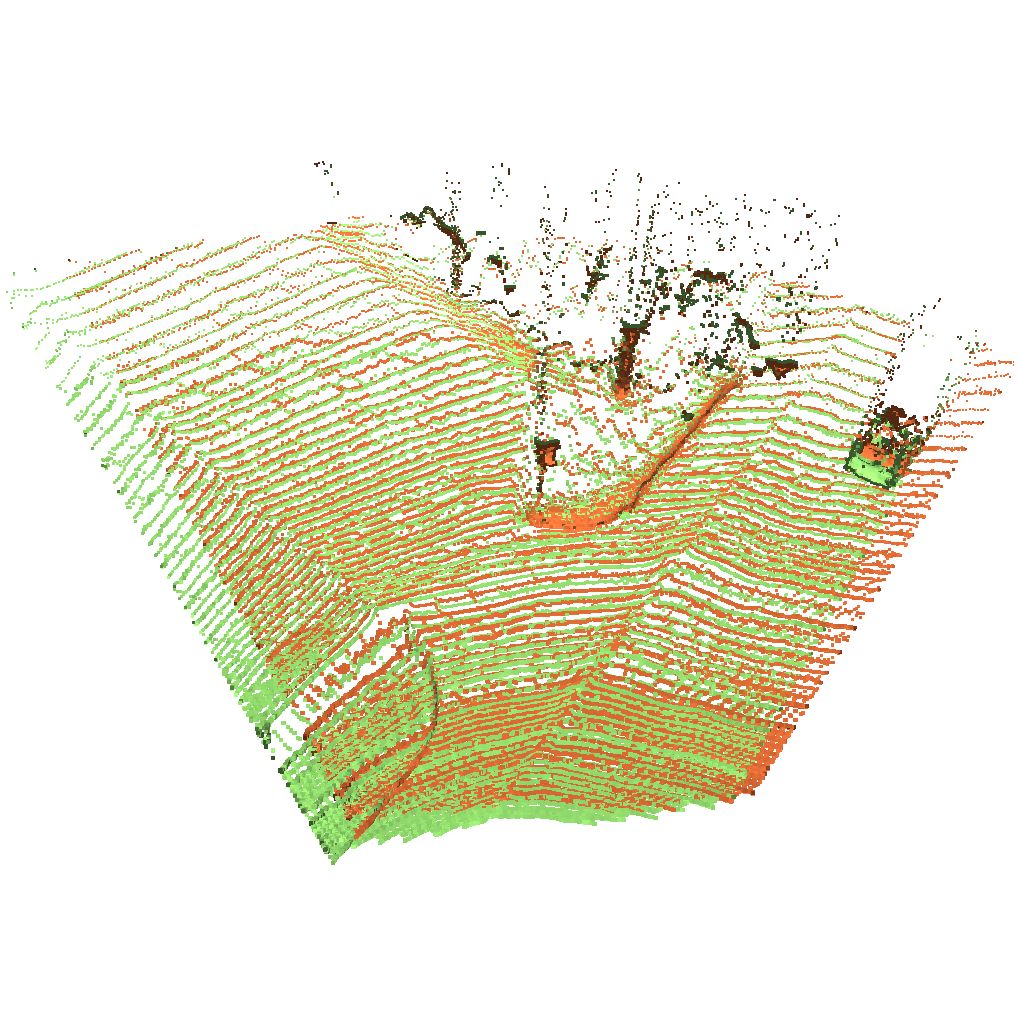}\vspace{0.2cm}\\
        \begin{tabular}{c}\vspace{-2.8cm}\\Kitti\\Train Data\\\#06-301\end{tabular} &
        \includegraphics[height=0.16\textwidth]{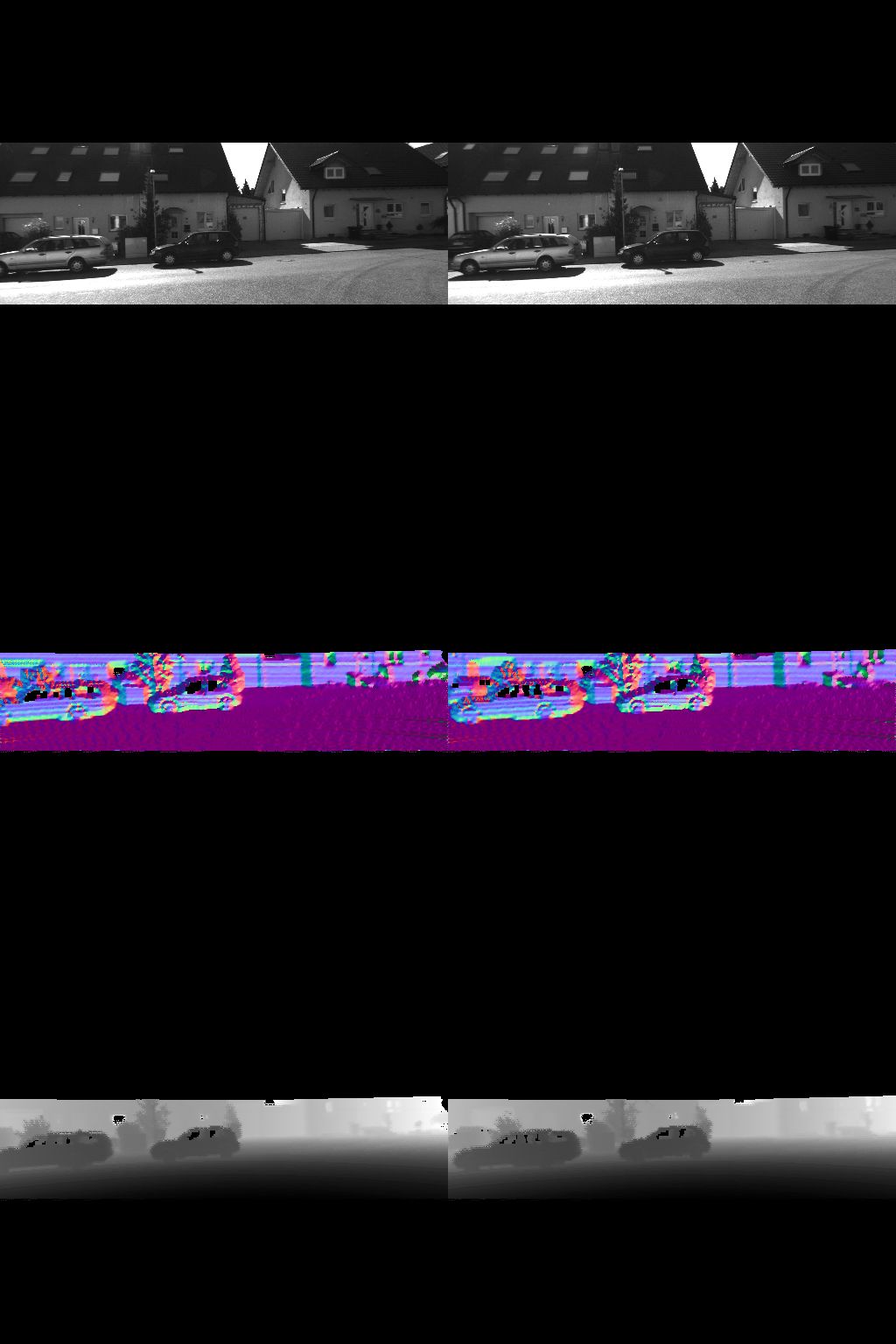} &
        \includegraphics[height=0.16\textwidth]{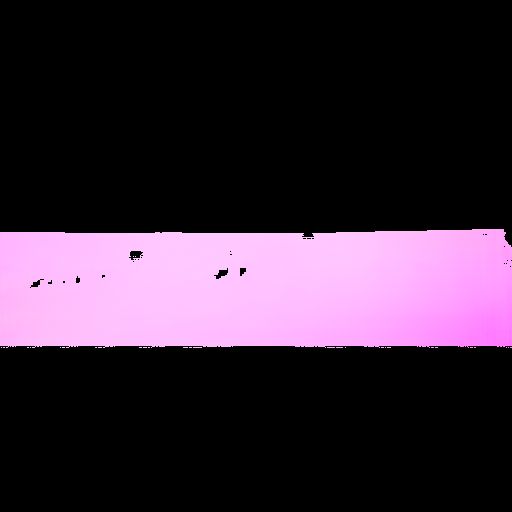} &
        \includegraphics[height=0.16\textwidth]{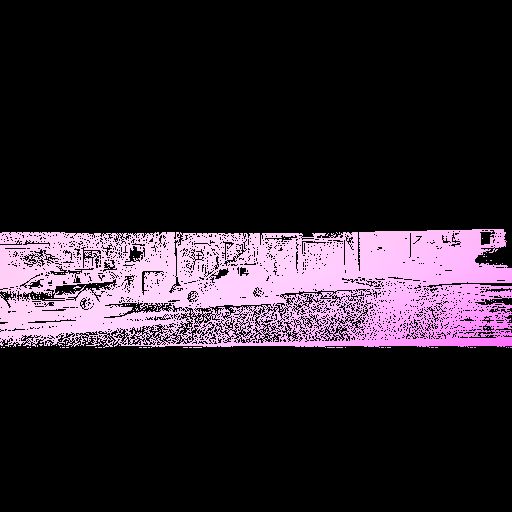} & 
        \includegraphics[height=0.16\textwidth]{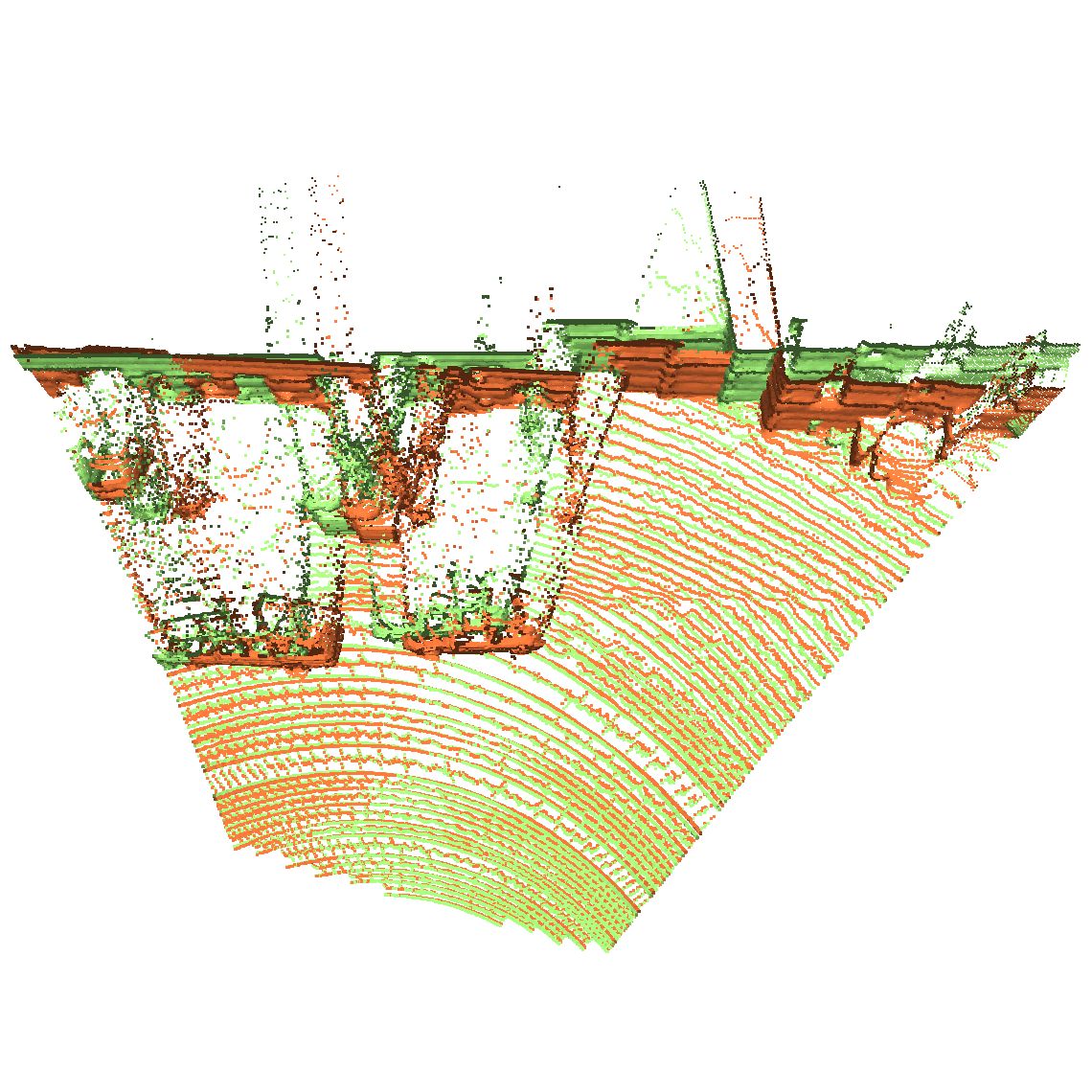} &
        \includegraphics[height=0.16\textwidth]{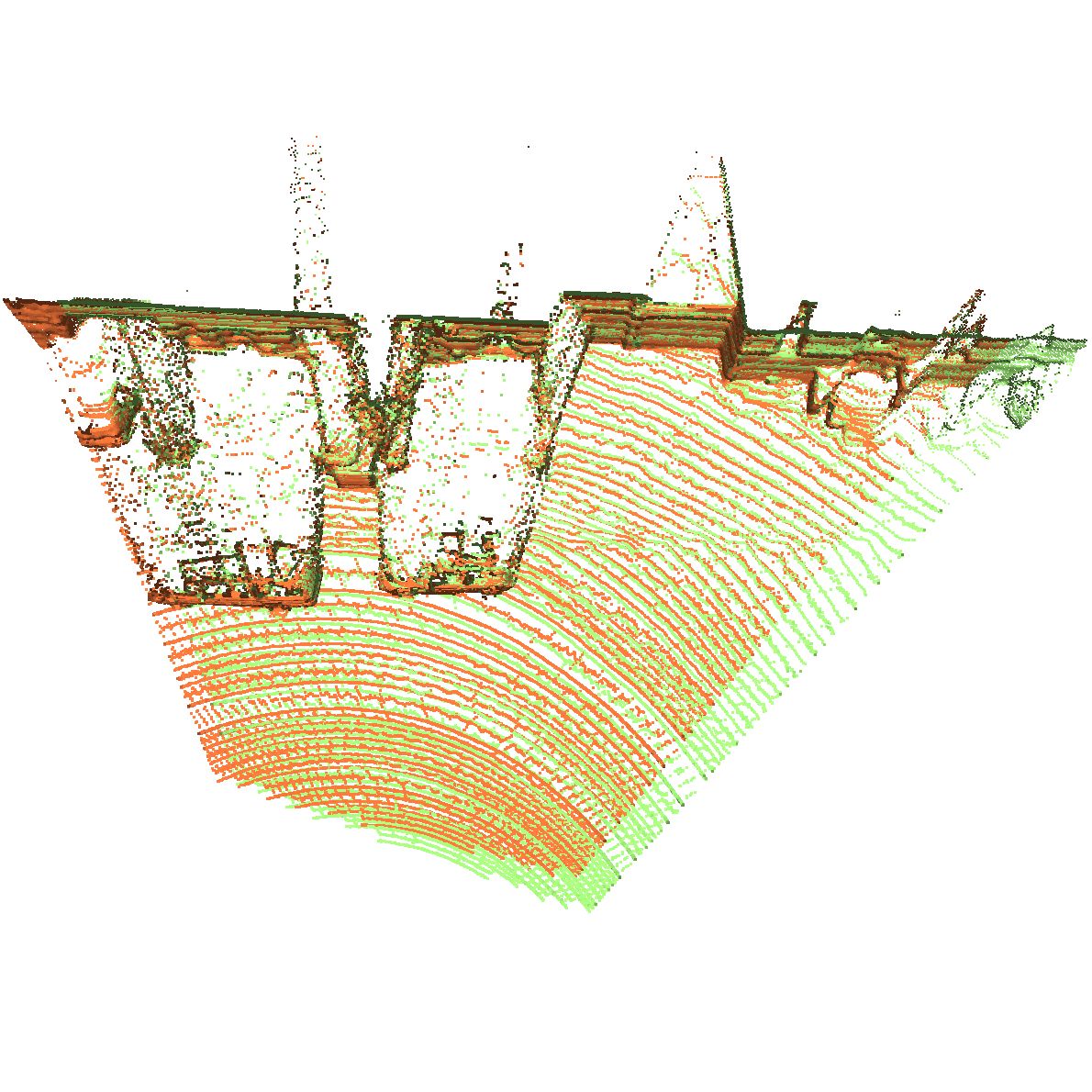}\vspace{1cm}\\
        \begin{tabular}{c}\vspace{-2.8cm}\\Kitti\\Test Data\\\#11-408\end{tabular} &
        \includegraphics[height=0.16\textwidth]{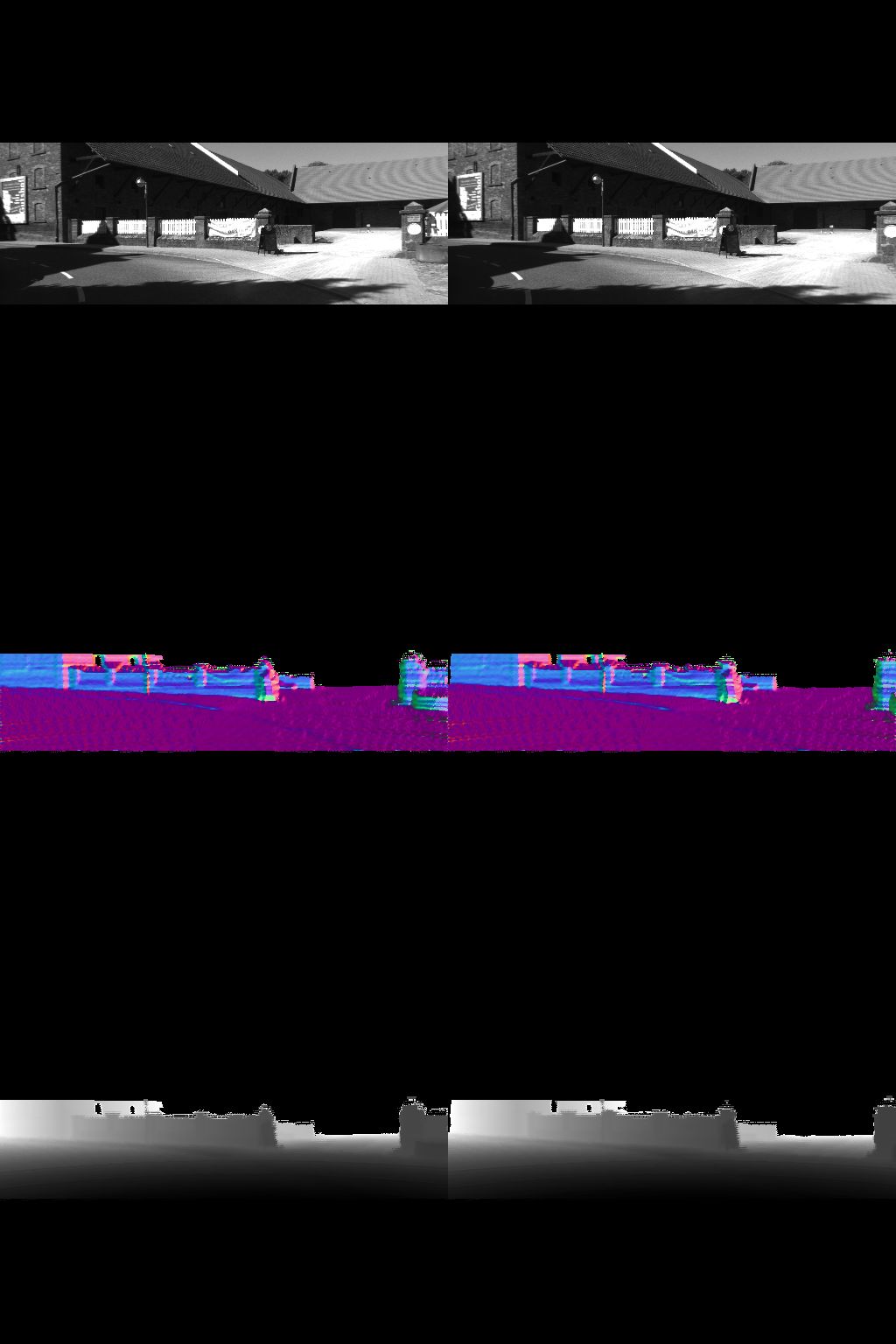} &
        \includegraphics[height=0.16\textwidth]{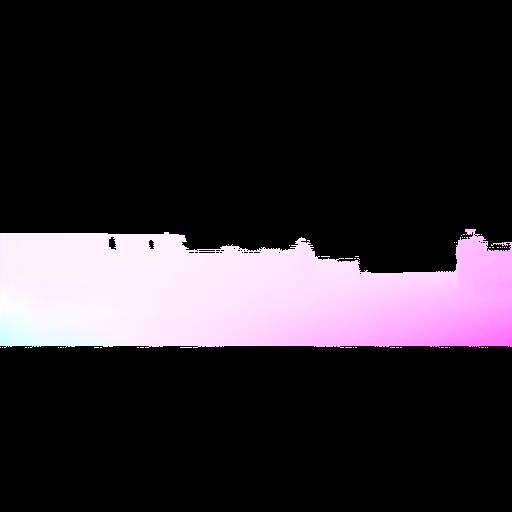} &\begin{tabular}{c}\vspace{-2.8cm}\\n.a.\end{tabular}& 
        \includegraphics[height=0.16\textwidth]{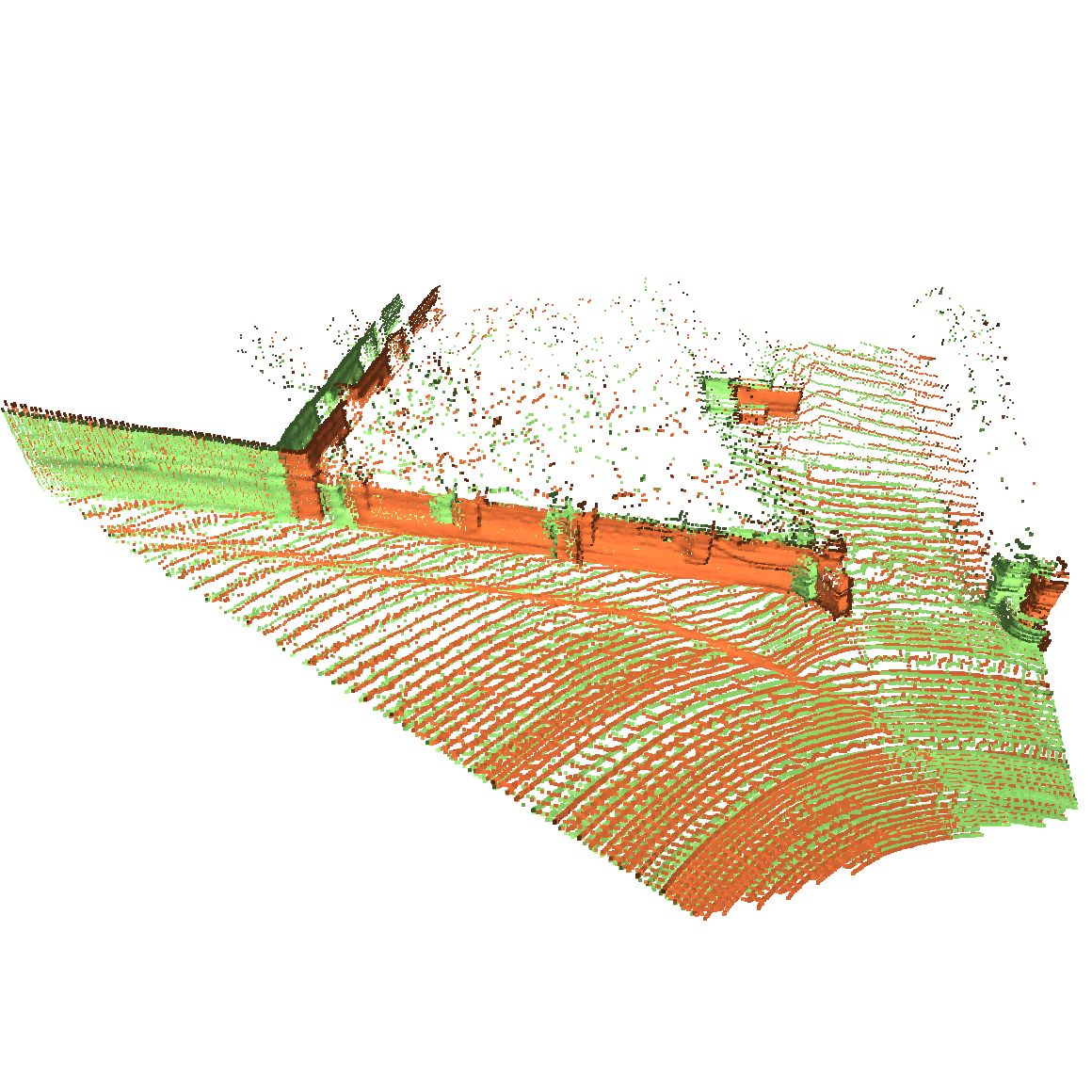} &
        \includegraphics[height=0.16\textwidth]{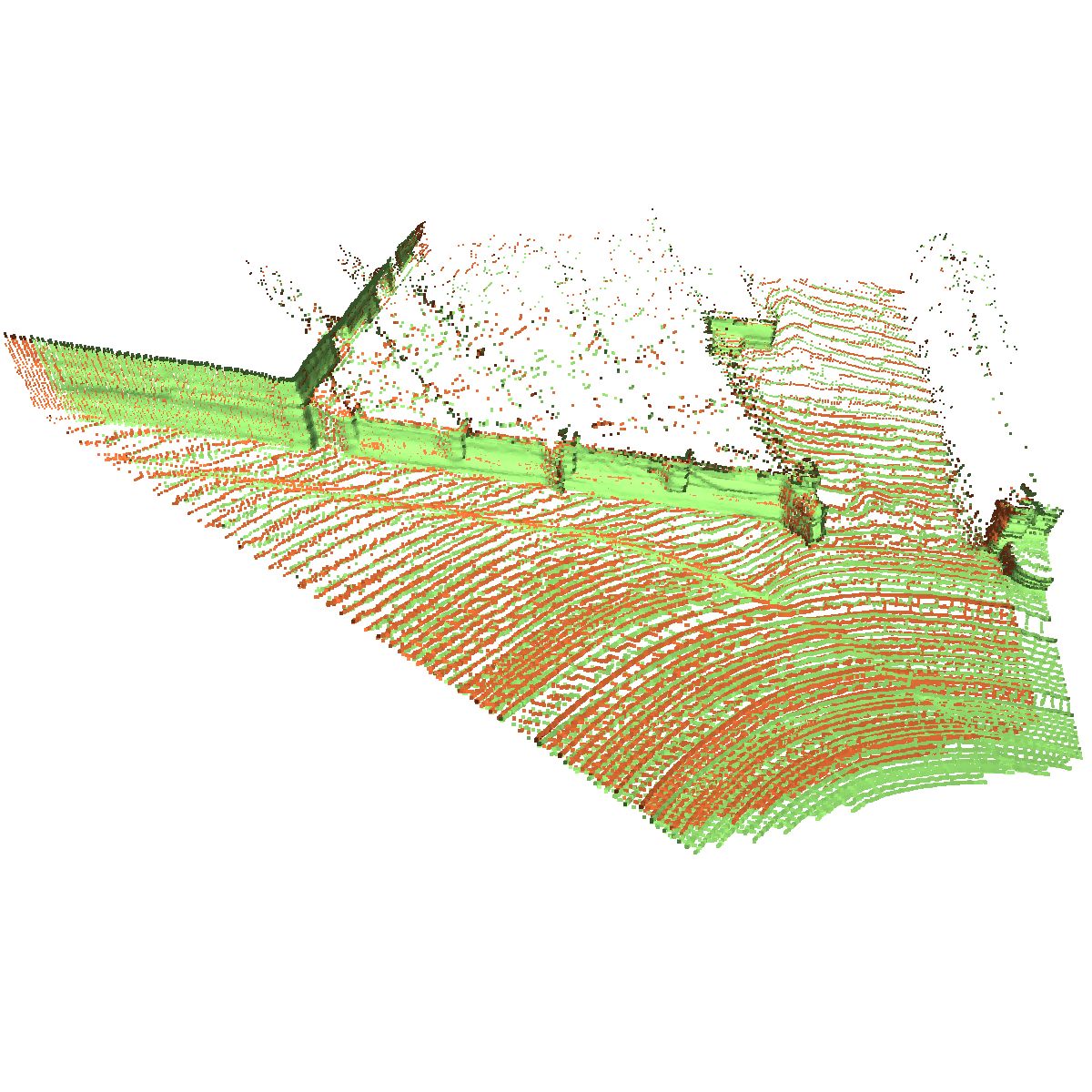}\vspace{0.2cm}\\
        \begin{tabular}{c}\vspace{-2.8cm}\\Kitti\\Test Data\\\#15-449\end{tabular} &
        \includegraphics[height=0.16\textwidth]{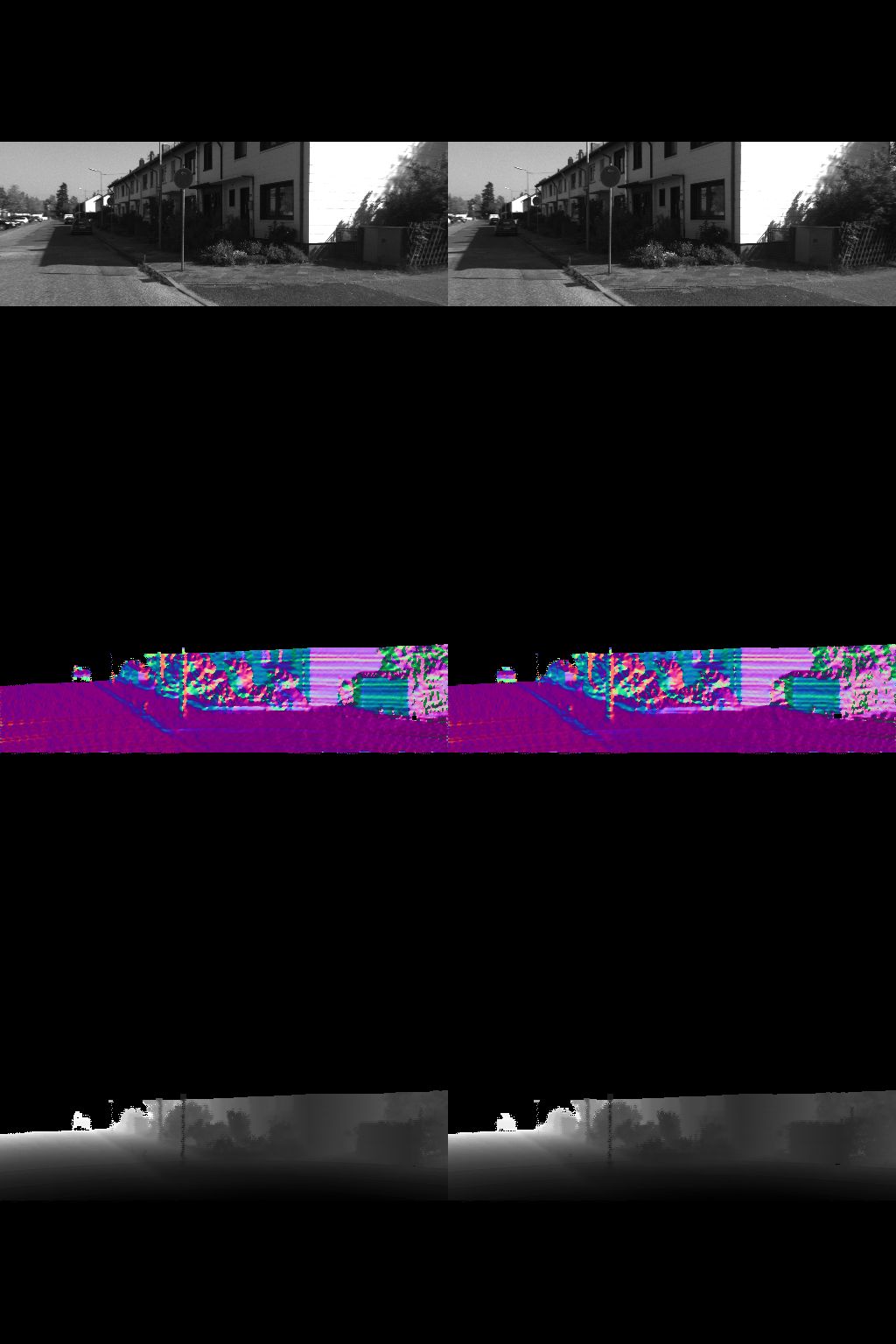} &
        \includegraphics[height=0.16\textwidth]{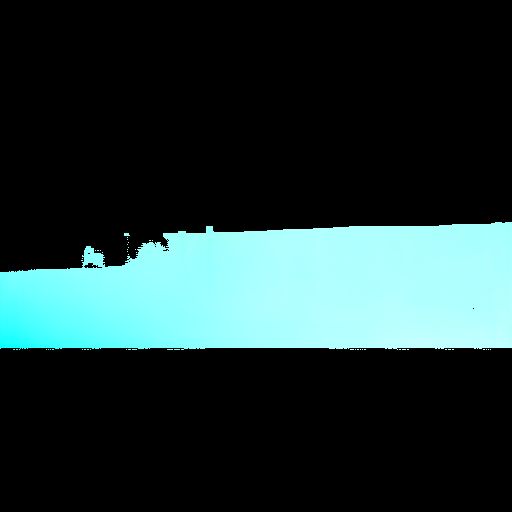} &\begin{tabular}{c}\vspace{-2.8cm}\\n.a.\end{tabular}& 
        \includegraphics[height=0.16\textwidth]{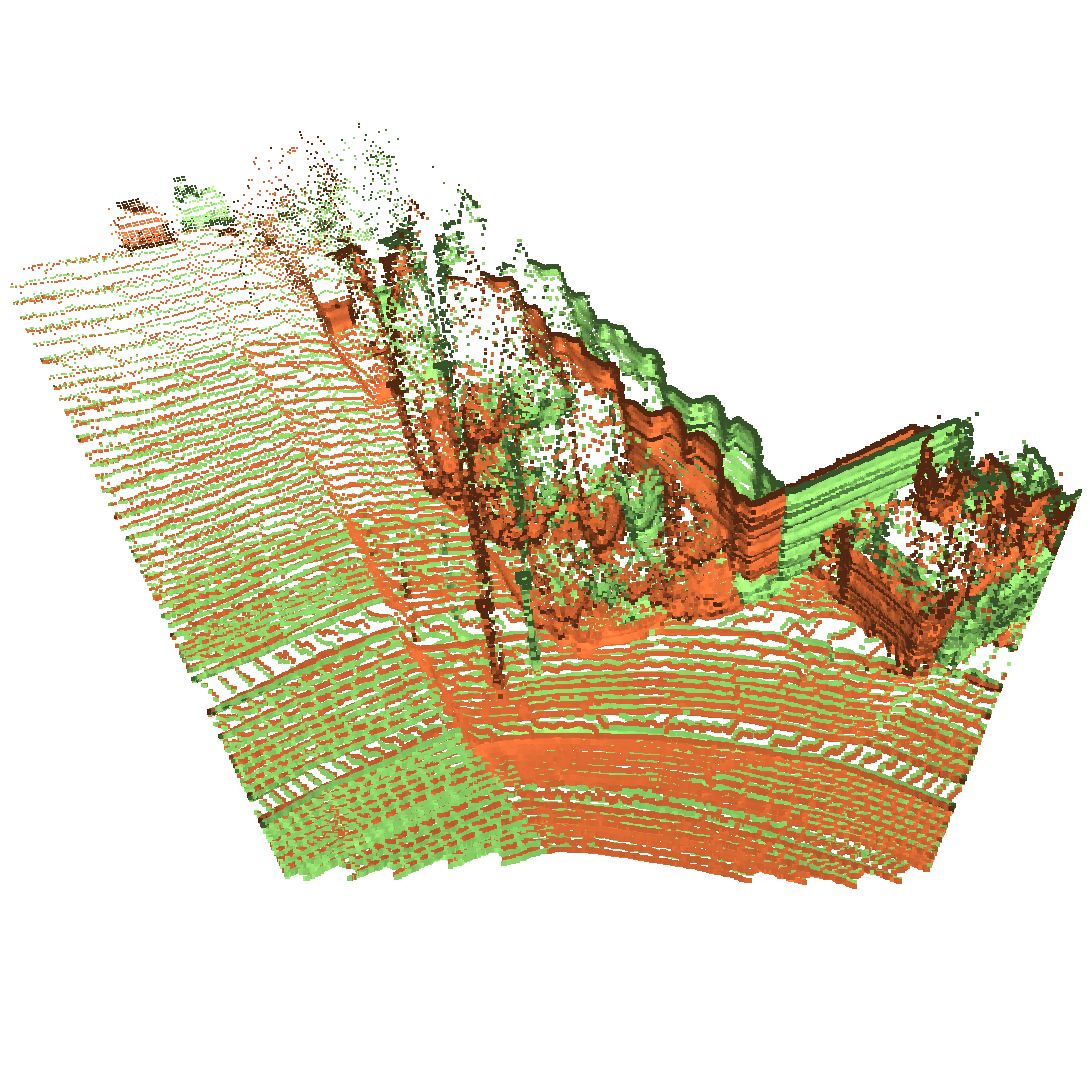} &
        \includegraphics[height=0.16\textwidth]{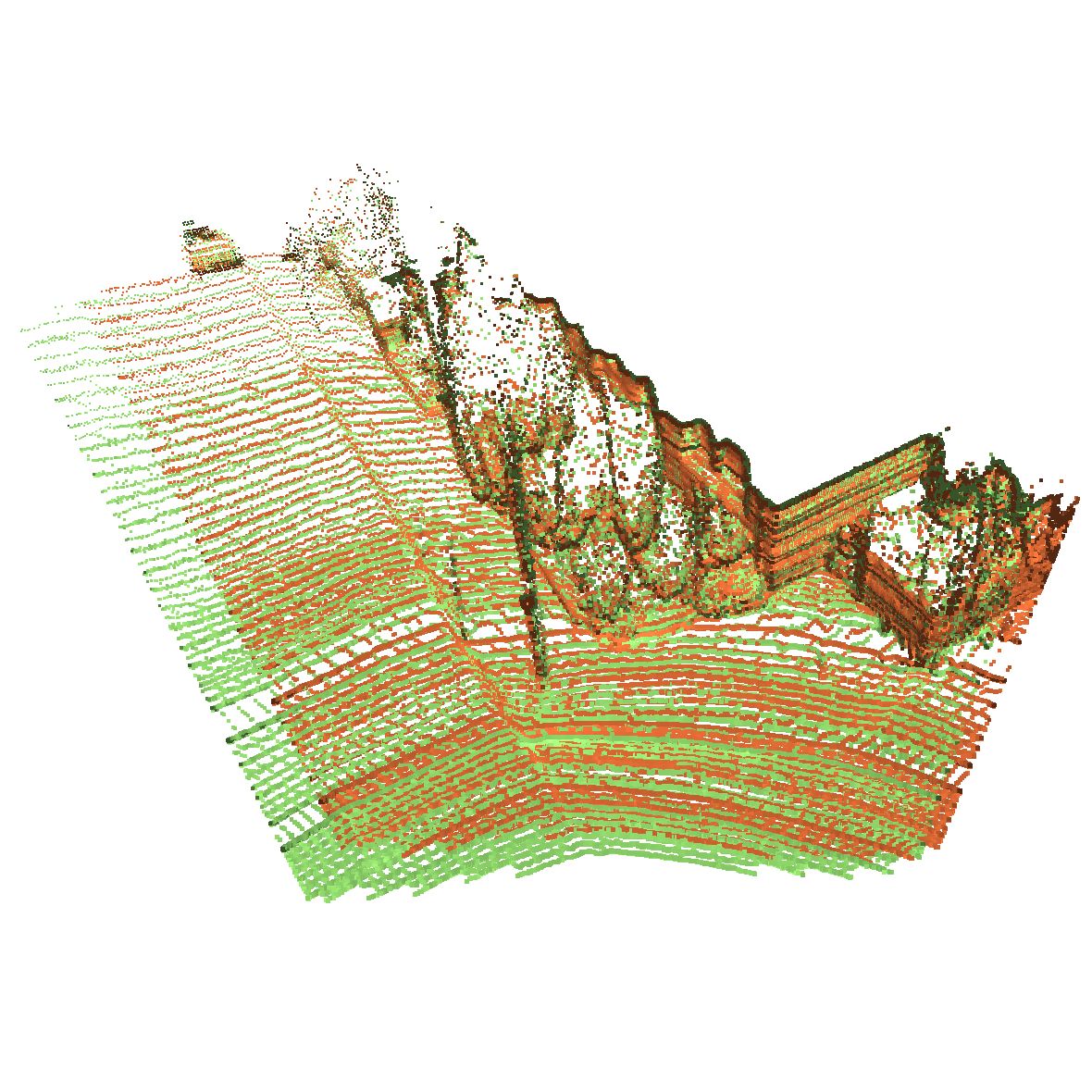}\vspace{0.2cm}\\
        \begin{tabular}{c}\vspace{-2.8cm}\\Kitti\\Test Data\\\#18-295\end{tabular} &
        \includegraphics[height=0.16\textwidth]{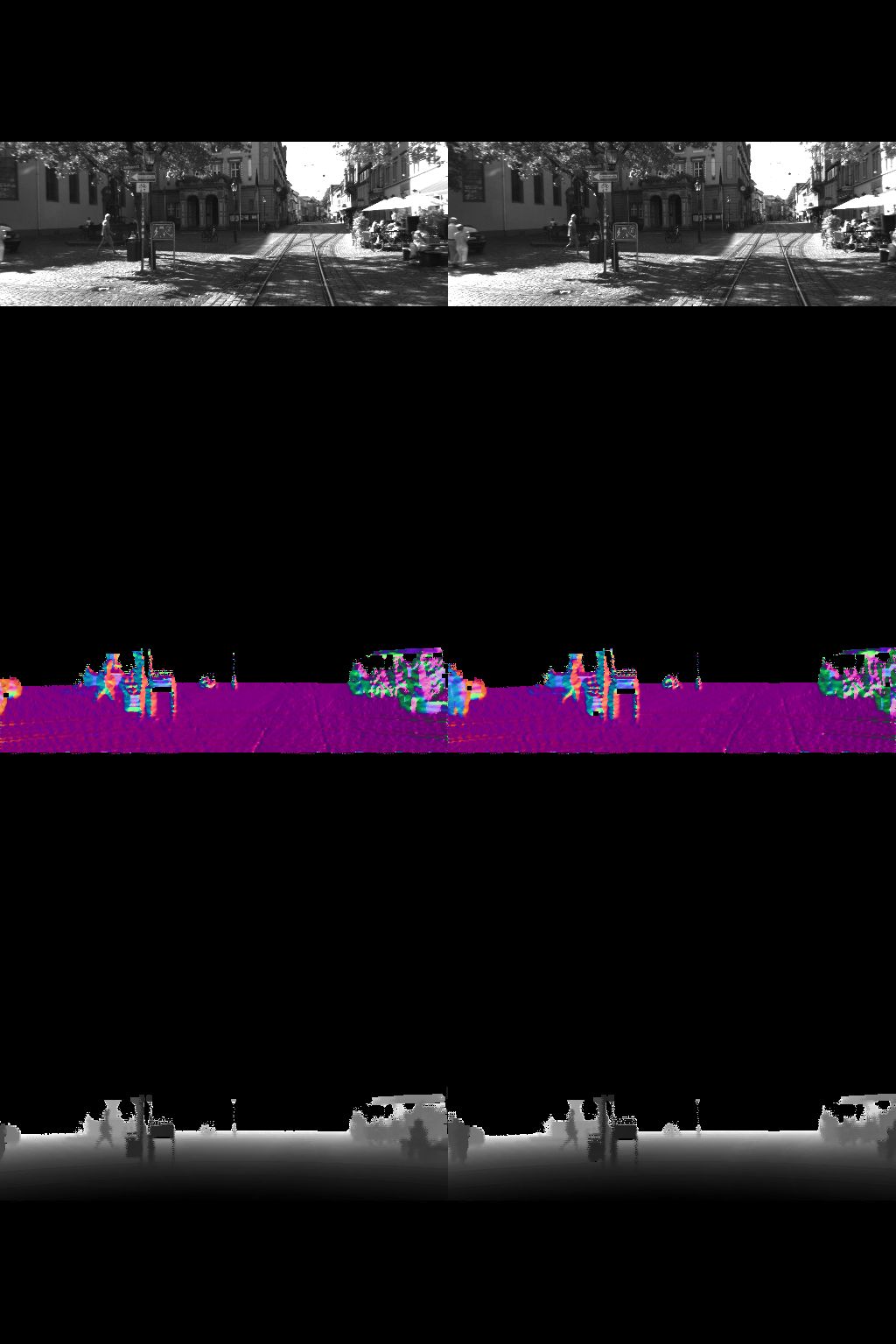} &
        \includegraphics[height=0.16\textwidth]{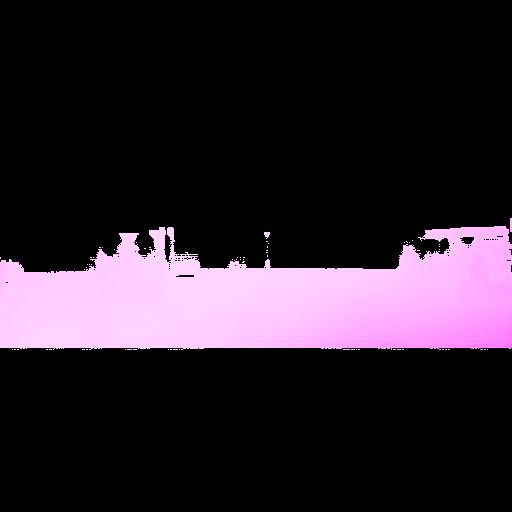} &\begin{tabular}{c}\vspace{-2.8cm}\\n.a.\end{tabular}& 
        \includegraphics[height=0.16\textwidth]{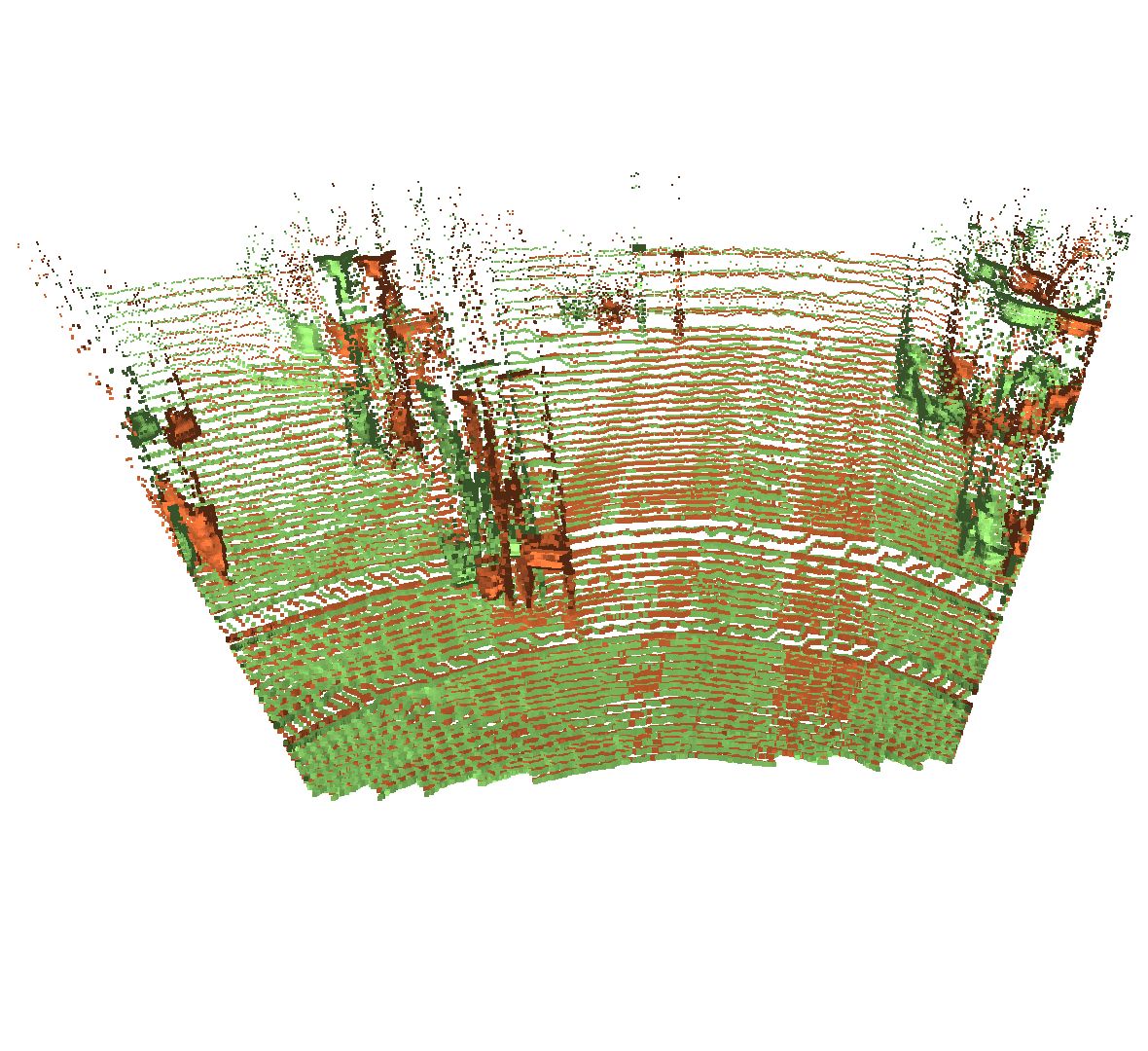} &
        \includegraphics[height=0.16\textwidth]{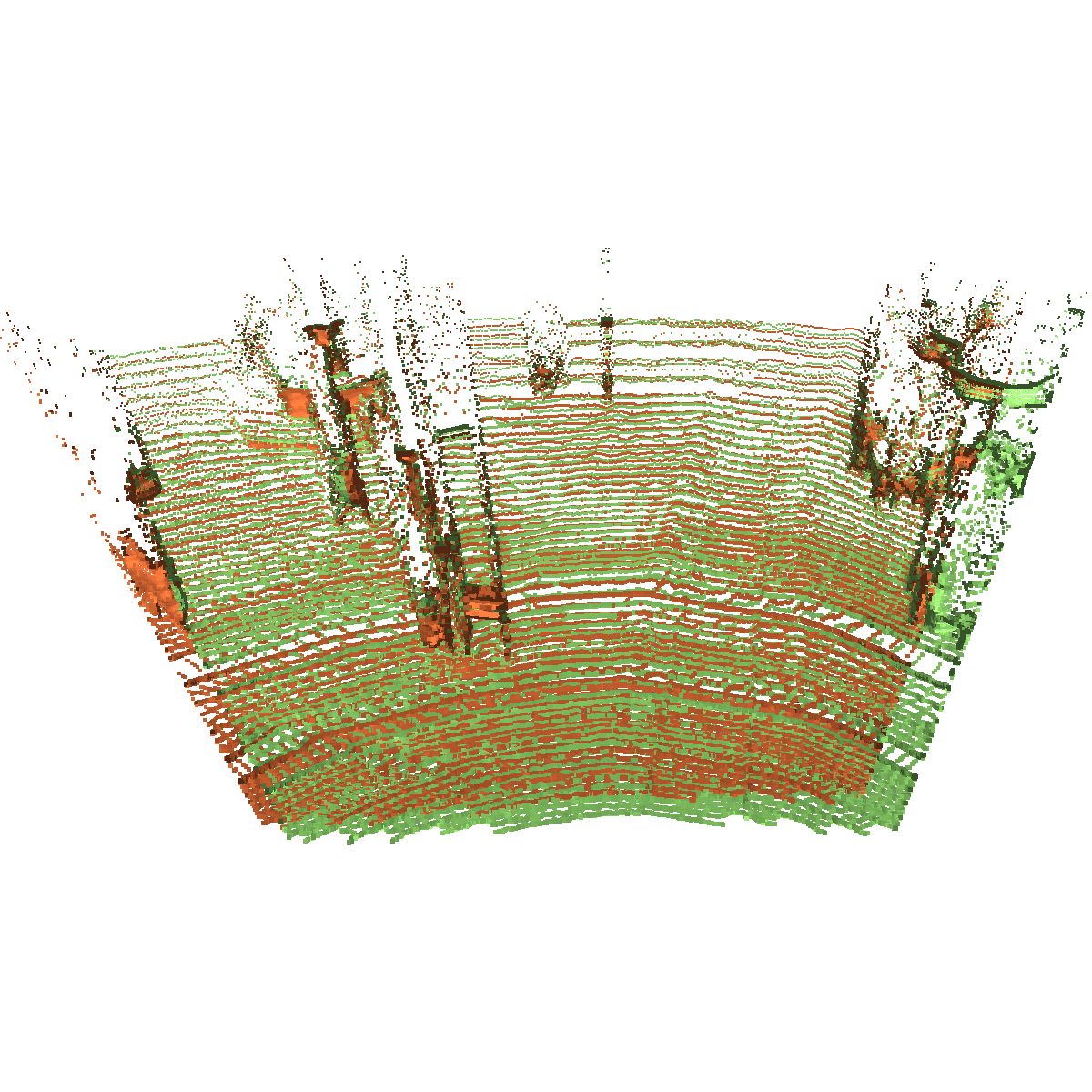}
    \end{tabular}
    \captionof{figure}{Qualitative results of the proposed method on training and test data of the \textit{Kitti Odometry} dataset. The method also works on this kind of scenario with less rotations and less shading changes than in the mainly investigated case, but also handles noise resulting from the lidar depth mesurement in the Kitti data. The network generalizes well from known training to unknown test data.}
    \label{Figure:Kitti}
\end{table*}

\subsection{Quantitative Evaluation}
For quantitative evaluation we first compare the different architectures (\textit{1 Step} and \textit{3 Step}) on the datasets published together with this work.
Table \ref{TableConsistentLight} shows the results on the full subsets with \textit{consistent light} and \textit{inconsistent light}.
In both cases the \textit{3 Step} method yields superior results in comparison to the standard procedure that directly predicts rotation and translation jointly.
Especially the resulting rotation is much more accurate, resulting in an alignment error that is up to 3 times smaller than in the popular standard prediction method.\vspace{-0.2cm}\\
\begin{center}
\begin{tabular}{|c|c|c|c|c|c|}
    \hline
    \multicolumn{2}{|c|}{Light} & \multicolumn{2}{|c|}{Consistent} & \multicolumn{2}{|c|}{Inconsistent}\\
    \hline   
    Data Type & Method & EPE & AE & EPE & AE\\
    \hline
    \hline
    Train Data & 1 Step & 1.83 & 0.035 & 2.33 & 0.035        \\
    Train Data & 3 Step & 1.83 & \textbf{0.012} & 2.33 & \textbf{0.013}\\
    Test Data  & 1 Step & 4.09 & 0.037 & 8.08 & 0.048         \\
    Test Data  & 3 Step & 4.09 & \textbf{0.023} & 8.08 & \textbf{0.035}\\
    \hline
\end{tabular}
\end{center}
\vspace{-0.2cm}
\captionof{table}{Quantitative comparison of the \textit{1 Step} and the proposed \textit{3 Step} methods to predict the pose from given warped vertex and normal maps.\vspace{0.3cm}}
\label{TableConsistentLight}

For completeness, we also trained the proposed architecture on the famous \textit{Kitti Odometry} dataset. 
As mentioned, the data does not reflect the situations, where the strengths of the proposed architecture comes up.
Also there are many procedures that are tuned to solve especially this common dataset.
Nevertheless our architecture is also able to deliver results that place within the ranking.
Table \ref{TableKitti} shows the methods, that are also based on point clouds and therefore somehow comparable to the method presented.
Our method would place around rank \#100, which shows the opportunity of the method to be also applicable to other tasks. 

\subsection{Predicted Dense Optical Flow}
\begin{figure}[h!]
\centering
    \begin{tabular}{p{0.07\textwidth}p{0.105\textwidth}p{0.1\textwidth}p{0.1\textwidth}}
    \ Input & Sparse GT & Predicted & Dense GT
    \end{tabular}
    \subfloat[Consistent Light: Train Object]{\includegraphics[height=0.12\textwidth]{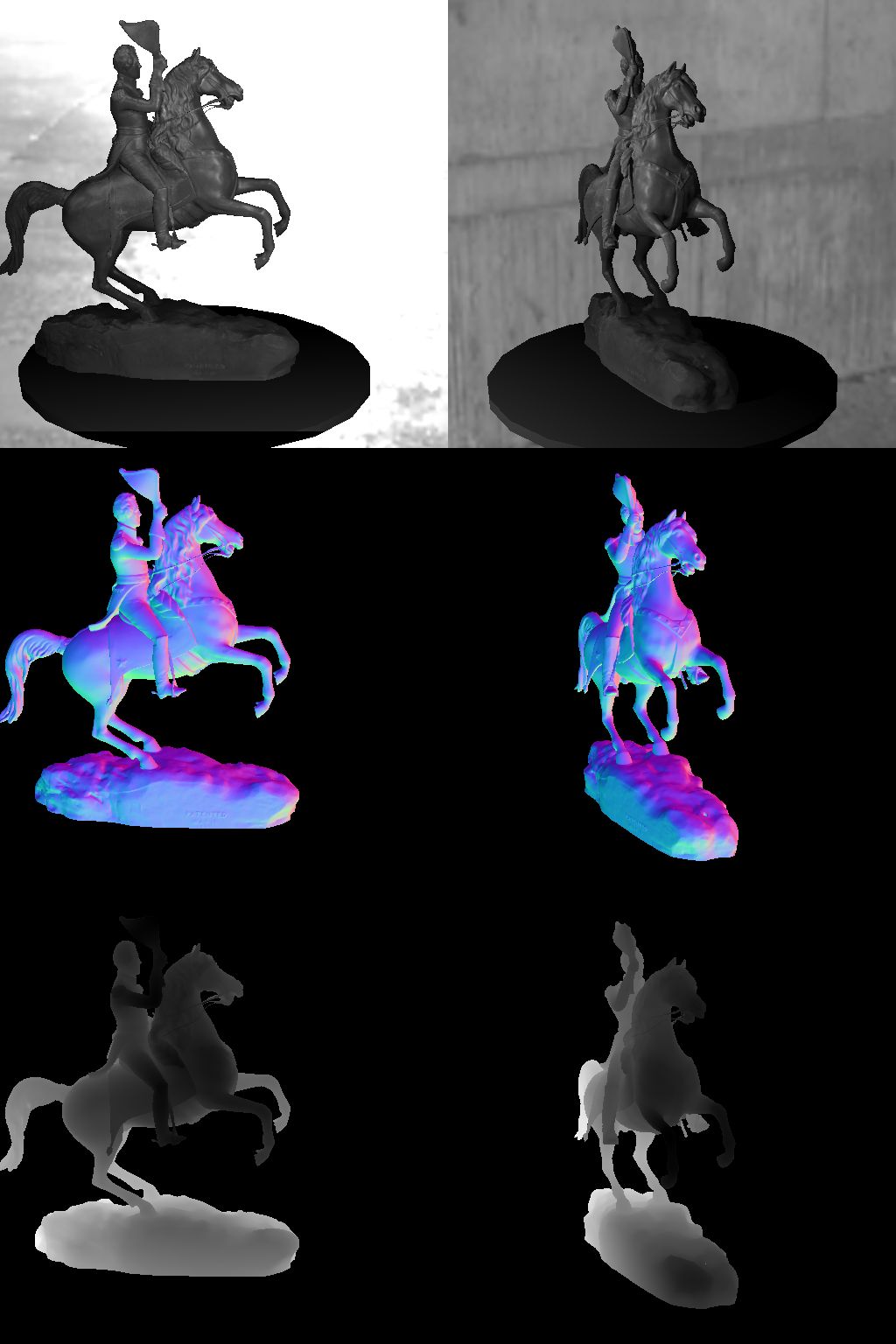}\ 
    \includegraphics[height=0.12\textwidth]{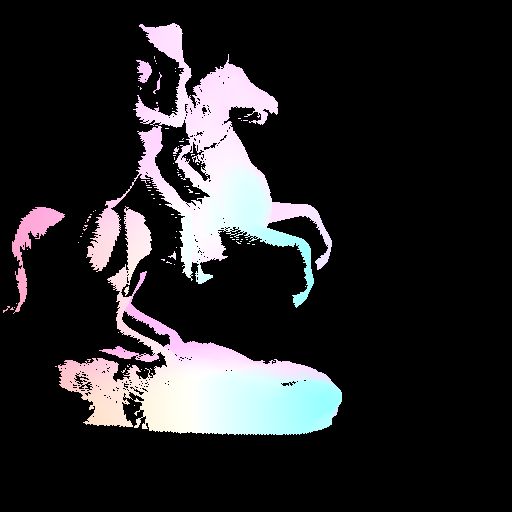}\ 
    \includegraphics[height=0.12\textwidth]{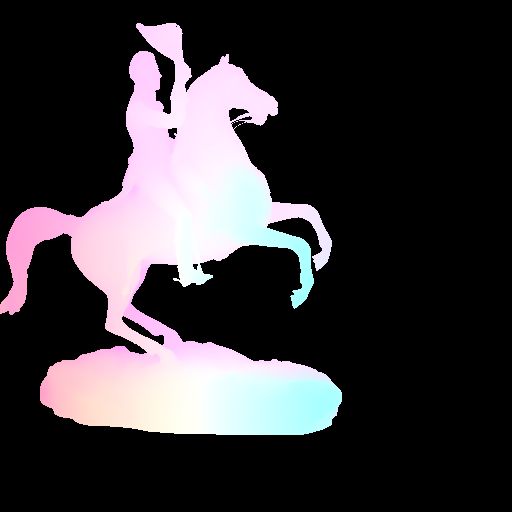}\ 
    \includegraphics[height=0.12\textwidth]{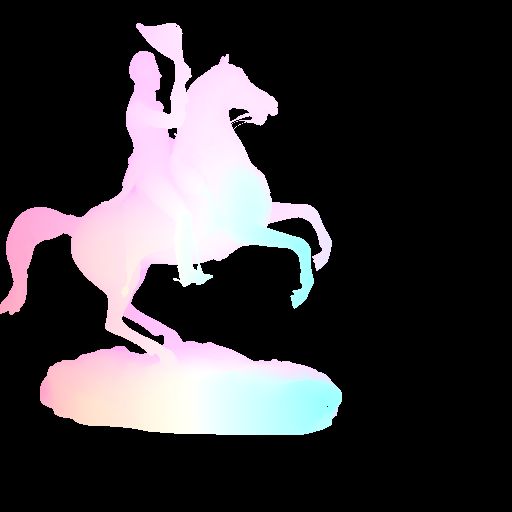}}\\
    \subfloat[Consistent Light: Test Object]{\includegraphics[height=0.12\textwidth]{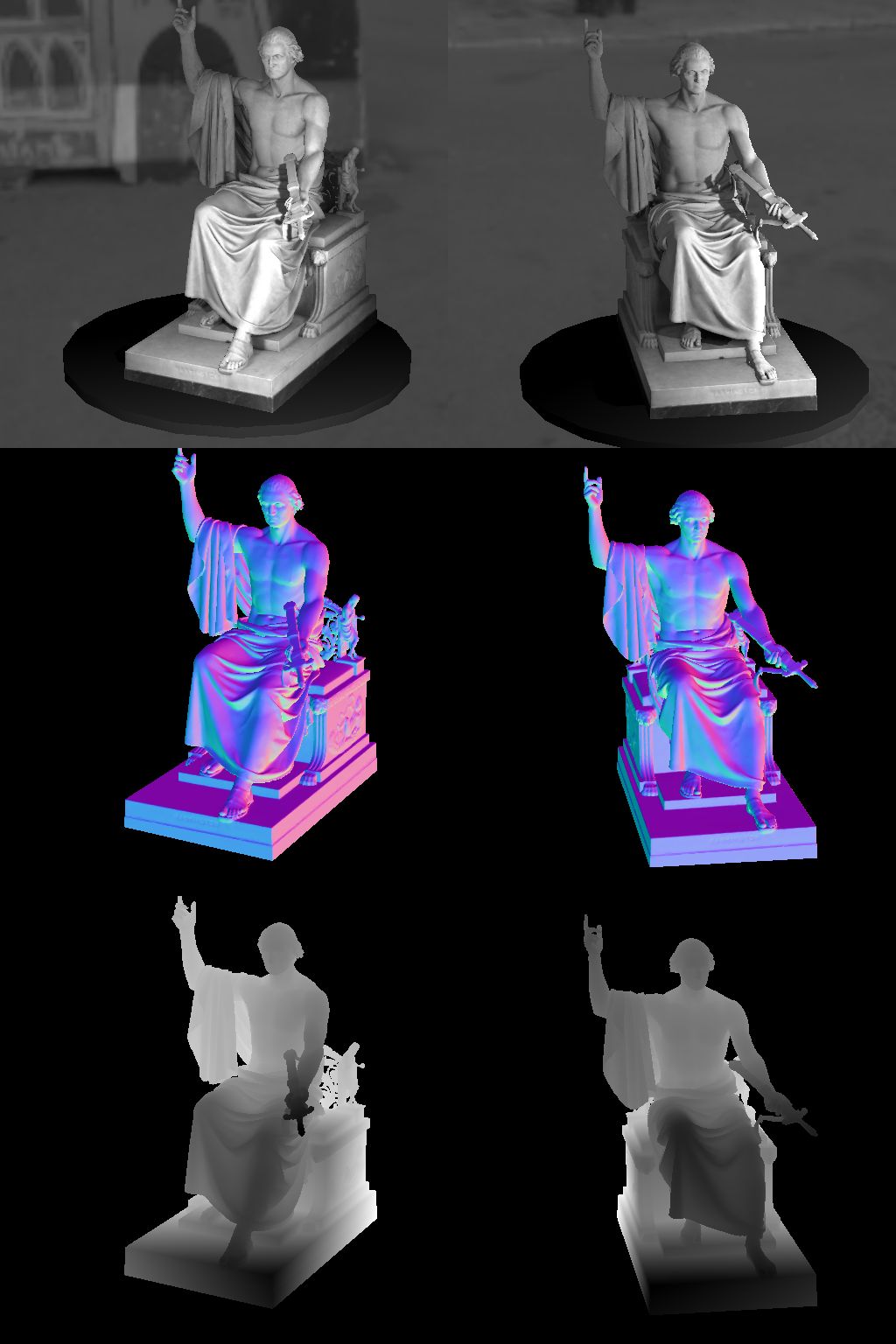}\ 
    \includegraphics[height=0.12\textwidth]{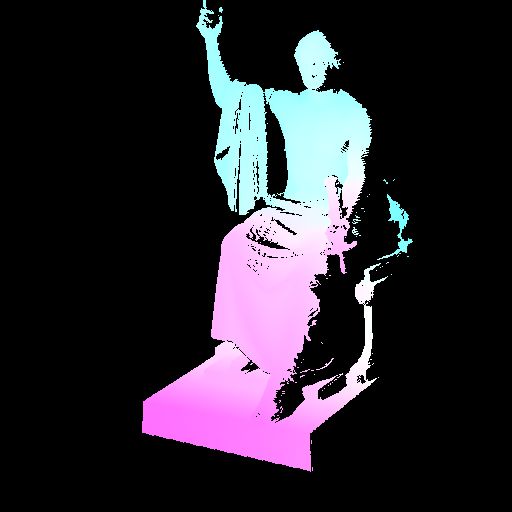}\ 
    \includegraphics[height=0.12\textwidth]{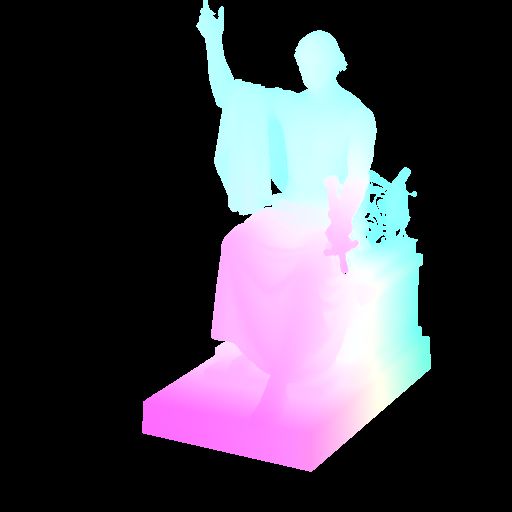}\ 
    \includegraphics[height=0.12\textwidth]{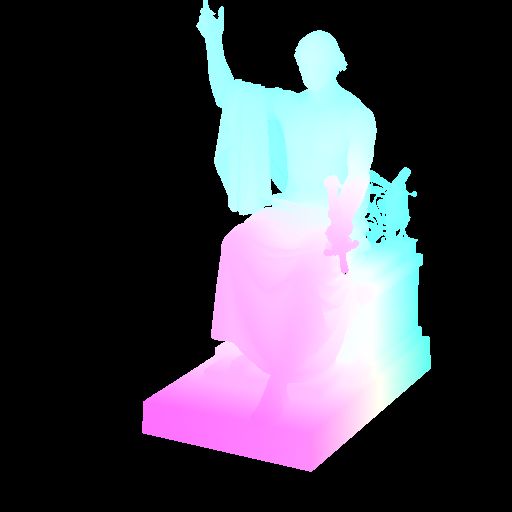}}\\
    \subfloat[Inconsistent Light: Train Object]{\includegraphics[height=0.12\textwidth]{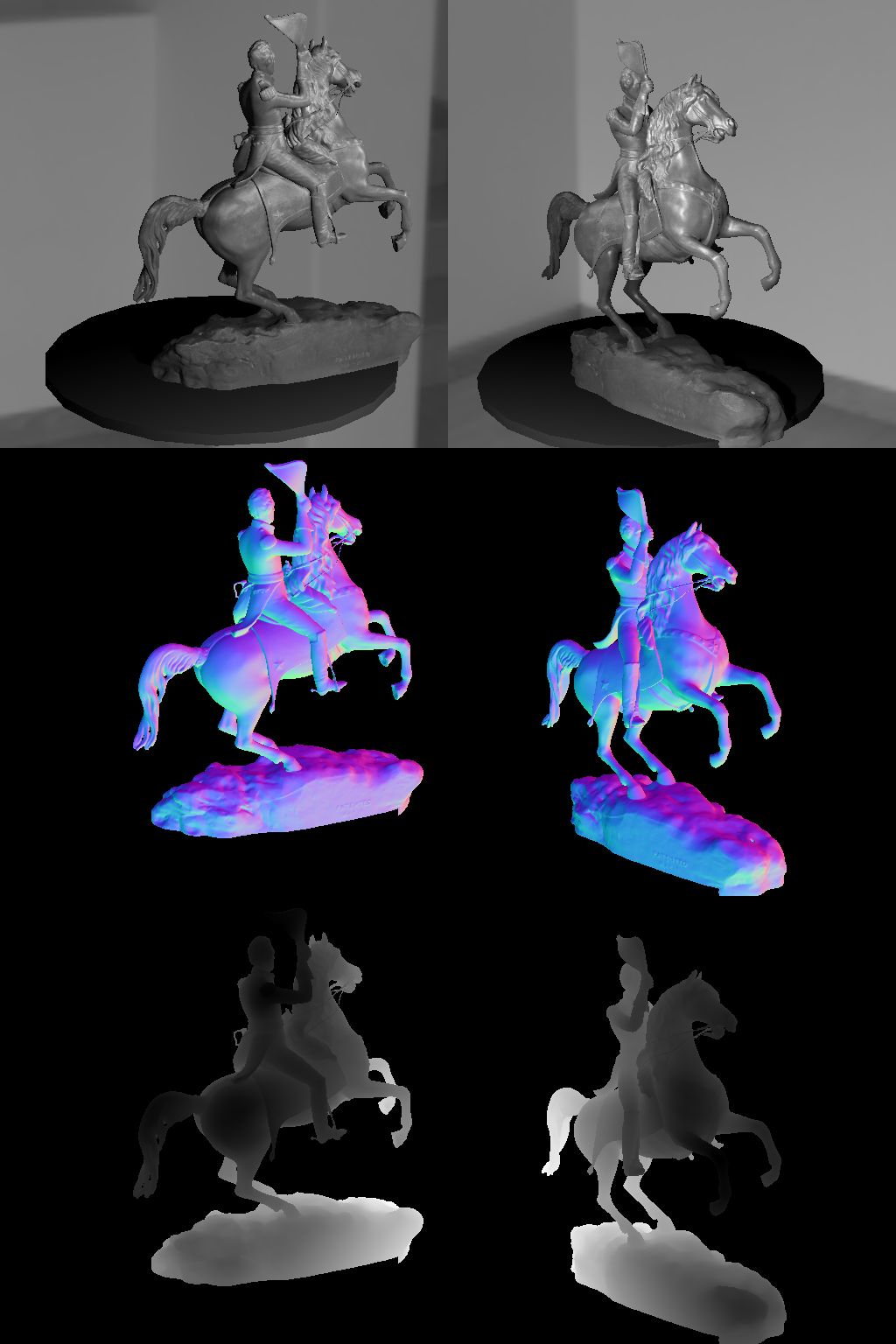}\ 
    \includegraphics[height=0.12\textwidth]{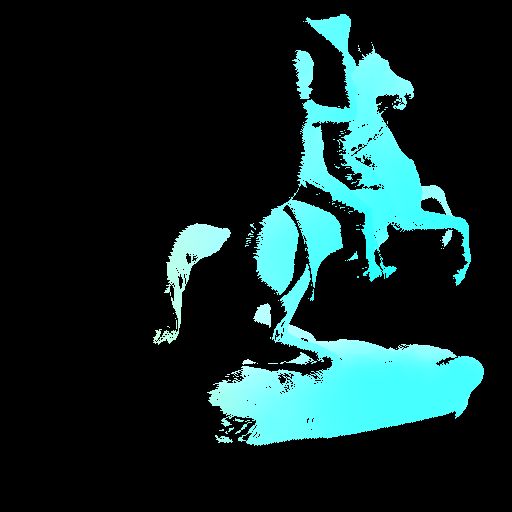}\ 
    \includegraphics[height=0.12\textwidth]{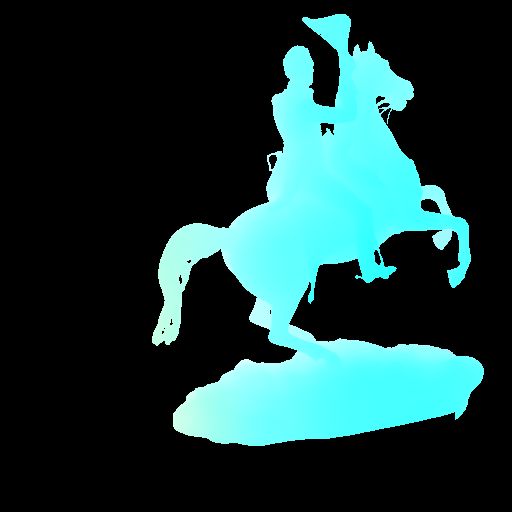}\ 
    \includegraphics[height=0.12\textwidth]{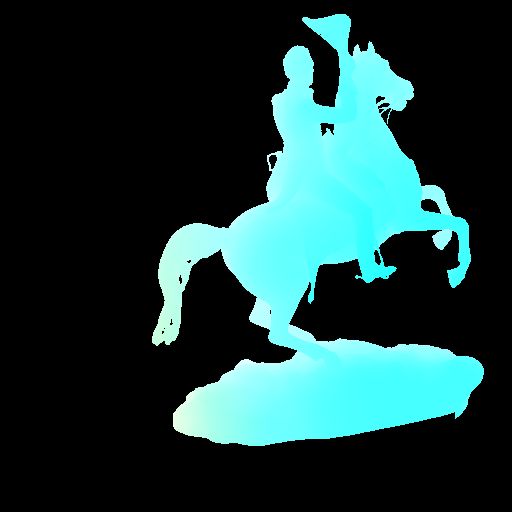}}\\
    \subfloat[Inconsistent Light: Test Object]{\includegraphics[height=0.12\textwidth]{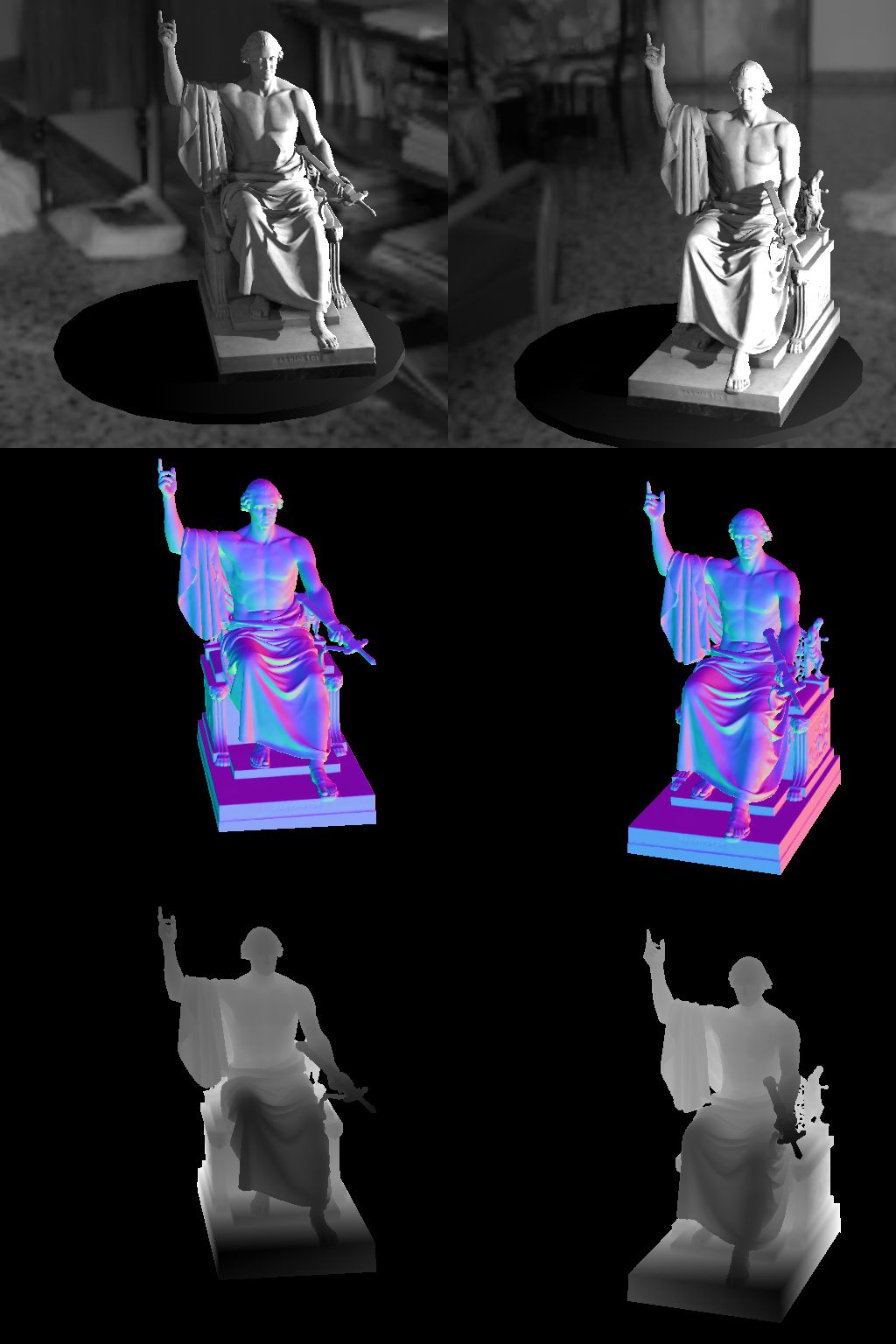}\ 
    \includegraphics[height=0.12\textwidth]{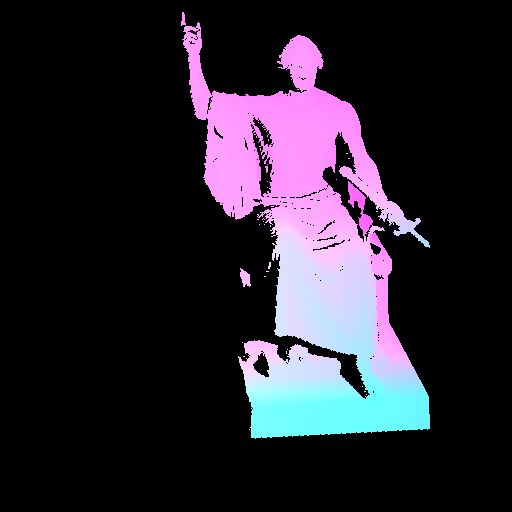}\ 
    \includegraphics[height=0.12\textwidth]{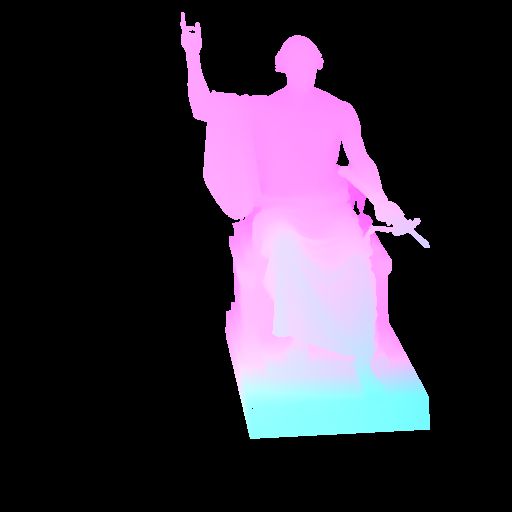}\ 
    \includegraphics[height=0.12\textwidth]{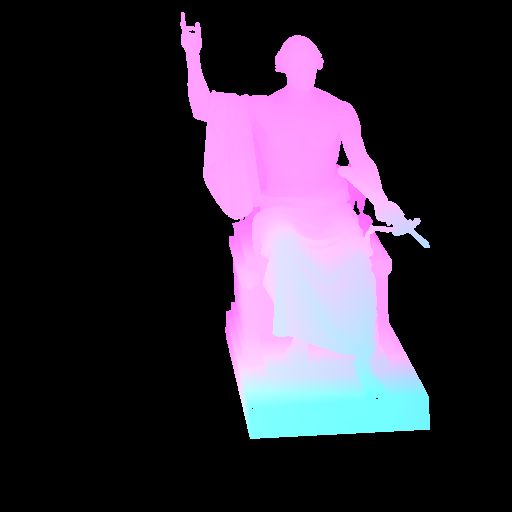}}
    \caption{Qualitative results of the predicted (dense) optical flow. The network allows to compute accurate flows for invisible pixels from context of visible parts as well for the consistent as for the inconsistent data.\vspace{0.05cm}}
    \label{Figure:DenseFlow}
\end{figure}

A special feature of the proposed method is its coarse to fine pyramidal optical flow base, combined with the rigid pose extraction.
Therefore one can assume, that the optical flow predicting sub-network learns rigidity relations from the extractability of the rigid pose from the dense optical flow.
As shown in Figure \ref{Figure:DenseFlow}, the ground truth optical flow (column 2), that has been used for training and evaluating the networks, is sparse, as it only contains the flow of points, that are visible in both views.
As the data is created synthetically, it is possible to also render dense ground truth optical flows (column 4), that contain the flow of points, that are occluded in one of the views and therefore may not be computable at all by the network.
As can be seen, the predicted optical flow (column 3) is dense.
It also predicts flow values for points that are not visible in both
views. 
These values result from context of other points, where the flow can be estimated stably.
The network learns how the flow behaves for rigid objects and transfers the knowledge to interpolated pixels.
This works as well for 
\begin{tabular}{|c|c|c|c|c|c|}
    \hline   
    Method & EPE & AE & R & t\\
    \hline
    \hline
    V-LOAM (\#2) & n.a. & n.a. & \textbf{0.0013} & \textbf{0.54} \\\hline
    LOAM (\#3) & n.a. & n.a. & 0.0013 & 0.55 \\\hline
    GLIM (\#5) & n.a. & n.a. & 0.0015 & 0.59 \\\hline
    CT-ICP (\#6) & n.a. & n.a. & 0.0014 & 0.59 \\\hline
    SDV-LOAM (\#7) & n.a. & n.a. & 0.0015 & 0.60 \\\hline
    CT-ICP2 (\#8) & n.a. & n.a. & 0.0012 & 0.60 \\\hline
    wPICP (\#11) & n.a. & n.a. & 0.0015 & 0.62 \\\hline
    FBLO (\#12) & n.a. & n.a. & 0.0014 & 0.62 \\\hline
    HMLO (\#14) & n.a. & n.a. & 0.0014 & 0.62 \\\hline
    filter-reg (\#16) & n.a. & n.a. & 0.0016 & 0.65 \\\hline
    MULLS (\#19) & n.a. & n.a. & 0.0019 & 0.65 \\\hline
    SMTD-LO (\#22) & n.a. & n.a. & 0.0020 & 0.66 \\\hline
    PICP (\#23) & n.a. & n.a. & 0.0018 & 0.67 \\\hline
    ELO (\#24) & n.a. & n.a. & 0.0021 & 0.68 \\\hline
    IMLS-SLAM (\#25) & n.a. & n.a. & 0.0018 & 0.69 \\\hline
    MC2SLAM (\#26) & n.a. & n.a. & 0.0016 & 0.69 \\\hline
    ISC-LOAM (\#28) & n.a. & n.a. & 0.0022 & 0.72 \\\hline
    Test-W (\#30) & n.a. & n.a. & 0.0033 & 0.79 \\\hline
    PSF-LO (\#31) & n.a. & n.a. & 0.0032 & 0.82 \\\hline
    S4-SLAM2 (\#35) & n.a. & n.a. & 0.0097 & 0.83 \\\hline
    LIMO2\_GP (\#39) & n.a. & n.a. & 0.0022 & 0.84 \\\hline
    CAE-LO (\#40) & n.a. & n.a. & 0.0025 & 0.86 \\\hline
    LIMO2 (\#42) & n.a. & n.a. & 0.0022 & 0.86 \\\hline
    CPFG-slam (\#44) & n.a. & n.a. & 0.0025 & 0.87 \\\hline
    SD-DEVO (\#49) & n.a. & n.a. & 0.0028 & 0.88 \\\hline
    PNDT LO (\#50) & n.a. & n.a. & 0.0030 & 0.89 \\\hline
    LIMO (\#58) & n.a. & n.a. & 0.0026 & 0.93 \\\hline
    SuMa-MOS (\#67) & n.a. & n.a. & 0.0033 & 0.99 \\\hline
    SuMa++ (\#69) & n.a. & n.a. & 0.0034 & 1.06 \\\hline
    DEMO (\#74) & n.a. & n.a. & 0.0049 & 1.14 \\\hline
    \begin{tabular}{c}STEAM-L\\WNOJ (\#83)\end{tabular} & n.a. & n.a. & 0.0058 & 1.22 \\\hline
    LiViOdo (\#84) & n.a. & n.a. & 0.0042 & 1.22 \\\hline
    STEAM-L (\#87) & n.a. & n.a. & 0.0061 & 1.26 \\\hline
    SALO (\#93) & n.a. & n.a. & 0.0051 & 1.37 \\\hline
    SuMa (\#95) & n.a. & n.a. & 0.0034 & 1.39 \\\hline
    \hline\begin{tabular}{c}Flow2PoseNet\\3 Step\end{tabular}
    & 1.18 & 0.019 & 0.0019 & 2.73\\\hline\hline
    Deep-CLR (\#134) & n.a. & n.a. & 0.0104 & 3.83 \\\hline\hline
    SLL (\#163) & n.a. & n.a. & 0.2645 & 90.05 \\\hline
\end{tabular}
\captionof{table}{Extract of the ranking of the \textit{Kitti Odometry} dataset showing point cloud based methods. The proposed method would be placed within the ranking, although rather at the end. Nevertheless, this shows the additional applicability of the method to other highly studied tasks.\vspace{0.4cm}}
\label{TableKitti}
objects, that are known from the training set (rows 1 and 3) as for test objects, that have never been used for training (rows 2 and 4) as well for the \textit{ConsistentLight} case (rows 1 and 2) as for the \textit{InconsistentLight} case (rows 3 and 4).
Table \ref{QuantitativeEvalDenseFlow} moreover shows, that the resulting \textit{Endpoint Errors (EPE)} do not dramatically increase for the invisible points, which indicates, that the network learns to predict flows for the invisible points from context, according to the behavior of rigid objects.

\begin{tabular}{|c|c|c|}
    \hline
    \multicolumn{3}{|c|}{Consistent Light}\\
    \hline   
    Data Type & Visible Points EPE & Invisible Points EPE\\
    \hline
    \hline
    Train Data & 2.7446 & 3.4978\\
    Test Data  & 3.6411 & 4.9284\\
    \hline
\end{tabular}
\vspace{0.2cm}\\
\begin{tabular}{|c|c|c|}
    \hline
    \multicolumn{3}{|c|}{Inconsistent Light}\\
    \hline   
    Data Type & Visible Points EPE & Invisible Points EPE\\
    \hline
    \hline
    Train Data & 3.6974 & 5.4024\\
    Test Data  & 4.7996 & 4.7703\\
    \hline
\end{tabular}
\captionof{table}{Quantitative results for the visible and invisible points in the evaluated scenes. The resulting \textit{Endpoint Errors (EPE)} do not heavily increase.
The network is still able to predict accurate flows from context of visible points and to generalize to the test data as well for the consistent as for the inconsistent data.\vspace{0.5cm}}
\label{QuantitativeEvalDenseFlow}

\section{Conclusion}
In this paper, a method has been presented that combines optical flow estimation of rigid scenes with a posterior pose estimation.
In this way, including several contributions, a method has been developed that allows scenes with difficult lighting conditions to be registered in a stable way.

Optical flow is thereby estimated accurately using geometric, shading and texture features. 
The variety of different feature types allows the system to be trained to be illumination resistant (using geometric and normal features) without having to completely sacrifice potentially important texture features. 

The pose is then stably estimated from the warped normals and vertex maps using a new 3-step procedure.
This has, compared to typical approaches that directly infer the pose, significant advantages especially in cases with strong rotations that often cause the considered shading changes.

The combination of optical flow and rigid pose estimation allows the pose to benefit from the features of different levels of the underlying coarse-to-fine flow approach, which means that the method is not dependent on highly accurate features and can also align smooth scenes with weak features.
In turn, the optical flow sub-network learns a typical flow behavior of rigid scenes from the posterior estimability of the pose. 
This allows accurate dense estimates to be achieved, even for occluded areas based on context and overall learned behavior.

{\small
\bibliographystyle{ieee}
\bibliography{egbib}
}
\end{document}